\pdfoutput=1

\documentclass[11pt]{article}
\usepackage{hyperref}
\usepackage{url}
\usepackage{xurl}
\usepackage{graphicx}
\usepackage{booktabs}
\usepackage{amssymb}
\usepackage{amsmath}
\usepackage{subcaption}
\usepackage{xcolor}
\usepackage{adjustbox} 
\usepackage{multirow}
\usepackage{wrapfig}
\usepackage{tikz}
\usepackage{xspace}
\usepackage{soul}
\newcommand{\ie}{\textit{i}.\textit{e}., }
\newcommand{\eg}{\textit{e}.\textit{g}., }
\usepackage{titletoc}

\definecolor{bg_blue}{RGB}{213,227,251}
\definecolor{bg_yellow}{RGB}{250,243,187}
\definecolor{bg_purple}{RGB}{177,167,207}
\definecolor{bg_red}{RGB}{200,169,188}
\definecolor{bg_green}{RGB}{192,213,175}
\definecolor{bg_skin}{RGB}{245,232,210}

\definecolor{red_color}{RGB}{255,0,0}
\newcommand{\redtext}[1]{\textcolor{red_color}{#1}}
\definecolor{yellow_color}{RGB}{255,202,47}

\definecolor{purple_color}{RGB}{64,103,139}
\newcommand{\purpletext}[1]{\textcolor{purple_color}{#1}}

\definecolor{dark_red}{RGB}{153, 31, 41}

\definecolor{green_color}{RGB}{130,139,78}
\newcommand{\greentext}[1]{\textcolor{green_color}{#1}}
\definecolor{brown_color}{RGB}{205,90,161}

\definecolor{lg_color}{RGB}{63,147,139}

\definecolor{com_color}{RGB}{0,0,139}

\definecolor{orange_color}{RGB}{255,148,63}
\newcommand{\orangetext}[1]{\textcolor{orange_color}{#1}}

\definecolor{gray_color}{RGB}{169,169,169}

\definecolor{lightgray}{RGB}{220,220,220}

\definecolor{lightgreen}{RGB}{179,207,176}

\definecolor{lightblue}{RGB}{181,209,230}

\newcommand{\model}{\mbox{\sc LLM-Rec}\xspace}

\usepackage{acl}

\usepackage{times}
\usepackage{latexsym}

\usepackage[T1]{fontenc}

\usepackage[utf8]{inputenc}

\usepackage{microtype}

\usepackage{inconsolata}
\usepackage{adjustbox}

\title{LLM-Rec: Personalized Recommendation via \\ Prompting Large Language Models}

\author{Hanjia Lyu\textsuperscript{1}, Song Jiang\textsuperscript{2}, Hanqing Zeng\textsuperscript{3}, Yinglong Xia\textsuperscript{3}, Qifan Wang\textsuperscript{3}, \\ \textbf{Si Zhang\textsuperscript{3}, Ren Chen\textsuperscript{3}, Christopher Leung\textsuperscript{3}, Jiajie Tang\textsuperscript{3}, Jiebo Luo\textsuperscript{1}} \\
\textsuperscript{1}University of Rochester \quad \textsuperscript{2}UCLA \quad \textsuperscript{3}Meta AI \\
\texttt{hlyu5@ur.rochester.edu, jluo@cs.rochester.edu}}

\begin{document}
\maketitle

\begin{abstract}
Text-based recommendation holds a wide range of practical applications due to its versatility, as textual descriptions can represent nearly any type of item. However, directly employing the original item descriptions may not yield optimal recommendation performance due to the lack of comprehensive information to align with user preferences. Recent advances in large language models (LLMs) have showcased their remarkable ability to harness commonsense knowledge and reasoning. In this study, we introduce a novel approach, coined \model, which incorporates four distinct prompting strategies of text enrichment for improving personalized text-based recommendations. Our empirical experiments reveal that using LLM-augmented text significantly enhances recommendation quality. Even basic MLP (Multi-Layer Perceptron) models achieve comparable or even better results than complex content-based methods. Notably, the success of \model lies in its prompting strategies, which effectively tap into the language model's comprehension of both general and specific item characteristics. This highlights the importance of employing diverse prompts and input augmentation techniques to boost the recommendation effectiveness of LLMs.
\end{abstract}

\section{Introduction}

Text-based recommendation systems exhibit a broad spectrum of applications, spanning across diverse domains and industries. This versatility mainly stems from the capability of natural language to effectively describe nearly \textit{any} type of items, encompassing not only products, movies, and books but also news articles and user-generated content, including short videos and social media posts~\citep{pazzani2007content, javed2021review, poirier2010towards,bai2022improving, wu-etal-2020-mind, oppermann2020vizcommender, chen2017joint, gupta2017scientific, wang2018explainable}. Nonetheless, these text-based recommendation systems are frequently challenged by the inherent limitation of \textbf{incomplete or insufficient information within item descriptions}, which hinders the task of accurately \textit{aligning} item characteristics with user preferences~\citep{perez2007building, dumitru2011demand}. The incompleteness may arise from two sources: a limited comprehension of the items themselves and an insufficient understanding of the users for whom recommendations are generated.

\begin{figure}[t]
    \centering
    \includegraphics[width=\linewidth]{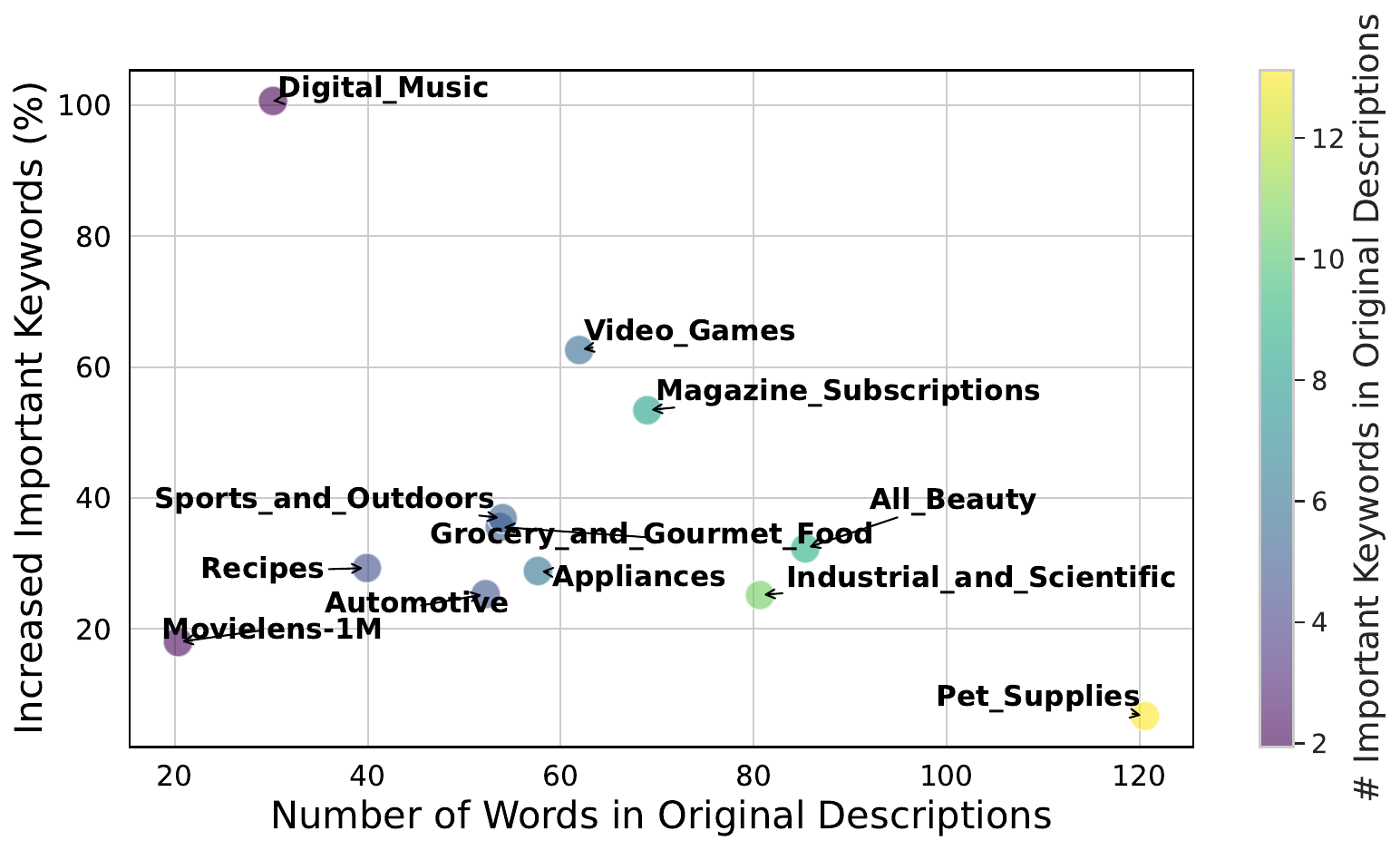}
    \caption{\model enhances original item descriptions by prompting LLMs to augment important keywords (\eg adjectives). It applies to various domains and is not limited to datasets with rich textual content.}
    \label{fig:applicable_domain}
\end{figure}

This challenge is not confined only to domains with well-defined and categorized items (\eg movies), but also extends to domains characterized by novel, unclassified, or less categorically structured items, as observed in the case of user-generated content. In the context of movie recommendations, a movie's description usually include the main actors, and a brief plot summary. However, this limited information may not capture crucial elements like genre, tone, cinematography style, or thematic depth, resulting in less effective recommendations.
As for user-generated content, imagine a social platform where users regularly post recipes which are often accompanied with brief textual descriptions like the name of the dish and a few ingredients, but limited details regarding preparation time, dietary restrictions, or flavor profiles. Consider a user who follows a vegan diet and is interested in discovering new plant-based recipes. Since the user-generated content often lacks comprehensive dietary information and may not explicitly mention terms like ``vegan'', ``plant-based'', or ``vegetarian'', in this scenario, the recommendation system, relying solely on the incomplete descriptions, may struggle to discern the vegan-friendliness of the recipes. 

The recent advances in the development of large language models (LLMs) underscore their exceptional capacity to store comprehensive world knowledge~\citep{peters2018dissecting, goldberg2019assessing, tenney2019you, petroni2019language}, engage in complex reasoning~\citep{wei2022chain, zhou2022least}, and function as versatile task solvers~\citep{zhao2023survey,ouyang2022training, kaplan2020scaling}. In light of this advancement and recognizing the challenge posed by incomplete item descriptions, our study introduces the \model framework. This approach is designed to exploit various prompting strategies to \textit{enrich input text} with the intrinsic capabilities of LLMs for personalized recommendations. By leveraging LLMs, which have been fine-tuned on extensive language datasets~\citep{ouyang2022training, touvron2023llama}, our goal is to unlock their potential in generating input text that is not only contextually aware but also of high quality, thereby elevating the overall recommendation quality. 

Through comprehensive empirical experiments, we evaluate the effectiveness of the \model framework. We find that integrating the augmented text as the new input achieves comparable or even superior recommendation performance compared to more advanced content-based recommendation approaches that rely solely on the original item descriptions. Further in-depth analyses reveal that the devised prompting strategies prompt LLMs to generate words that represent both general and specific item characteristics. It is applicable in a diverse range of domains and is not limited to datasets with rich textual information (Figure~\ref{fig:applicable_domain}). Our study provides insights into the impact of different prompting strategies on recommendation performance and sheds light on the potential of leveraging LLMs for personalized recommendation.

\vspace{-.2cm}
\section{Related Work}
\vspace{-.2cm}
\model closely aligns with two research directions: (1) augmentation in text-based recommendation, and (2) LLM for recommendation. A comprehensive discussion is provided in Appendix~\ref{sec:detailed_related_work}. 

\noindent\textbf{Augmentation in Text-based Recommendation.} Text-based recommendation systems leverage natural language processing and machine learning techniques to provide personalized recommendations to users based on textual information~\citep{lops2019trends, qiang2020short}. However, the performance of such systems can be compromised when dealing with incomplete or insufficient textual information. To address this limitation, several studies have suggested strategies for enhancing textual information. For instance, \citet{li2010contextual} proposed to extract contextual cues from online reviews, leveraging these narratives to uncover users' preferences and underlying factors influencing their choices~\citep{sachdeva2020useful}. Other approaches infer linguistic attributes from diverse sources, including emotion, sentiment, and topic, to refine the modeling of both items and users~\citep{sun2015mining, sailunaz2019emotion, ramage2010characterizing,chen2010short}. Furthermore, some works explore the integration of external knowledge bases to enrich the contextual understanding of items~\citep{di2012linked, musto2018semantics}. In a more recent development, \citet{bai2022improving} introduced an approach that employs pre-trained language models to generate additional product attributes, such as product names, to augment item contextual information. Diverging from these prior approaches, our contribution is the \model framework, which employs large language models to enhance input text, providing a versatile solution for recommendations. A more detailed discussion on the distinctions between \model and these related work can be found in Section~\ref{sec:discussions_and_conclusions}.

\begin{figure*}[ht]
    \centering
    \includegraphics[width=0.9\linewidth]{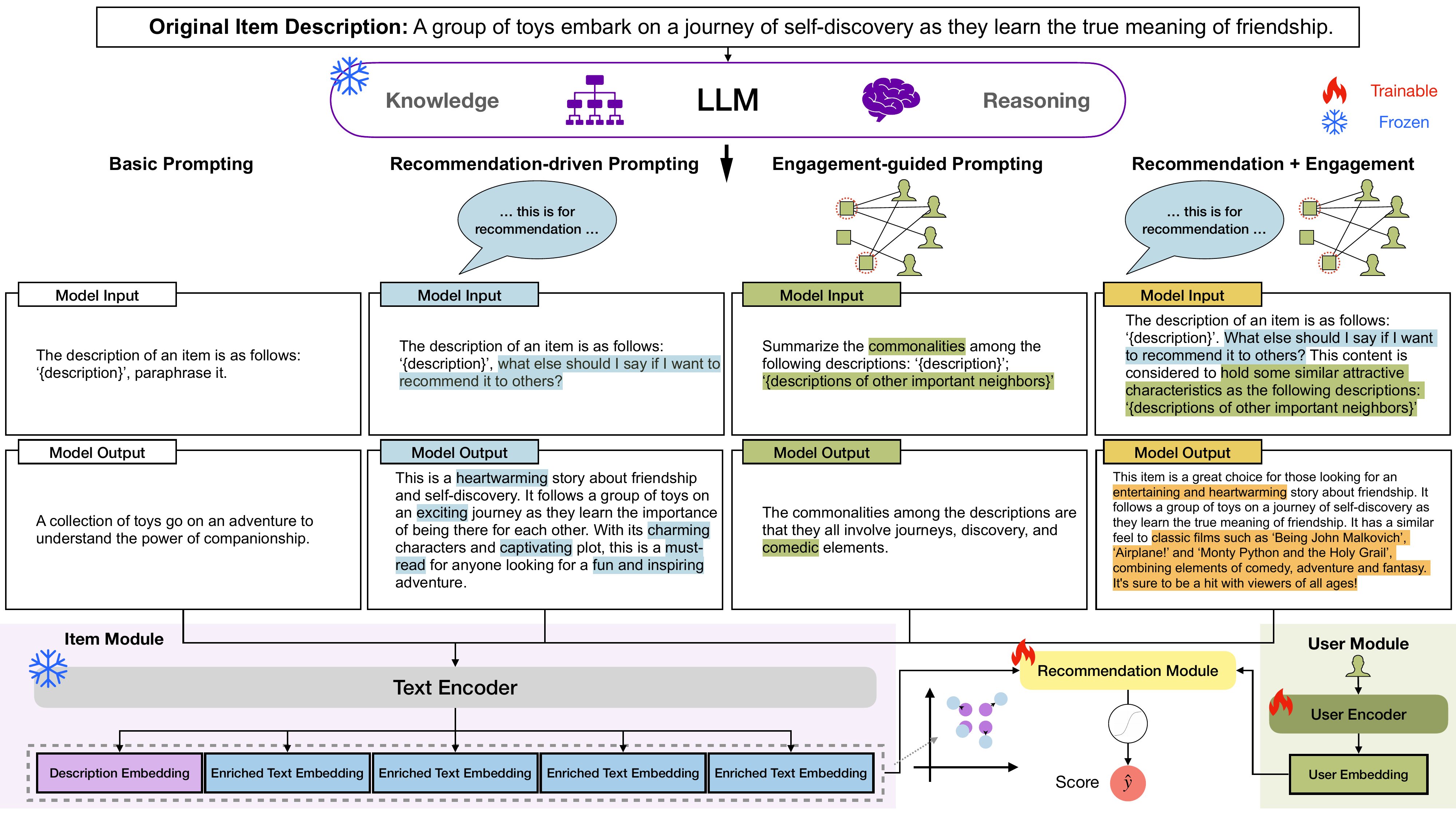}
    \caption{\model employs four prompting strategies to augment the original item descriptions which often contain incomplete information for recommendation. The augmented text is then concatenated to form the new input for the following recommendation module. \model plays a crucial role in enabling large language models to provide relevant context and help better align with user preferences. Prompts and augmented texts are highlighted.}
    \label{fig:llm-rec}
\end{figure*}
\noindent\textbf{LLM for Recommendation.}
Due to LLMs' remarkable text generation ability, many studies have leveraged LLMs as a data augmentation tool~\citep{dai2023auggpt, li2022elevater}. \citet{liu2023llava} used an LLM to produce multimodal language-image instruction-following datasets. Through a process of instruction tuning using this generated data, their proposed framework demonstrated an impressive aptitude in advancing vision and language comprehension. There have also been efforts to use LLMs to augment the input side of personalized recommendation. For instance, \citet{chen2023palr} incorporated user history behaviors, such as clicks, purchases, and ratings, into LLMs to generate user profiles. These profiles were then combined with the history interaction sequence and candidate items to construct the final recommendation prompt. LLMs were subsequently employed to predict the likelihood of user-item interaction based on this prompt. \citet{xi2023towards} introduced a method that leverages the reasoning knowledge of LLMs regarding user preferences and the factual knowledge of LLMs about items. However, our study focuses specifically on using LLMs' knowledge and reasoning ability to generate augmented input text that better captures the characteristics and nuances of items, leading to improved recommendation performance.

\section{LLM-Rec}

When composing a summary for recommendation purposes, it is customary to infuse it with specific emphases grounded in the author's \textit{comprehension} of the movie. This might involve accentuating the movie's distinctive attributes that set it apart from other movies. For instance, one may opt to incorporate genre information as a crucial element for classifying the movie. However, the decision to leverage the concept of genre for enhancing the summary is predicated on the author's understanding that the genre is a meaningful construct, effectively aligning the summary with the preferences and expectations of the intended audience. This paper aims to explore the potential of large language models when prompted to generate informative item descriptions and subsequently how to leverage this augmented text for enhancing personalized recommendations. Figure~\ref{fig:llm-rec} shows the diagram of \model. 
Specifically, our study focuses on investigating \textit{four} distinct LLM prompting strategies for description enrichment, namely basic prompting, recommendation-driven prompting, engagement-guided prompting, and the combination of recommendation-driven and engagement-guided prompting. The enriched text is then fed into the final recommendation module.

\noindent\textbf{Basic Prompting.} The concept of basic prompting closely resembles the task of crafting a general movie summary. Within this scope, we consider three basic prompting variants and refer to them as $p_{para}$, $p_{tag}$, and $p_{infer}$, respectively in the following experiments. $p_{para}$ instructs LLMs to paraphrase the original item description, emphasizing the objective of maintaining the same information without introducing any additional details. Given the original content description, the prompt we use is {\it ``The description of an item is as follows} `\{{\tt description}\}', {\it paraphrase it.''} $p_{tag}$ aims to guide LLMs to summarize the content description by using tags, striving to generate a more concise overview that captures key information. The corresponding prompt is  \textit{``The description of an item is as follows} `\{{\tt description}\}', {\it summarize it with tags.''} $p_{infer}$ instructs LLMs to deduce the characteristics of the original content description and provide a categorical response that operates at a broader, less detailed level of granularity. We use the following prompt in the experiments: \textit{``The description of an item is as follows} `\{{\tt description}\}', {\it what kind of emotions can it evoke?''} 

\noindent\textbf{Recommendation-driven Prompting.} This prompting strategy is to add a recommendation-driven instruction, into the basic prompting, resembling the task of creating a paragraph intended for making recommendations. We refer to the three recommendation-driven prompting as $p_{para}^{rec}$, $p_{tag}^{rec}$, and $p_{infer}^{rec}$, respectively in the following experiments, aligning with their counterparts in the basic prompting strategy. $p_{para}^{rec}$ represents the prompt: \textit{``The description of an item is as follows} `\{{\tt description}\}', {\it \purpletext{what else should I say if I want to recommend it to others?}''} The prompt for $p_{tag}^{rec}$ is \textit{``The description of an item is as follows} `\{{\tt description}\}', {\it what tags should I use \purpletext{if I want to recommend it to others?}''} The prompt for $p_{infer}^{rec}$ is \textit{``The description of an item is as follows} `\{{\tt description}\}', {\it \purpletext{recommend it to others} with a focus on the emotions it can evoke.''} 

\noindent\textbf{Engagement-guided Prompting.} As previously elucidated, the deficiency in item descriptions can also emanate from a limited comprehension of the user cohort for whom the recommendations are being generated. Typically, item descriptions are initially formulated for broad, general purposes, devoid of specific targeting toward particular user groups. As a result, they often fall short in capturing the intricate nuances of items required for a more fine-grained alignment with individual user preferences. The goal of the engagement-guided prompting strategy is to leverage user behavior, specifically the interactions between users and items (\ie user-item engagement) to devise prompts with the intention to steer LLMs towards a more precise comprehension of the attributes within the items, thus generating more insightful and contextually relevant descriptions that align more closely with the preferences of intended users. We refer to this variant as $p^{eng}$. To create the engagement-guided prompt, we combine the description of the target item, denoted as $d_{target}$, with the descriptions of $T$ \textbf{important} neighbor items, represented as \greentext{$d_{1}, d_{2}, \cdots, d_{T}$}. The importance is measured based on \textbf{user engagement}. More details can be found in Appendix~\ref{sec:importance}. The exact prompt of this prompting strategy is \textit{``\greentext{Summarize the commonalities among the following descriptions:}} ‘\{{\tt description}\}’; \greentext{‘\{{\tt descriptions of other important neighbors}\}’.}{\it''}

\noindent\textbf{Recommendation-driven + Engagement-guided Prompting.} It intends to incorporate both the recommendation-driven and engagement-guided instructions, which we denote as $p^{rec+eng}$: \textit{``The description of an item is as follows:} ‘\{{\tt description}\}’. {\it \purpletext{What else should I say if I want to recommend it to others?} This content is considered to \greentext{hold some similar attractive characteristics as the following descriptions:}} \greentext{‘\{{\tt descriptions of other important neighbors}\}’.}{\it''}   

\noindent\textbf{How does \mbox{\sc \textbf{LLM-Rec}}\xspace affect personalized recommendation?} 
In our experiments, we discover that first and foremost, \model stands out as a versatile yet simple framework, largely unrestricted by the type of items. Our experimental results on two datasets including the items that are categorically structured and extensively studied to items that are relatively novel and unclassified such as user-generated content, consistently demonstrate the substantial improvement in personalized recommendations. Simple models, such as MLP, can achieve performance on par with, or even better than, more advanced and complex models with the augmented text. This finding underscores the potential of simplified training to address challenges due to more complex models. More importantly, compared to other knowledge-based text augmentation methods, \model achieves superior recommendation performances and requires considerably less domain expertise compared to prior studies, making it much more accessible for implementation.
 
Second, \model contributes to increased recommendation transparency and explainability. The ability to directly investigate the augmented text not only enhances our understanding of the recommendation models but also offers insights into the characteristics of the items. It is invaluable for both users and system designers seeking to comprehend the rationale behind recommendations.

\section{Experiments}\label{sec:exp}

\begin{table*}[t]
    \centering
    \adjustbox{max width=\textwidth}{
    \begin{tabular}{lccccccc}
    \toprule[1.1pt]
   & & \multicolumn{3}{c}{\textbf{Movielens-1M}}   & \multicolumn{3}{c}{\textbf{Recipe}}  \\
  &    & Precision@10  & Recall@10      & NDCG@10       & Precision@10      & Recall@10    &NDCG@10         \\ \midrule
  \multicolumn{2}{l}{Item Popularity}  & 0.0426 \scriptsize{$\textcolor{gray}{\pm 0.0019}$}     & 0.0428   \scriptsize{$\textcolor{gray}{\pm 0.0028}$}       & 0.0530 \scriptsize{$\textcolor{gray}{\pm 0.0035}$}  & 0.0116  \scriptsize{$\textcolor{gray}{\pm 0.0025}$} &  0.0274 \scriptsize{$\textcolor{gray}{\pm 0.0083}$} &  0.0201 \scriptsize{$\textcolor{gray}{\pm 0.0053}$}  \\
  \multicolumn{2}{l}{MLP} & 0.2922 \scriptsize{$\textcolor{gray}{\pm 0.0019}$}& 0.2455 \scriptsize{$\textcolor{gray}{\pm 0.0031}$}& 0.3640 \scriptsize{$\textcolor{gray}{\pm 0.0039}$}& 0.0325 \scriptsize{$\textcolor{gray}{\pm 0.0021}$}& 0.0684 \scriptsize{$\textcolor{gray}{\pm 0.0066}$}& 0.0580 \scriptsize{$\textcolor{gray}{\pm 0.0054}$}\\
\multicolumn{2}{l}{AutoInt~\citep{song2019autoint}}                                                & 0.2149  \scriptsize{$\textcolor{gray}{\pm 0.0078}$} &  0.1706  \scriptsize{$\textcolor{gray}{\pm 0.0075}$}    & 0.2698 \scriptsize{$\textcolor{gray}{\pm 0.0092}$} & 0.0351 \scriptsize{$\textcolor{gray}{\pm 0.0032}$}  & 0.0772 \scriptsize{$\textcolor{gray}{\pm 0.0102}$}& \underline{0.0658} \scriptsize{$\textcolor{gray}{\pm 0.0089}$} \\
\multicolumn{2}{l}{DCN-V2~\citep{wang2021dcn}}   & 0.2961  \scriptsize{$\textcolor{gray}{\pm 0.0050}$}   & 0.2433 \scriptsize{$\textcolor{gray}{\pm 0.0057}$}   & 0.3689   \scriptsize{$\textcolor{gray}{\pm 0.0033}$}      & \underline{0.0360} \scriptsize{$\textcolor{gray}{\pm 0.0036}$} & \underline{0.0786} \scriptsize{$\textcolor{gray}{\pm 0.0104}$} & 0.0653  \scriptsize{$\textcolor{gray}{\pm 0.0085}$}\\
\multicolumn{2}{l}{EDCN~\citep{chen2021enhancing}} & 0.2935 \scriptsize{$\textcolor{gray}{\pm 0.0036}$} & 0.2392 \scriptsize{$\textcolor{gray}{\pm 0.0051}$} & 0.3678 \scriptsize{$\textcolor{gray}{\pm 0.0053}$} &0.0354 \scriptsize{$\textcolor{gray}{\pm 0.0030}$}& 0.0772 \scriptsize{$\textcolor{gray}{\pm 0.0091}$}& 0.0652 \scriptsize{$\textcolor{gray}{\pm 0.0071}$}\\
\midrule

\multirow{2}{*}{TagGPT~\cite{li2023taggpt}} &  {\sc Llama-2-7B}   & 0.2991 \scriptsize{$\textcolor{gray}{\pm 0.0017}$}  &0.2556 \scriptsize{$\textcolor{gray}{\pm 0.0038}$} & 0.3723  \scriptsize{$\textcolor{gray}{\pm 0.0023}$} & 0.0353  \scriptsize{$\textcolor{gray}{\pm 0.0024}$}& 0.0751 \scriptsize{$\textcolor{gray}{\pm 0.0067}$}&  0.0641 \scriptsize{$\textcolor{gray}{\pm 0.0057}$}\\ 
& {\sc GPT-3}   & 0.3001 \scriptsize{$\textcolor{gray}{\pm 0.0027}$} &0.2569 \scriptsize{$\textcolor{gray}{\pm 0.0028}$}& 0.3747 \scriptsize{$\textcolor{gray}{\pm 0.0042}$}& 0.0356  \scriptsize{$\textcolor{gray}{\pm 0.0032}$}& 0.0752 \scriptsize{$\textcolor{gray}{\pm 0.0084}$}&  0.0637 \scriptsize{$\textcolor{gray}{\pm 0.0068}$}\\ 

\multicolumn{2}{l}{KAR~\citep{xi2023towards}}   & 0.3056  \scriptsize{$\textcolor{gray}{\pm 0.0026}$}   & 0.2623 \scriptsize{$\textcolor{gray}{\pm 0.0034}$}   & 0.3824   \scriptsize{$\textcolor{gray}{\pm 0.0042}$}      & 0.0298 \scriptsize{$\textcolor{gray}{\pm 0.0018}$} & 0.0611 \scriptsize{$\textcolor{gray}{\pm 0.0049}$} & 0.0525  \scriptsize{$\textcolor{gray}{\pm 0.0043}$}\\

\multicolumn{2}{l}{\quad - augmented with ground truth}   & 0.3075  \scriptsize{$\textcolor{gray}{\pm 0.0015}$}   & 0.2636 \scriptsize{$\textcolor{gray}{\pm 0.0035}$}   & 0.3853   \scriptsize{$\textcolor{gray}{\pm 0.0027}$}      & - & - & - \\

\midrule

\multirow{4}{*}{\model} &  \multirow{2}{*}{{\sc Llama-2-7B}}   & \underline{0.3102} \scriptsize{$\textcolor{gray}{\pm 0.0014}$}  &\underline{0.2712}  \scriptsize{$\textcolor{gray}{\pm 0.0026}$} & \underline{0.3867}  \scriptsize{$\textcolor{gray}{\pm 0.0027}$} & 0.0359  \scriptsize{$\textcolor{gray}{\pm 0.0024}$}& 0.0770 \scriptsize{$\textcolor{gray}{\pm 0.0076}$}&  0.0632 \scriptsize{$\textcolor{gray}{\pm 0.0052}$}\\ 
& & (\greentext{+6.16$\%$}) &  (\greentext{+10.47$\%$})  &  (\greentext{+6.24$\%$})&  (\greentext{+10.46$\%$}) &  (\greentext{+12.57$\%$}) &  (\greentext{+8.97$\%$})\\
& \multirow{2}{*}{{\sc GPT-3}}    & \textbf{0.3150} \scriptsize{$\textcolor{gray}{\pm 0.0023}$} &\textbf{0.2766} \scriptsize{$\textcolor{gray}{\pm 0.0030}$}& \textbf{0.3951} \scriptsize{$\textcolor{gray}{\pm 0.0035}$}& \textbf{0.0394}  \scriptsize{$\textcolor{gray}{\pm 0.0033}$}& \textbf{0.0842} \scriptsize{$\textcolor{gray}{\pm 0.0098}$}&  \textbf{0.0706} \scriptsize{$\textcolor{gray}{\pm 0.0084}$}\\ 

& & (\greentext{+7.80$\%$}) &  (\greentext{+12.67$\%$})  &  (\greentext{+8.54$\%$})&  (\greentext{+21.23$\%$}) &  (\greentext{+23.10$\%$}) &  (\greentext{+21.72$\%$})\\
             
             \bottomrule[1.1pt]
\end{tabular}}
\caption{Average recommendation performance between \model and baseline approaches across five different train/test splits. The best results are highlighted in \textbf{bold}, the second-best results are \underline{underlined}, and relative gains compared to the MLP baseline are indicated in \greentext{green}.}
    \label{tab:sota_comp}
\end{table*}
\subsection{Experiment Setup}\label{sec:exp_setup}

\noindent\textbf{Datasets and Baslines.} Two widely adopted recommendation benchmarks are used, Movielens-1M~\citep{harper2015movielens} for movie recommendation, and Recipe~\citep{majumder2019generating} for recipe recommendation.  
To assess \model's efficacy, we compare it against two distinct categories of baselines. The first category includes content-based baselines that takes solely the original item descriptions as input. The second category includes different text augmentation methods.  Details including dataset statistics, preprocessing specifics, baselines, model training, hyper-parameter settings and implementation are discussed in Appendix~\ref{sec:detailed_exp_setting}. 

\noindent\textbf{Language Models.} Two large language models are selected for experiments. The first is {\sc GPT-3}~\citep{brown2020language}, particularly its variant {\tt text-davinci-003}. This variant is an advancement over the InstructGPT models~\citep{ouyang2022training}. We select this variant due to its ability to consistently generate high-quality writing, effectively handle complex instructions, and demonstrate enhanced proficiency in generating longer form content~\citep{raf2023davinci}. The second is {\sc Llama-2}~\citep{touvron2023llama2}, which is an open-sourced model that has shown superior performance across various external benchmarks in reasoning, coding, proficiency, and knowledge tests. Specifically, we use the {\sc Llama-2-Chat} variant of 7B parameters.

\noindent\textbf{Evaluation Protocols.} 
We follow the same evaluation methodology of \citet{wei2019mmgcn}. We randomly divide the dataset into training, validation, and test sets using an 8:1:1 ratio. Negative training samples are created by pairing users and items without any recorded interactions (note that these are pseudo-negative samples). For the validation and test sets, we pair each observed user-item interaction with $n$ items that the user has not previously interacted with. Here we follow the methodology outlined in the previous study~\cite{wei2019mmgcn} and set $n$ to $1,000$. It is important to note that there is \textit{no} overlap between the negative samples in the training set and the unobserved user-item pairs in the validation and test sets. This ensures the independence of the evaluation data. We use metrics such as Precision@K, Recall@K and NDCG@K to evaluate the performance of top-K recommendations, where $K$ is set to $10$. We report the average scores across five different splits of the testing sets. The recommendation module of \model is the combination of an MLP model and a dot product.

\subsection{Main Results}\label{sec:main_results}

\textbf{Integrating the text augmented by \mbox{\sc \textbf{LLM-Rec}}\xspace boosts recommendation performance.} Table~\ref{tab:sota_comp} shows the average recommendation performance between \model and baseline approaches across five different splits. In addition to the aforementioned baselines, we include another baseline relies only on item popularity and does not involve any learning process, referred to as Item Popularity. Note that \model uses the same recommendation module as the MLP baseline. It can be seen that \model exhibits significant relative gains over the MLP method. For instance, \model achieves improvements in NDCG@10 from 6.24\% to 8.54\% in Movielens-1M and from 8.97\% to 21.72\% in Recipe. These improvements highlight the value of the augmented text in the recommendation process. More importantly, when comparing \model with the other three content-based methods, we observe that \model empowers simple MLP models to attain comparable or even superior performance, surpassing other more complex feature-based recommendation methods.

\begin{figure*}[t]
    \centering
    \includegraphics[width=\textwidth]{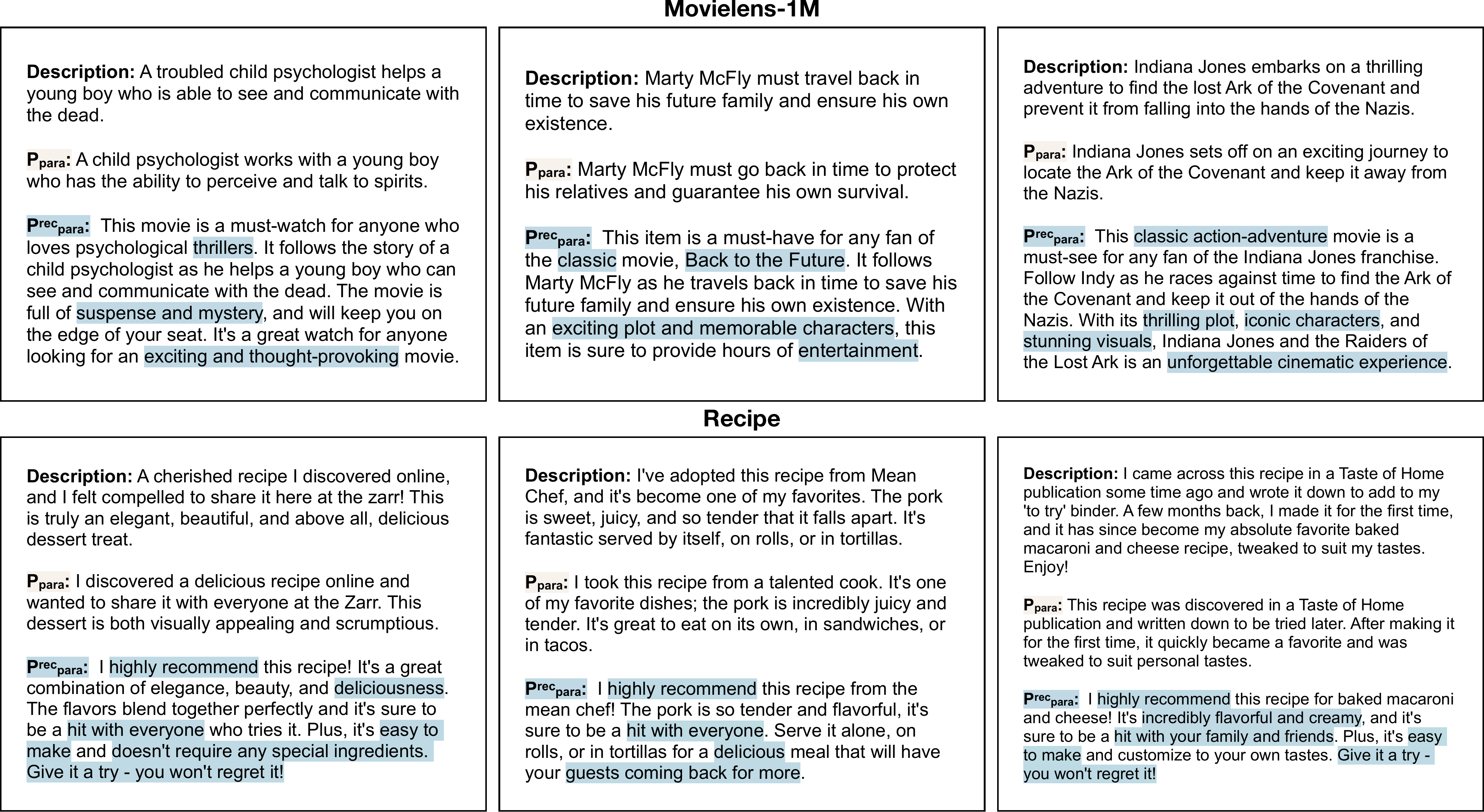}
    \caption{Example responses generated by {\sc GPT-3}. The additional information augmented via the recommendation-driven prompting is highlighted in \purpletext{blue}. Examples generated by {\sc Llama-2} can be found in Table~\ref{tab:example_llama2_para}.}
    \label{fig:case_para}
\end{figure*}

\noindent\textbf{What extra information does \mbox{\sc \textbf{LLM-Rec}}\xspace incorporate that contributes to its performance improvement?}
To better understand why \model is effective, particularly regarding the additional information contributed by its prompting approach, we conduct both qualitative and quantitative studies. We find that the augmented content contains more detailed and expressive descriptions, emphasizing item characteristics which helps in understanding items more comprehensively than with their original descriptions and contributing to the performance improvement. Figure~\ref{fig:case_para} shows example responses generated by {\sc GPT-3} with $p_{para}$ and $p_{para}^{rec}$. The first example suggests that the response via $p_{para}^{rec}$ categorizes the movie as a \purpletext{psychological thriller} and recommends it as a \purpletext{must-watch} for fans of this genre. It also positions the movie as both \purpletext{exciting} and \purpletext{thought-provoking}, appealing to those looking for more than just \purpletext{entertainment}. These distinctive words describe user preferences and item characteristics including genre description, descriptive elements, and viewer recommendation. While Figure~\ref{fig:case_para} might suggest that the LLM-generated text for the Recipe dataset adds only modifiers, these phrases, like ``easy to make,'' actually reflect key characteristics valued in the Recipe dataset, such as simplicity. Some authors may also add \#easytomake to their recipe descriptions~\cite{majumder2019generating}. Consistent patterns are also observed when comparing the responses of $p_{tag}$ with $p_{tag}^{rec}$ (Tables~\ref{tab:example_gpt_tag} and \ref{tab:example_llama2_tag}), and $p_{infer}$ with $p_{infer}^{rec}$ (Tables~\ref{tab:example_gpt_infer} and \ref{tab:example_llama2_infer}). A more thorough analysis shows that \model can be applied to diverse item domains and it is not restricted to datasets with rich textual information. Please see Appendix~\ref{appendix:applicability}.

We hypothesize that these generated words contribute to improving recommendation performance. To further validate this hypothesis, we design two variants of the response generated by {\sc GPT-3}, namely $p_{para}^{mask}$ and $p_{para}^{keyword}$. To construct $p_{para}^{mask}$, we mask the words that appear in the response of $p_{para}^{rec}$ but are absent in the response of $p_{para}$. To construct $p_{para}^{keyword}$, we append the words that (1) appear in the response of $p_{para}^{rec}$ and (2) are pre-defined user-preference-related words such as genres to the end of the response of $p_{para}$ (more details in Appendix~\ref{appendix:keyword_construction}). These two variants are then fed into MLP models to form baselines. Comparing the performance of $p_{para}^{rec}$ and $p_{para}^{mask}$ in Table~\ref{tab:mask_keyword}, we observe a discernible decline in recommendation performance when words unique to the response of $p_{para}^{rec}$ are selectively masked. This outcome highlights the pivotal role played by the supplementary insights introduced through the augmented text. Further, our investigation reveals that the incorporation of vital keywords, as opposed to the inclusion of all response words, can yield even superior recommendation performance. This may be attributed to potential discrepancies or extraneous elements within the response of $p_{para}^{rec}$.

\begin{table}[t]
  \centering
  \begin{adjustbox}{width=0.8\columnwidth,center}
  \begin{tabular}{lcc} 
    \toprule[1.1pt]
                & \textbf{Movielens-1M} & \textbf{Recipe}\\
                \midrule
$p_{para}$      & 0.3746   \small{$\textcolor{gray}{\pm 0.0028}$}                 & 0.0611 \small{$\textcolor{gray}{\pm 0.0053}$}       \\
$p_{para}^{keyword}$   &   \begin{tabular}[c]{@{}c@{}}0.3822 \small{$\textcolor{gray}{\pm 0.0049}$}\\(\greentext{\textbf{+2.03\%}})\end{tabular}          &  \begin{tabular}[c]{@{}c@{}}0.0615 \small{$\textcolor{gray}{\pm 0.0060}$}\\ (\greentext{\textbf{+0.65\%}})\end{tabular}        \\
\midrule
$p_{para}^{rec}$ & 0.3777  \small{$\textcolor{gray}{\pm 0.0028}$}               & 0.0646   \small{$\textcolor{gray}{\pm 0.0072}$}         \\
$p_{para}^{mask}$      &              \begin{tabular}[c]{@{}c@{}}0.3769 \small{$\textcolor{gray}{\pm 0.0040}$}\\(\redtext{\textbf{-0.21\%}})\end{tabular}    &         \begin{tabular}[c]{@{}c@{}}0.0611 \small{$\textcolor{gray}{\pm 0.0066}$}\\(\redtext{\textbf{-0.52\%}})\end{tabular}     \\
    \bottomrule[1.1pt]
  \end{tabular}
  \end{adjustbox}
   \caption{Average NDCG@10 across five splits.}
  \label{tab:mask_keyword}
\end{table}

\noindent\textbf{\mbox{\sc \textbf{LLM-Rec}}\xspace augmentation outperforms other text augmentation methods for recommendation.} We compare \model with two recent advancements in the field of using LLMs to augment item information, specifically Knowledge Augmented Recommendation (KAR) as proposed by \citet{xi2023towards}, and TagGPT, as proposed by \citet{li2023taggpt}. KAR introduces a fusion of domain knowledge and prompt engineering to generate factual knowledge pertaining to the items (for detailed implementation information, see Appendix~\ref{appendix:implementation}).  Since the augmented information may not necessarily be correct, we further implement a variant with ground truth knowledge. It aligns with strategies akin to those introduced by \citet{di2012linked}, who harnessed external databases to enhance item information. In a manner consistent with this approach, we incorporate genre information into the item descriptions. Note that genre constitutes one of the metadata components in Movielens-1M. Such categorical characteristics are absent in Recipe. Therefore, we only apply this variant to the Movielens-1M dataset.

\begin{figure}[t]
    \centering
    \includegraphics[width=0.98\linewidth]{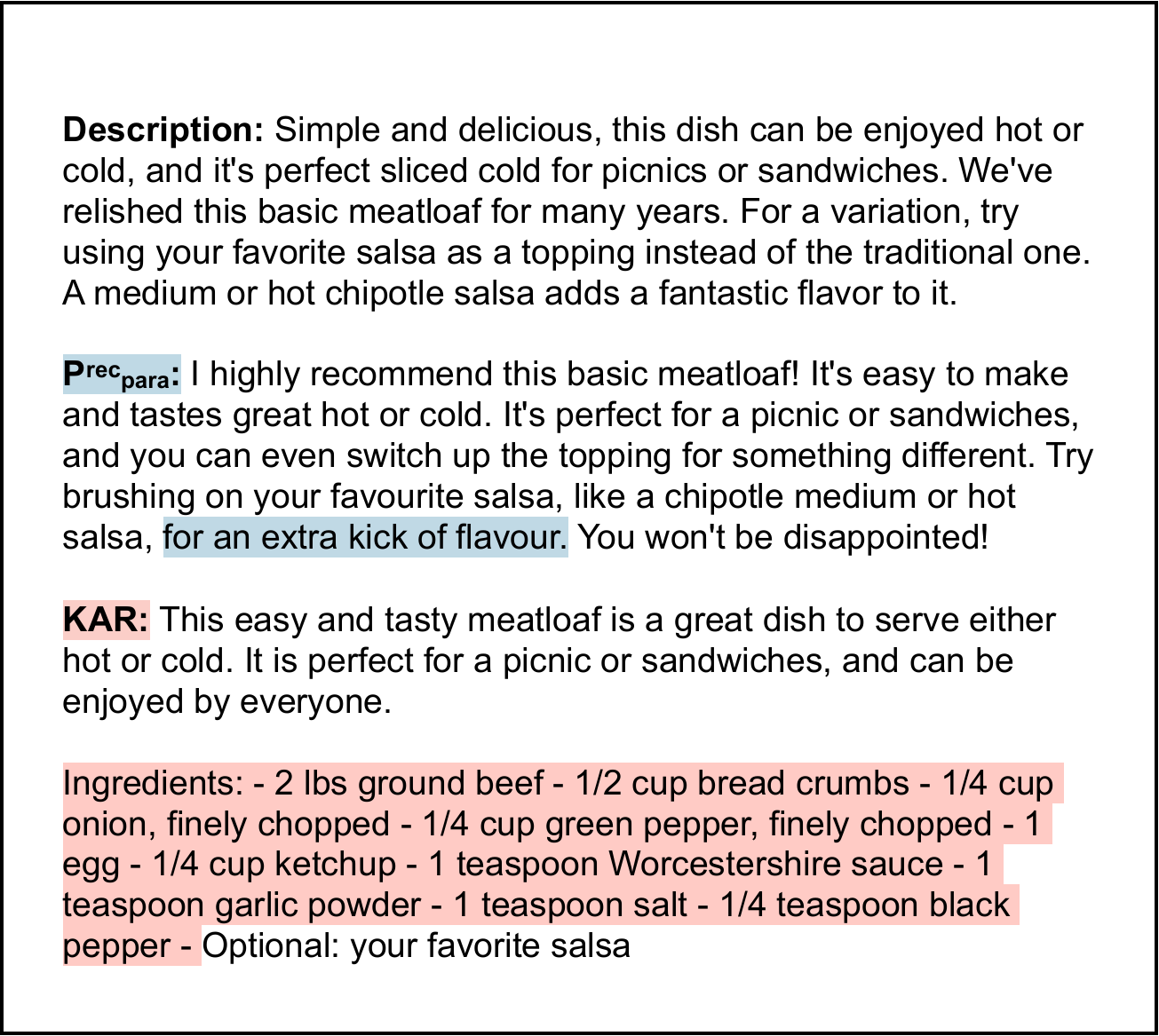}
    \caption{An example response generated using knowledge augmentation prompts~\cite{xi2023towards}. The additional information is highlighted in \redtext{red}.}
    \label{fig:kar_recipe_example_short}
\end{figure}

As shown in Table~\ref{tab:sota_comp}, the incorporation of knowledge-based text augmentation offers significant improvements in recommendation performance for well-classified items, such as movies. However, it becomes evident that this approach faces limitations when applied to items, like user-generated content, that are inherently more novel and dynamic in nature. The example response in Figure~\ref{fig:kar_recipe_example_short} shows that one key reason that knowledge augmentation approaches do not yield optimal improvement regarding recommendation performance may lie in the potential \textbf{mismatch} between the generated knowledge and the target item. For instance, while the generated ingredient information may be correct for most meatloaf recipes, it could be entirely wrong for a specific recipe without additional context. In contrast to these knowledge augmentation methods, \model's recommendation-driven prompts provide augmented information that describes the target item at a broader, less granular level, especially when compared to KAR. More importantly, \model does not require domain knowledge throughout the entire process.

In terms of the second text augmentation baseline, TagGPT~\cite{li2023taggpt}, which extracts tags using LLMs, several key observations can be made. First, we note an improvement in recommendation performance using tag generation compared to the baseline methods. Second, the prompts specifically designed within our \model framework demonstrate superior effectiveness compared to those used in TagGPT.

\subsection{Ablation Study}\label{sec:ablation_study}

\textbf{How does each prompting strategy perform?} We conduct an ablation study to examine the impact on recommendation performance when models use either original item descriptions alone or a combination of these descriptions with augmented text derived from one of four distinct prompting strategies. The results, presented in Table~\ref{tab:ablation_different_prompt_short}, reveal a noteworthy and consistent enhancement in recommendation performance across various prompting strategies within two benchmark datasets.

We also note variations in the performance of these strategies across different domains, aligning with our expectations. In Movielens-1M, the strategy combining recommendation-driven and engagement-guided approaches yields the best results. Conversely, in Recipe, the recommendation-driven strategy alone proves most effective. This variability suggests that combining multiple objectives in a single prompting strategy does not always lead to superior performance. When LLMs are tasked with generating descriptions serving multiple purposes, the balance of information becomes crucial. If neighboring item descriptions vary widely, it can challenge the model's ability to generate useful content, potentially leading to less optimal improvements. To address this, \model integrates all enriched text and leverages the subsequent recommendation module to effectively model the extra information. Future work can investigate different prompt designs, aiming to effectively achieve multiple objectives simultaneously.

\begin{table}[t]
\centering
\begin{adjustbox}{width=0.8\columnwidth,center}
\begin{tabular}{lcc}
\toprule[1.1pt]
        & \textbf{Movielens-1M} & \textbf{Recipe}         \\
\midrule
 \begin{tabular}[l]{@{}l@{}}Item\\Description\end{tabular}&  0.3640 \small{$\textcolor{gray}{\pm 0.0039}$}& 0.0580 \small{$\textcolor{gray}{\pm 0.0054}$} \\\midrule

 Basic   & \begin{tabular}[c]{@{}c@{}}0.3747 \small{$\textcolor{gray}{\pm 0.0042}$}\\(\greentext{\textbf{+2.94\%}})\end{tabular} &   \begin{tabular}[c]{@{}c@{}}0.0644 \small{$\textcolor{gray}{\pm 0.0068}$}\\(\greentext{\textbf{+11.03\%}})\end{tabular}\\
 Rec     & \begin{tabular}[c]{@{}c@{}}0.3786 \small{$\textcolor{gray}{\pm 0.0041}$}\\(\greentext{\textbf{+4.01\%}})\end{tabular} &  \begin{tabular}[c]{@{}c@{}}\textbf{0.0649} \small{$\textcolor{gray}{\pm 0.0069}$}\\(\greentext{\textbf{+11.90\%}})\end{tabular}\\
            Eng     & \begin{tabular}[c]{@{}c@{}}0.3801 \small{$\textcolor{gray}{\pm 0.0032}$}\\(\greentext{\textbf{+4.42\%}})\end{tabular} & \begin{tabular}[c]{@{}c@{}}0.0628 \small{$\textcolor{gray}{\pm 0.0077}$}\\(\greentext{\textbf{+8.28\%}})\end{tabular}\\
        Rec+Eng &   \begin{tabular}[c]{@{}c@{}}\textbf{0.3802} \small{$\textcolor{gray}{\pm 0.0037}$}\\(\greentext{\textbf{+4.45\%}})\end{tabular}&  \begin{tabular}[c]{@{}c@{}}0.0625 \small{$\textcolor{gray}{\pm 0.0060}$}\\(\greentext{\textbf{+7.76\%}})\end{tabular}\\
                        \bottomrule[1.1pt]       
\end{tabular}
\end{adjustbox}
\caption{Average NDCG@10 across five different splits among different prompting strategies. }
    \label{tab:ablation_different_prompt_short}

\end{table}

\begin{figure}[t]
 \centering

 \includegraphics[width=0.9\linewidth]{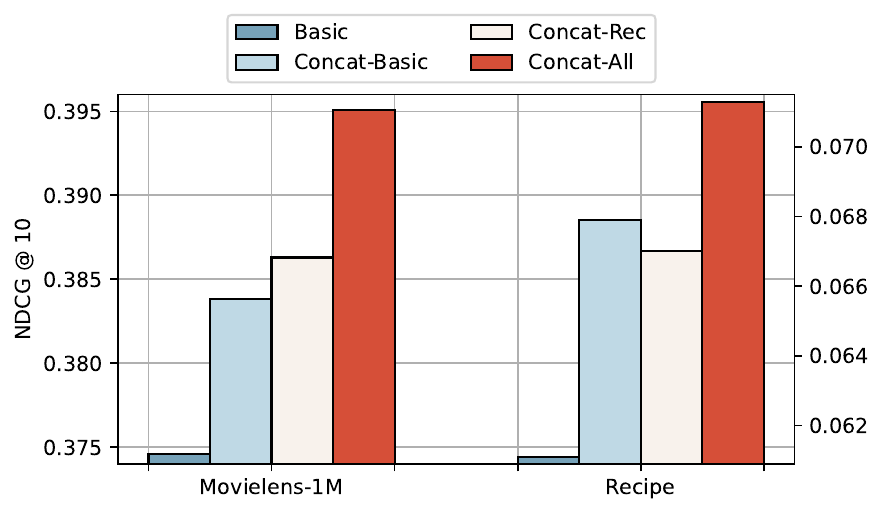}
  \caption{The ablation study shows that recommendation benefits from concatenating the embeddings of the input text augmented by LLM.  
  }\label{fig:concat}
\end{figure}

\noindent\textbf{How does concatenating the augmented responses affect recommendation?} In Table~\ref{tab:sota_comp}, we show that the MLP model, which combines all augmented text with the original description, outperforms more advanced models that rely solely on the original description as input. Now we take a deeper look at the quality of the combined augmented text. We employ the same recommendation module (\ie an MLP with a dot product) and evaluate the recommendation performance of various concatenation combinations. As shown in Figure~\ref{fig:concat}, the model denoted as {\tt Basic} uses the embeddings of text augmented through $p_{para}$. {\tt Concat-Basic} represents the model that concatenates the embeddings of the input text augmented by all basic prompts. Additionally, {\tt Concat-Rec} is the model that employs the concatenation of the embeddings of input text augmented by all recommendation-driven prompts. Lastly, {\tt Concat-All} stands for the model that combines the embeddings of input text augmented by all four prompting strategies. Our findings reveal that concatenating more information \textit{consistently} enhances recommendation performance. This emphasizes the added value of incorporating augmented text as opposed to relying solely on the original content description. Complete results of Figure~\ref{fig:concat} can be found in Figure~\ref{fig:ablation_different_concat}.

\noindent\textbf{How to effectively integrate the augmented responses to maximize improvement?} Table~\ref{tab:concat_short} shows the recommendation performances of other concatenation variants: (1) \textbf{Duplication}: We duplicate the embeddings of the original item description to match the dimension size of the embeddings of {\tt Concat-All}; (2) \textbf{Text concatenation}: Instead of concatenating the embeddings of all response (\ie {\tt Concat-All}), we concatenate the responses first, and then convert it to embeddings. Through a comparative analysis of the model's performance, contrasting the first variant with {\tt Concat-All}, it becomes evident that the observed improvement in performance is not attributable to changes in embedding size. Further, by comparing the performance of the second variant against {\tt Concat-All}, we discover that in scenarios where the text encoder remains unmodified, the most effective strategy to integrate all enriched information is by first converting the text into embeddings and then concatenating these embeddings. This approach surpasses the method of concatenating text prior to its conversion into embeddings. Future research can explore the potential of modifying the text encoder to further enhance model efficiency and effectiveness.

\begin{table}[t]
    \centering
    \begin{adjustbox}{width=0.9\columnwidth,center}
    \begin{tabular}{lcc}
    \toprule[1.1pt]
    & \textbf{Movielens-1M}   & \textbf{Recipe}  \\ 
        \midrule
      Item Description & 0.3640 \small{$\textcolor{gray}{\pm 0.0039}$} & 0.0580 \small{$\textcolor{gray}{\pm 0.0054}$}\\
      \midrule
  Duplication & 0.3567 \small{$\textcolor{gray}{\pm 0.0043}$}& 0.0590 \small{$\textcolor{gray}{\pm 0.0068}$}\\
  Text Concatenation & 0.3853 \small{$\textcolor{gray}{\pm 0.0027}$} & 0.0591 \small{$\textcolor{gray}{\pm 0.0065}$} \\

             {\tt Concat-All} (ours)     &  \textbf{0.3951}  \small{$\textcolor{gray}{\pm 0.0035}$}&  \textbf{0.0706}  \small{$\textcolor{gray}{\pm 0.0084}$}\\ 
             \bottomrule[1.1pt]
\end{tabular}
\end{adjustbox}
\caption{Average NDCG@10 across five splits among different methods of extra information integration.}
    \label{tab:concat_short}
\end{table}

\noindent\textbf{Does modifying the word choice in the designed prompts significantly affect the augmented output?} To investigate this, we construct one variant prompt for each of \model's prompts, ensuring they convey the same meaning but with different word choices.  As shown in Table~\ref{tab:appendix_variant_quantitative}, despite variations in wording of the prompts, the responses remain largely similar. The observed lower cosine similarity is primarily attributed to differences in the format of the responses which can be mitigated through various strategies, such as additional fine-tuning of the model or incorporating specific instructions within the prompts.

\begin{table}[t]
  \centering
  \begin{adjustbox}{width=0.65\columnwidth,center}
\begin{tabular}{lll}
\toprule[1.1pt]
\multicolumn{2}{l}{Prompting Strategy}    & Variant      \\\midrule
\multirow{3}{*}{Basic} & $p_{para}$      & 0.8859 \scriptsize{$\textcolor{gray}{\pm 0.0898}$} \\
& $p_{tag}$         & 0.9112 \scriptsize{$\textcolor{gray}{\pm 0.1000}$} \\
& $p_{infer}$     & 0.6819 \scriptsize{$\textcolor{gray}{\pm 0.1500}$} \\\midrule
\multirow{3}{*}{Rec} & $p^{rec}_{para}$   & 0.7011 \scriptsize{$\textcolor{gray}{\pm 0.1369}$} \\
& $p^{rec}_{tag}$   &0.8627 \scriptsize{$\textcolor{gray}{\pm 0.1248}$} \\
& $p^{rec}_{infer}$ & 0.8458 \scriptsize{$\textcolor{gray}{\pm 0.0652}$}  \\\midrule
Eng & $p^{eng}$        & 0.6218 \scriptsize{$\textcolor{gray}{\pm 0.1012}$} \\\midrule
Rec+Eng & $p^{rec+eng}$    &  0.8542 \scriptsize{$\textcolor{gray}{\pm 0.0802}$} \\
\bottomrule[1.1pt]
\end{tabular}
\end{adjustbox}
\caption{Average cosine similarity between the variant prompt responses and those generated from \model.}
  \label{tab:appendix_variant_quantitative}
\end{table}

\section{Discussions and Conclusions}\label{sec:discussions_and_conclusions}

In this study, we have investigated the effectiveness of \model as a simple yet impactful mechanism for improving recommendation through LLMs. Our approach is among those early attempts~\cite{lin2023can, chen2023large} that leverage LLMs for text augmentation in recommendation. There are three key contributions that distinguish our work from the concurrent ones. 
First, while previous work, such as KAR~\cite{xi2023towards}, focuses on design augmentation algorithm for a specific recommendation model, our model focuses on input text augmentation with LLMs, which is suitable for any content-based backbone recommendation models, demonstrating the flexibility of our approach. 
Second, in addition to our recommendation-driven augmentation using LLMs, we also design engagement-guided prompts to augment the input, which contains more personalized item characteristics. 
Overall, we conduct comprehensive experiments, with different combinations of prompting strategies, to not only illustrate the superior performance of our approach but also uncover the underlying rationale of the improvements.

We introduce \model, which enhances recommendation by augmenting the original item descriptions which often contains incomplete information for effective recommendations using large language models. 
We observed from extensive experiments that combining augmented input text and original item descriptions yields notable improvements in recommendation quality. These findings show the potential of using LLMs and strategic prompting techniques to enhance the accuracy and relevance of recommendation with an easier training process. By incorporating additional context, we enable the recommendation algorithms to capture more nuanced information and generate recommendations that better align with user preferences.

\section*{Limitations}\label{sec:limitations}
In this study, we have investigated the effectiveness of \model as a simple yet effective mechanism for improving recommendation through large language models. While effective, \model does have its limitations. First, there is extra computational cost associated with \model framework. The primary computational load comes from the augmentation phase including the augmented output text length. Our findings indicate that selecting important words for inclusion, rather than incorporating all response words, can lead to improved recommendation performance, as evidenced in Table~\ref{tab:mask_keyword}. Our future work will explore the balance between the number of words generated and the resulting performance enhancements. Second, similar to many LLM-based studies, \model faces challenges in promptly incorporating the latest knowledge. We plan to investigate methods in future work for LLMs to autonomously gather and summarize current knowledge from external sources, thereby improving text augmentation effectiveness.

\section*{Acknowledgments}
Luo was supported in part by NSF Award \#2238208.

\bibliography{custom}

\clearpage
\appendix

\section{Additional Details of Experiment Setup}\label{sec:detailed_exp_setting}

\subsection{Datasets}\label{sec:appendix_dataset}

\begin{table}[h]
  \centering
  \adjustbox{max width=0.8\linewidth}{
  \begin{tabular}{llll} 
    \toprule[1.1pt]
    Dataset      & \# Interaction & \# Item & \# User \\
\midrule
MovieLens-1M & 1,000,209      & 3,706   & 6,040   \\
Recipe       & 132,246        & 4,125   & 2,589  \\
    \bottomrule[1.1pt]
  \end{tabular}
  }
  \caption{Benchmark Statistics.}
  \label{tab:ablation_position}
\end{table}

\textbf{MovieLens-1M}~\citep{harper2015movielens} is a highly recognized benchmark dataset commonly used for evaluating item recommendation systems.\footnote{License: \url{https://files.grouplens.org/datasets/movielens/ml-1m-README.txt}} It contains a vast collection of 1,000,209 ratings provided by 6,040 MovieLens users, covering 3,900 movies. Each user has at least 20 ratings. Following  \citet{he2017neural}, we convert the rating data into implicit feedback. More specifically, each entry is marked as 0 or 1 indicating whether the user has rated the corresponding item. The original movie data only contain movie titles and genres. We employ {\sc GPT-3} ({\tt text-davinci-003}) to generate the content description of each movie using the following prompt: {\it ``Summarize the movie} \{{\tt title}\} {\it with one sentence. The answer cannot include the movie title.''} The response from {\sc GPT-3} is used as the item description. {\tt Temperature} is set at 0 to generate more focused and deterministic responses. Note that inclusion of the movie title is entirely \textit{optional}. We opt not to include the title intentionally, as our design for \model emphasizes its role as a general prompting framework. This versatility is important, as it is intended to function across a wide array of item types, including those that may not possess pre-defined titles, such as short videos. 

\textbf{Recipe}~\citep{majumder2019generating} is another benchmark dataset we use to assess the recommendation performance. This dataset consists of recipe details and reviews sourced from \url{Food.com}.\footnote{License: \url{https://www.kaggle.com/datasets/shuyangli94/food-com-recipes-and-user-interactions}} The metadata includes ratings, reviews, recipe names, descriptions, ingredients, directions, and so on. For instance, an example recipe description is \textit{``All the delicious flavors of mac n' cheese transformed into a warm, comforting bowl of soup!''}. In our evaluation, we employ the recipe descriptions as item descriptions for the four prompting strategies. Similar to the MovieLens-1M dataset, we apply filtering criteria, excluding users with fewer than 20 ratings and items with fewer than 30 ratings. Note that all original descriptions presented as examples in this paper have been paraphrased to protect user privacy.

The selection of these benchmarks is mainly motivated by two factors. First, we select movies and recipes as they represent two distinct types of content. Movies, being more categorically organized and widely researched, contrast sharply with recipes, which are diverse, user-generated content from social media platforms, often lacking a strict categorical structure and presenting more novelty. Second, the nature of their descriptions differs significantly: movie descriptions typically comprise narrative summaries, whereas recipe descriptions are instructional. Evaluating our model on these varied datasets allows for a comprehensive analysis of how different prompting strategies affect recommendation outcomes, providing valuable insights into their effectiveness across diverse content types.

\subsection{Baselines}\label{sec:baselines}
To assess \model's efficacy, we compare it against two distinct categories of baselines. The first category includes baselines that takes solely the original item descriptions as input. This includes models from MLP to more complex content-based approaches. Specifically, we choose three more advanced, content-based recommendation models. AutoInt is a multi-head self-attentive neural network with residual connections designed to explicitly model feature interactions within a low-dimensional space~\citep{song2019autoint}. DCN-V2 represents an enhanced version of DCN and incorporates feature crossing at each layer~\citep{wang2021dcn,wang2017deep}. Lastly, EDCN~\citep{chen2021enhancing} introduces a bridge module and a regulation module to collaboratively capture layer-wise interactive signals and learn discriminative feature distributions for each hidden layer in parallel networks, such as DCN. The purpose of this comparison is to evaluate the added value of augmented text in improving recommendation outcomes. 

The second category includes different text augmentation methodologies. Here, \model is evaluated against two recent advancements in the field of using LLMs to augment item information. The first method is Knowledge Augmented Recommendation (KAR) as proposed by \citet{xi2023towards}. KAR introduces a fusion of domain knowledge and prompt engineering to generate factual knowledge pertaining to the items (for implementation details, see Appendix~\ref{appendix:kar_details}) In contrast to KAR's approach, \model places a particular emphasis on the innate common-sense reasoning capabilities of large language models and notably does not mandate domain expertise. Since the augmented information may not necessarily be correct, we further implement a variant with ground truth knowledge. It aligns with strategies akin to those introduced by \citet{di2012linked}, who harnessed external databases to enhance item information. The second method, TagGPT, proposed by \citet{li2023taggpt}, extracts tags using LLMs, similar to one of our prompting strategies for item descriptions.

Although collaborative Filtering (CF) is another widely used technique in recommendation systems, given our primary focus on addressing the issue of incomplete item descriptions, we do not conduct experiments under CF settings~\citep{li2023exploring}. Instead, we concentrate on comparing our method with other input augmentation approaches.

\subsection{Item and User Modules}\label{sec:item_and_user_module}
We use Sentence-BERT~\citep{reimers2019sentence} to derive the textual embeddings from the original item description and augmented text. The embedding model is {\tt all-MiniLM-L6-v2}. 
We directly apply it to convert the natural language to embeddings without fine-tuning it, freezing only the text-encoder and not the user encoder to avoid the additional computational cost associated with training and fine-tuning the sentence transformer model.
For users, we employ an embedding table to convert user ID into latent representations. For both MovieLens-1M and Recipe datasets, the output dimension is set at 128. 
We have considered using embeddings derived from the LLMs. However, as our goal is to propose a general framework that can leverage both open-source and proprietary models, we do not pursue further exploration of this aspect in the current study.

\subsection{Model Training}\label{sec:model_training}
To facilitate the model training process, we employ the binary cross-entropy loss, expressed as:
\begin{align}
\begin{split}
L= & -\sum_{(u, i) \in Y} [y_{u,i}\cdot \log \hat{y}_{u, i} + \\ & (1 - y_{u, i}) \cdot \log (1-\hat{y}_{u, i})]
\end{split}
\end{align}
where $(u, i)$ represents the user-item pair, and $Y$ denotes the set that contains all positive and negative samples. The variable $y_{u,i}$ serves as a label, with a value of 1 indicating that user $u$ has engaged with item $i$, and 0 representing the absence of interaction. The prediction score $\hat{y}_{u,i}$, ranging from 0 to 1, reflects the likelihood of user $u$ interacting with item $i$. In our dataset, each instance of user-item interaction is considered a positive sample. Alongside these positive samples, we incorporate negative samples by randomly pairing users with items that lack any prior recorded interactions. To mitigate the risk of overfitting and enhance training efficiency, we implement an early stopping mechanism. The window size and evaluation frequency are both configured to be 5. It is noteworthy that we have also explored the viability of employing the Bayesian Personalized Ranking (BPR) Loss~\citep{rendle2012bpr} within our framework. However, subsequent experimentation reveals that the BPR Loss does not offer superior performance when compared to the binary cross-entropy loss. Consequently, we opt to use the binary cross-entropy loss as our loss function.

\subsection{Hyper-parameter Settings}\label{sec:hyperparameters}
\textbf{Large Language Models.} We perform experiments with two large language models. For {\sc GPT-3} ({\tt text-davinci-003}), {\tt temperature}, {\tt max\_token}, {\tt top\_p}, {\tt frequency penalty}, and {\tt presence penalty} are set as 0, 512, 1, 0.0, and 0.6, respectively. For {\sc Llama-2} (7B {\sc Llama-2-Chat}), we set {\tt do\_sample} to be {\tt True}, {\tt top\_k} 10, and the {\tt num\_return\_sequences} 1. {\sc Llama-2}'s generation is conducted on 8 NVIDIA GeForce RTX 2080 Ti GPUs, each equipped with 11 GB of memory.

\noindent\textbf{Recommendation Modules} We initialize the model parameters randomly, following a Gaussian distribution. To optimize the framework, we employ the AdamW algorithm~\cite{loshchilov2017decoupled} with a weight decay value of 0.0005. For the MLP model, the hyper-parameter grids for the learning rate and dropout rate are $\{0.0001, 0.0005, 0.001\}$ and $\{0.1, 0.3, 0.5\}$, respectively. For AutoInt~\citep{song2019autoint}, the hyper-parameter grids for the learning rate, dropout rate, hidden layer size, number of attention layers, and attention heads are $\{0.001, 0.005, 0.01\}$, $\{0.1, 0.3, 0.5\}$, $\{16, 32, 64, 128\}$, $\{1, 2\}$, and $\{1, 2\}$, respectively. For DCN-V2~\citep{wang2021dcn}, the learning rate, dropout rate, hidden layer size, and number of cross layers are searched in $\{0.001, 0.005, 0.01\}$, $\{0.1, 0.3, 0.5\}$, $\{16, 32, 64, 128\}$, $\{1, 2\}$, and $\{1, 2\}$, respectively. Since the network structure of EDCN~\citep{chen2021enhancing} is similar with DCN-V2~\citep{wang2021dcn}, we apply the hyper-parameter settings of DCN-V2 to EDCN. The performance is evaluated every five epochs, and the early stop mechanism is configured to have a patience of 5. We set the batch size to 4096 for all baselines except for AutoInt which is 1024 due to the memory limitation. Settings that achieve the highest Recall@K on the validation set are chosen for the evaluation on the testing set. 

\subsection{Importance Measurement for Engagement-guided Prompting}\label{sec:importance}
In our study, we show an example of using Personalized PageRank (PPR)~\citep{brin1998pagerank} score as the metric to find the important neighbor items. PPR is a widely employed technique for finding significant neighbors in recommendation systems~\citep{10.1145/3219819.3219890}. In particular, we first construct the user-item bipartite graph $G=(V, E)$. In this notation, $G$ represents the bipartite graph, $E$ denotes the set of nodes, and $E$ represents the set of edges. There are two types of nodes including users $V_{user} \subset V$ and items $V_{item} \subset V \:(V_{user} \cup V_{item} = V, V_{user} \cap V_{item} = \varnothing)$. An edge $e \in E$ between a user node $v \in V_{user}$ and an item node $i \in V_{item}$ is created if this user interacts with this item. 

Next, we proceed by calculating the Personalized PageRank (PPR) score for each item node. The PPR value $\pi(s, t)$, where $s$ is the source node and $t$ is the target node, signifies the probability that a random walk initiated from node $s$ concludes at node $t$. This value offers a quantified measure of their relative importance from the standpoint of an individual node~\citep{lofgren2015efficient}. For every item node, we construct a set of significant neighboring items. By identifying the top $T$ item nodes with the highest PPR scores, we pinpoint important neighbor items guided by user engagement. The rationale behind this approach lies in the observation that when users frequently engage with two items, there tends to be a greater similarity between these two items through the lens of user preferences. By incorporating this information, we aim to capture user preferences more effectively, leading to enhanced performance in content recommendation. For both datasets, we set $T=3$. For the Movielens-1M dataset, we find the important neighbor items that share the same genre as the target item.

\subsection{Implementation Details}\label{appendix:implementation}

Our methods are implemented and experiments are conducted using PyTorch. The computation of PPR scores is facilitated by the use of the {\tt torch-ppr} library. Each experiment is run on one NVIDIA A100 GPU with 80 GB of memory at a time. Further, we adapt the codes of the DeepCTR\footnote{\url{https://github.com/shenweichen/DeepCTR}} and DeepCTR-Torch\footnote{\url{https://github.com/shenweichen/DeepCTR-Torch}} repositories to implement AutoInt~\citep{song2019autoint}, DCN-V2~\citep{wang2021dcn}, and EDCN~\citep{chen2021enhancing}. Table~\ref{tab:tab_all_prompt} summarizes all prompts of \model.

\begin{table*}[t]
    \centering
    \adjustbox{max width=\linewidth}{
    \begin{tabular}{c l}
    \toprule[1.1pt]
    $p_{para}$      &  {\it ``The description of an item is as follows} `\{{\tt description}\}', {\it paraphrase it.''} \\
    $p_{tag}$     &  \textit{``The description of an item is as follows} `\{{\tt description}\}', {\it summarize it with tags.''}\\
    $p_{infer}$ & \textit{``The description of an item is as follows} `\{{\tt description}\}', {\it what kind of emotions can it evoke?''} \\ \midrule
    $p_{para}^{rec}$ & \textit{``The description of an item is as follows} `\{{\tt description}\}', {\it \purpletext{what else should I say if I want to recommend it to others?}''} \\
    $p_{tag}^{rec}$ & \textit{``The description of an item is as follows} `\{{\tt description}\}', {\it what tags should I use \purpletext{if I want to recommend it to others?}''}\\
    $p_{infer}^{rec}$ & \textit{``The description of an item is as follows} `\{{\tt description}\}', {\it \purpletext{recommend it to others} with a focus on the emotions it can evoke.''} \\ \midrule
     $p^{eng}$ & \textit{``\greentext{Summarize the commonalities among the following descriptions:}} ‘\{{\tt description}\}’; \greentext{‘\{{\tt descriptions of other important neighbors}\}’.}{\it''}\\ \midrule
    $p^{rec+eng}$   & \begin{tabular}[c]{@{}l@{}} \textit{``The description of an item is as follows:} ‘\{{\tt description}\}’. {\it \purpletext{What else should I say if I want to recommend it to others?}} \\ {\it This content is considered to \greentext{hold some similar attractive characteristics as the following descriptions:}} \greentext{‘\{{\tt descriptions of other important neighbors}\}’.}{\it''}\end{tabular} \\
        \bottomrule[1.1pt]
    \end{tabular}}
    \caption{\model prompts.}
    \label{tab:tab_all_prompt}
\end{table*}

\subsection{KAR Augmentation Details}\label{appendix:kar_details}

In KAR, \citet{xi2023towards} applied a specific prompt to elicit factual knowledge about movies of the Movielens-1M dataset~\citep{harper2015movielens}. The prompt instructed the model to: \textit{``Introduce movie {\tt \{item description\}} and describe its attributes precisely (including but not limited to {\tt scenario-specific factors})''}. In their study, the {\tt item description} was the movie titles. Human experts were enlisted to refine the answers generated by LLMs in response to the question: \textit{``List the importance factors or features that determine whether a user will be interested in a movie.''} These refined factors were then considered as the {\tt \{scenario-specific factors\}}, including \textit{genre, actors, directors, theme, mood, production quality, and critical acclaim}. Because the responses generated using these prompts were not publicly released, we re-implement the same methodology, employing LLMs to generate the factual knowledge of items. In the case of the Recipe dataset~\citep{majumder2019generating}, we use recipe description as the {\tt item description}. The same approach is then adopted to identify {\tt scenario-specific factors}. Initially, the prompt is adapted to: \textit{``List the importance factors or features that determine whether a user will be interested in a recipe.''} Subsequently, the answers generated by {\sc ChatGPT} are validated (see Table~\ref{tab:appendix_KAR_factors}). The resulting set of {\tt scenario-specific factors} for Recipe comprises a diverse range of attributes, including \textit{dietary preferences, ingredients, cuisine type, cooking time, nutritional value, allergies, taste preferences, skill level, occasion, cost, health and wellness goals, food allure, reviews and ratings, cooking equipment, personal experience, season and weather, cultural or ethical considerations, creativity and variety, simplicity, popularity and trends}. These prompts are then employed to enrich the factual knowledge of both movies and recipes using {\sc GPT-3} ({\tt text-davinci-003}). For illustrative examples of the responses, please refer to Table~\ref{tab:appendix_KAR_examples}. KAR is also composed of a preference reasoning prompt for user information augmentation. Since we only focus on the item side, only the item factual prompt is implemented to examine how different focuses on LLMs' ability between \model and KAR affect recommendation performance.

\subsection{Keywords Construction}\label{appendix:keyword_construction}
The keyword generation process differs between the Movielens-1M and Recipe datasets. For Movielens-1M, the keywords are derived from genre labels, which are intrinsic components of the dataset's metadata. In the case of Recipe, the process involves multiple steps. Initially, we compile a list of unique words found in the responses generated through the recommendation-driven prompting strategy. Subsequently, we filter out stopwords and proceed to construct unigrams and bigrams using the {\tt NLTK} package. Following this, a manual review is conducted to identify phrases that appear at least five times in the corpus. These phrases are then scrutinized to determine whether they contain words relevant for categorizing recipes. The final list of keywords for Recipe contain ``easy'', ``homemade'', ``baking'', ``health'', ``healthy'', ``dessert'', and ``dinner''. These keywords collectively serve as indicative descriptors for recipes within the dataset.

\begin{table*}[ht]
    \centering
    \small
    \adjustbox{max width=\textwidth}{
    \begin{tabular}{p{\textwidth}}
        \toprule[1.1pt]

        The factors or features that determine whether a user will be interested in a recipe can vary from person to person, but some important factors commonly include:

\textbf{1. Dietary Preferences:} Whether the recipe aligns with the user's dietary restrictions, such as vegetarian, vegan, gluten-free, or keto.

\textbf{2. Ingredients:} The availability and appeal of the ingredients used in the recipe.

\textbf{3. Cuisine Type:} Whether the recipe falls within a cuisine the user enjoys, like Italian, Mexican, or Asian.

\textbf{4. Cooking Time:} The user's available time for cooking, as some may prefer quick and easy recipes, while others enjoy longer cooking processes.

\textbf{5. Nutritional Value:} Users may consider the nutritional content of the recipe, such as calorie count, protein, or fiber.

\textbf{6. Allergies:} Any known food allergies that need to be accommodated in the recipe.

\textbf{7. Taste Preferences:} Whether the flavors and taste profile of the recipe match the user's preferences.

\textbf{8. Skill Level:} Some users may look for recipes that match their cooking skills, such as beginner-friendly or gourmet.

\textbf{9. Occasion:} If the recipe is suitable for a particular occasion, like a holiday, party, or a regular weekday meal.

\textbf{10. Cost:} The affordability of the ingredients and whether it fits within the user's budget.

\textbf{11. Health and Wellness Goals:} Recipes that align with specific health or weight management goals, like low-carb or heart-healthy.

\textbf{12. Food Allure:} The visual appeal and presentation of the dish through images and descriptions.

\textbf{13. Reviews and Ratings:} User reviews and ratings on the recipe, which can indicate its popularity and quality.

\textbf{14. Cooking Equipment:} Whether the user has the necessary kitchen tools and appliances for the recipe.

\textbf{15. Personal Experience:} Previous experiences with similar recipes or dishes may influence interest.

\textbf{16. Season and Weather:} Seasonal ingredients or dishes that match the current season or weather.

\textbf{17. Cultural or Ethical Considerations:} Recipes that respect cultural, ethical, or environmental values.

\textbf{18. Creativity and Variety:} The desire for trying something new and innovative in cooking.

\textbf{19. Simplicity:} Some users may prefer straightforward, minimal-ingredient recipes.

\textbf{20. Popularity and Trends:} Trending recipes or those recommended by friends and family may pique a user's interest.

These factors can significantly impact a user's decision to explore a recipe, and personalized recommendation systems often consider several of these factors to suggest recipes tailored to individual preferences.\\

        \bottomrule[1.1pt]
    \end{tabular}}
    \caption{Response generated by {\sc ChatGPT} about the question \textit{``List the importance factors or features that determine whether a user will be interested in a recipe.''} }
    \label{tab:appendix_KAR_factors}
\end{table*}
\begin{table*}[ht]
    \centering
    \small
    \adjustbox{max width=\textwidth}{
    \begin{tabular}{p{\textwidth}}
        \toprule[1.1pt]
        \textbf{Prompt:} Introduce movie Batman Forever and describe its attributes precisely (including but not limited to genre, actors, directors, theme, mood, production quality, and critical acclaim)\\

        \textbf{Response:} Batman Forever is a 1995 American superhero film directed by Joel Schumacher and produced by Tim Burton. It is the third installment in the Batman film series, based on the DC Comics character of the same name. The film stars Val Kilmer as Bruce Wayne/Batman, alongside Tommy Lee Jones, Jim Carrey, Nicole Kidman, Chris O'Donnell, Michael Gough, and Pat Hingle. The film follows Batman as he battles Two-Face and The Riddler, both of whom are determined to wreak havoc in Gotham City. The film's theme is the struggle between good and evil, and the power of redemption. The mood of the film is dark and brooding, with a hint of humor. The production quality of Batman Forever is high, with impressive special effects and cinematography. The film was well-received by critics, who praised its action sequences, performances, and visual style. It was a box office success, grossing over \$336 million worldwide.\\\midrule
        
        \textbf{Prompt:} Introduce recipe `Numerous Amish bread recipes have been shared, but none seem to include the starter. Therefore, I thought it would be helpful to provide this. Happy baking! Here's the Amish Friendship Bread and its starter.' and describe its attributes precisely (including but not limited to dietary preferences, ingredients, cuisine type, cooking time, nutritional value, allergies, taste preferences, skill level, occasion, cost, health and wellness goals, food allure, reviews and ratings, cooking equipment, personal experience, season and weather, cultural or ethical considerations, creativity and variety, simplicity, popularity and trends) \\
        
        \textbf{Response:} Amish Friendship Bread is a sweet, yeasty bread that is made with a starter that is passed from person to person. It is a traditional Amish recipe that has been around for generations.
        
        Dietary Preferences: Vegetarian, Gluten-Free (if using gluten-free flour)
        
        Ingredients: All-purpose flour, sugar, active dry yeast, milk, butter, eggs, vanilla extract, cinnamon, and salt.
        
        Cuisine Type: American
        
        Cooking Time: 10 minutes to prepare the starter, 1 hour to prepare the dough, and 1 hour to bake the bread.Nutritional Value: High in carbohydrates, low in fat and protein.
        
        Allergies: Contains dairy, eggs, and wheat.
        
        Taste Preferences: Sweet, yeasty, and slightly spiced.
        
        Skill Level: BeginnerOccasion: Breakfast, brunch, or snack.
        
        Cost: Low
        
        Health and Wellness Goals: A healthier alternative to store-bought breads.
        
        Food Allure: Comfort food that is easy to make.
        
        Reviews and Ratings: Generally positive reviews.
        
        Cooking Equipment: Large bowl, measuring cups and spoons, wooden spoon, loaf pan, and oven.
        
        Personal Experience: Delicious and easy to make.
        
        Season and Weather: Can be made year-round.
        
        Cultural or Ethical Considerations: Traditional Amish recipe.
        
        Creativity and Variety: Can be adapted to include different flavors and ingredients.
        
        Simplicity: Easy to make with basic ingredients.
        
        Popularity and Trends: Popular among home bakers.
\\
        \bottomrule[1.1pt]
    \end{tabular}}
    \caption{Response generated via the item factual prompt proposed by \citet{xi2023towards}. }
    \label{tab:appendix_KAR_examples}
\end{table*}

\clearpage

\clearpage
\section{Additional Analysis}\label{appendix: additional_analysis}

\subsection{Additional Discussions on Augmented Text}\label{appendix:augmented_text_understanding}
Overall, we observe that \model effectively elicits LLMs to enrich item descriptions with additional information. This augmentation results in descriptions that are not only more detailed but also more expressive, enhancing the original item descriptions.  Tables~\ref{tab:example_gpt_tag} and \ref{tab:example_gpt_infer} show example responses generated by {\sc GPT-3} comparing the recommendation-driven and basic promptings in terms of {\tt tag} and {\tt infer}. Tables~\ref{tab:example_llama2_para}, \ref{tab:example_llama2_tag}, and \ref{tab:example_llama2_infer} show example responses generated by {\sc Llama-2} comparing the recommendation-driven and basic promptings. Responses from the recommendation prompting strategies provides several additional pieces of information and context compared to the responses from the basic prompting strategies. Take the responses in Table~\ref{tab:example_gpt_tag} as an example, the recommendation prompting strategy introduces new themes like ``Supernatural'', ``Paranormal'', and ``Psychological Thriller'', which are not present in the responses from the basic promptings. These themes suggest a broader and more specific context for the story, indicating not just communication with the dead, but also elements of horror and suspense. The term ``Troubled Child'' adds a new dimension to the ``Young Boy'' mentioned in the first sentence, suggesting that the child's character may face internal conflicts or challenges.

\begin{figure*}[t]
    \centering
    \includegraphics[width=\linewidth]{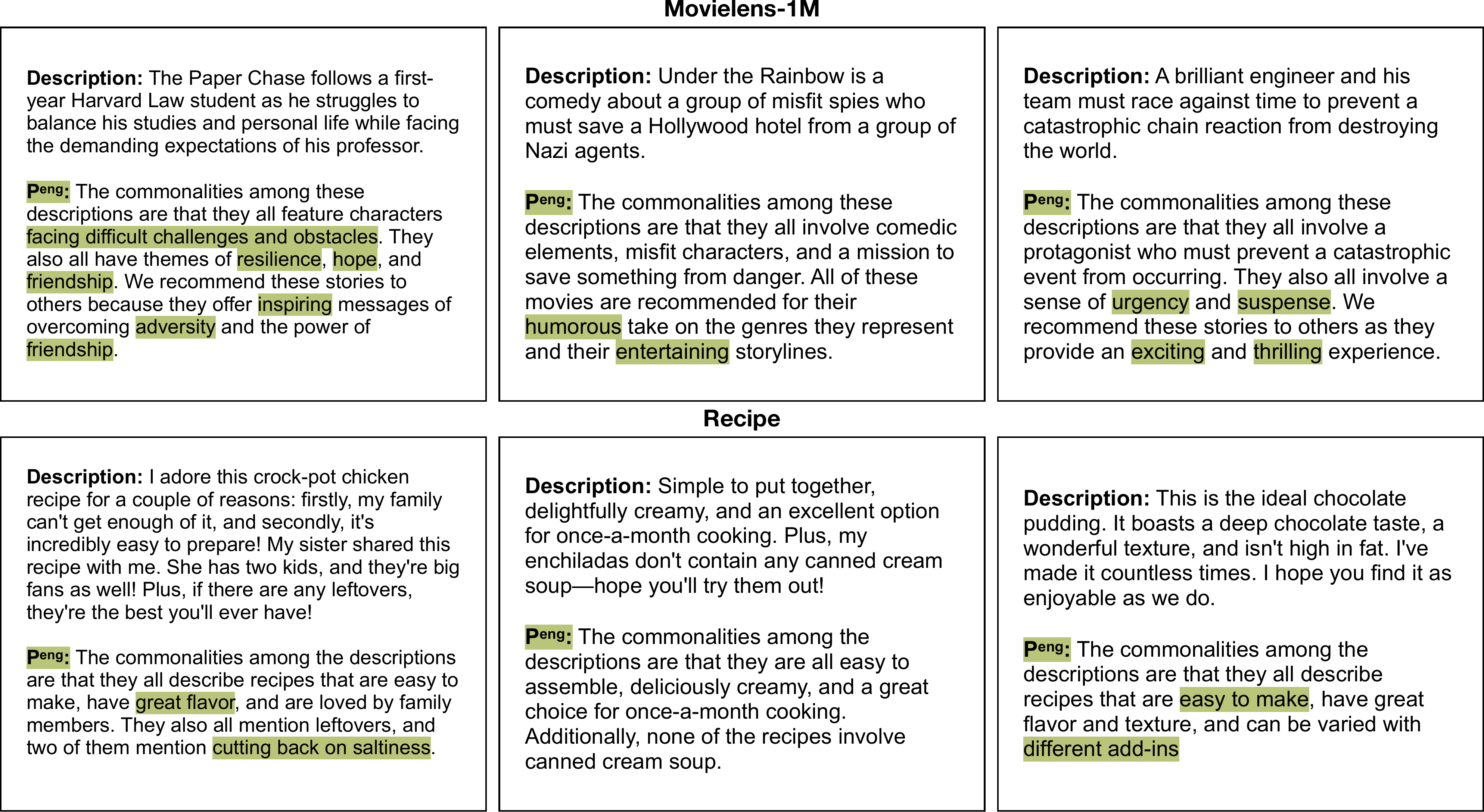}
    \caption{Example responses generated by {\sc GPT-3}. The additional information augmented via the engagement-guided prompting is colored \greentext{green}. We choose the example responses generated by {\sc GPT-3} for illustration. Examples generated by {\sc Llama-2} can be found in Appendix~\ref{appendix:augmented_text_understanding}.}
    \label{fig:case_eng}
\end{figure*}

\noindent\textbf{What extra information does engagement-guided strategy prompt LLMs to augment?} Consistent with our previous experiments, we curate exemplary responses obtained from $p^{eng}$ for closer examination (Figure~\ref{fig:case_eng}). Our analysis reveals a distinct pattern compared to what we have observed with recommendation-driven prompting. There are primarily two scenarios to consider. First, if the descriptions of the important neighbor items and the target items exhibit high similarity, the impact of $p^{eng}$ resembles that of $p_{para}$, as exemplified in the second Recipe example in Figure~\ref{fig:case_eng}. Second, $p^{eng}$ guides LLMs to generate additional information, which may be derived from the descriptions of the important neighbor items. Consequently, how the engagement-guided strategy influences LLMs' text generation—whether it aligns with one of the behaviors we have described, both of them, or even other unexplored patterns—largely depends on the composition of the important neighbor items. This composition, in turn, is contingent on the neighbor sampling method which is out of the scope of our study. We leave a more in-depth exploration of this topic to future research endeavors.

Interestingly, the recommendation-driven + engagement-guided prompting strategy is able to generate text that shares similar characteristics with both sub-strategies. How they quantitatively form the final generation remains an open question. Examples can be found in Table~\ref{tab:example_gpt_rec_eng}.

Table~\ref{tab:example_llama2_eng} shows example responses of {\sc Llama-2} to the engagement-guided prompting strategy. Table~\ref{tab:example_gpt_rec_eng} shows example responses of {\sc GPT-3} to the recommendation-driven and engagement-guided prompting strategy. Overall, the 7B {\sc Llama-2-Chat} performs poorly compared to {\sc GPT-3}. In some cases, there is no generated content as we have also observed in Appendix~\ref{sec:example_responses_appendix}.

\subsection{Additional Discussions on Applicable Item Domains and Available Textual Information}\label{appendix:applicability}

The applicability of \model beyond movies and recipes, particularly in domains with sparse textual information, remains a question. To address this, we conduct an analysis to determine \model's efficacy in enriching text across various domains and text lengths. We use item descriptions from ten distinct domains in the Amazon review dataset~\cite{ni2019justifying}, which includes product metadata like descriptions, category, price, brand, and image features. The selected domains are all beauty, appliances, automotive, digital music, grocery and gourmet food, pet supplies, sports and outdoors, video games, magazine subscriptions, and industrial and scientific. For each domain, we sample 50 items and prompt {\sc GPT-3} using $p^{rec}_{para}$ (\ie recommendation-driven prompting).

Our previous discussions highlight that the most valuable information for improving recommendation performance typically aligns with expressive words pertinent to item characteristics. While no single metric directly quantifies this added information, we use the increase in the number of adjectives as a proxy. Additionally, the total word count serves as a straightforward metric to approximate the volume of augmented information. By comparing the augmented texts with the original item descriptions, we calculate the percentage increase in the number of adjectives. Note that the adjective increase is computed as a ratio of the difference in adjective count to the original word count.

As Figure~\ref{fig:applicable_domain} demonstrates, \model effectively enriches item descriptions across multiple domains, including those lacking in rich textual content. For instance, the average word counts in movie and digital music descriptions are only 20.34 and 30.18 words, respectively. \model enhances expressiveness, with a notable increase in the use of adjectives.

\subsection{Additional Experiments on Applying LLM-Rec to Other Baselines}\label{appendix_sec: llm-rec_to_other_baselines}
We extend the application of \model to other text-based recommendation systems and replicate the experiments. The results, as presented in Table~\ref{tab:other_baselines_with_llm_rec}, indicate that \model can be easily adapted to various text-based recommendation systems and generally enhances recommendation performance compared to using the original text.

\begin{table*}[t]
    \centering
    \adjustbox{max width=\textwidth}{
    \begin{tabular}{lccccccc}
    \toprule[1.1pt]
   & & \multicolumn{3}{c}{\textbf{Movielens-1M}}   & \multicolumn{3}{c}{\textbf{Recipe}}  \\
  &    & Precision@10  & Recall@10      & NDCG@10       & Precision@10      & Recall@10    &NDCG@10         \\ \midrule
\multicolumn{2}{l}{AutoInt~\citep{song2019autoint}}                                                & 0.2149  \scriptsize{$\textcolor{gray}{\pm 0.0078}$} &  0.1706  \scriptsize{$\textcolor{gray}{\pm 0.0075}$}    & 0.2698 \scriptsize{$\textcolor{gray}{\pm 0.0092}$} & 0.0351 \scriptsize{$\textcolor{gray}{\pm 0.0032}$}  & 0.0772 \scriptsize{$\textcolor{gray}{\pm 0.0102}$}& 0.0658 \scriptsize{$\textcolor{gray}{\pm 0.0089}$} \\

\multicolumn{2}{l}{\quad - Basic} & 0.2590  \scriptsize{$\textcolor{gray}{\pm 0.0038}$} &  0.2193  \scriptsize{$\textcolor{gray}{\pm 0.0049}$}    & 0.3224 \scriptsize{$\textcolor{gray}{\pm 0.0052}$} & 0.0361 \scriptsize{$\textcolor{gray}{\pm 0.0030}$}  & 0.0797 \scriptsize{$\textcolor{gray}{\pm 0.0097}$}& 0.0664 \scriptsize{$\textcolor{gray}{\pm 0.0076}$} \\

\multicolumn{2}{l}{\quad - Rec} & 0.2593  \scriptsize{$\textcolor{gray}{\pm 0.0035}$} &  0.2197  \scriptsize{$\textcolor{gray}{\pm 0.0068}$}    & 0.3242 \scriptsize{$\textcolor{gray}{\pm 0.0059}$} & 0.0357 \scriptsize{$\textcolor{gray}{\pm 0.0029}$}  & 0.0794 \scriptsize{$\textcolor{gray}{\pm 0.0096}$}& 0.0660 \scriptsize{$\textcolor{gray}{\pm 0.0079}$} \\

\multicolumn{2}{l}{\quad - Eng} & 0.2323  \scriptsize{$\textcolor{gray}{\pm 0.0011}$} &  0.1857  \scriptsize{$\textcolor{gray}{\pm 0.0032}$}    & 0.2899 \scriptsize{$\textcolor{gray}{\pm 0.0031}$} & 0.0349 \scriptsize{$\textcolor{gray}{\pm 0.0028}$}  & 0.0764 \scriptsize{$\textcolor{gray}{\pm 0.0090}$}& 0.0642 \scriptsize{$\textcolor{gray}{\pm 0.0072}$} \\

\multicolumn{2}{l}{\quad - Rec+Eng} & 0.2620  \scriptsize{$\textcolor{gray}{\pm 0.0021}$} &  0.2230  \scriptsize{$\textcolor{gray}{\pm 0.0037}$}    & 0.3270 \scriptsize{$\textcolor{gray}{\pm 0.0022}$} & 0.0349 \scriptsize{$\textcolor{gray}{\pm 0.0029}$}  & 0.0759 \scriptsize{$\textcolor{gray}{\pm 0.0099}$}& 0.0647 \scriptsize{$\textcolor{gray}{\pm 0.0075}$} \\\midrule

\multicolumn{2}{l}{DCN-V2~\citep{wang2021dcn}}   & 0.2961  \scriptsize{$\textcolor{gray}{\pm 0.0050}$}   & 0.2433 \scriptsize{$\textcolor{gray}{\pm 0.0057}$}   & 0.3689   \scriptsize{$\textcolor{gray}{\pm 0.0033}$}      & 0.0360 \scriptsize{$\textcolor{gray}{\pm 0.0036}$} & 0.0786 \scriptsize{$\textcolor{gray}{\pm 0.0104}$} & 0.0653  \scriptsize{$\textcolor{gray}{\pm 0.0085}$}\\

\multicolumn{2}{l}{\quad - Basic}   & 0.3083  \scriptsize{$\textcolor{gray}{\pm 0.0033}$}   & 0.2601 \scriptsize{$\textcolor{gray}{\pm 0.0051}$}   & 0.3842   \scriptsize{$\textcolor{gray}{\pm 0.0054}$}      & 0.0365 \scriptsize{$\textcolor{gray}{\pm 0.0028}$} & 0.0802 \scriptsize{$\textcolor{gray}{\pm 0.0093}$} & 0.0658  \scriptsize{$\textcolor{gray}{\pm 0.0084}$}\\

\multicolumn{2}{l}{\quad - Rec}   & 0.3062  \scriptsize{$\textcolor{gray}{\pm 0.0031}$}   & 0.2572 \scriptsize{$\textcolor{gray}{\pm 0.0049}$}   & 0.3831   \scriptsize{$\textcolor{gray}{\pm 0.0041}$}      & 0.0362 \scriptsize{$\textcolor{gray}{\pm 0.0035}$} & 0.0794 \scriptsize{$\textcolor{gray}{\pm 0.0108}$} & 0.0670  \scriptsize{$\textcolor{gray}{\pm 0.0095}$}\\

\multicolumn{2}{l}{\quad - Eng}   & 0.2990  \scriptsize{$\textcolor{gray}{\pm 0.0024}$}   & 0.2496 \scriptsize{$\textcolor{gray}{\pm 0.0020}$}   & 0.3725   \scriptsize{$\textcolor{gray}{\pm 0.0021}$}      & 0.0356 \scriptsize{$\textcolor{gray}{\pm 0.0032}$} & 0.0786 \scriptsize{$\textcolor{gray}{\pm 0.0094}$} & 0.0647  \scriptsize{$\textcolor{gray}{\pm 0.0076}$}\\

\multicolumn{2}{l}{\quad - Rec+Eng}   & 0.3114  \scriptsize{$\textcolor{gray}{\pm 0.0021}$}   & 0.2641 \scriptsize{$\textcolor{gray}{\pm 0.0038}$}   & 0.3882   \scriptsize{$\textcolor{gray}{\pm 0.0028}$}      & 0.0357 \scriptsize{$\textcolor{gray}{\pm 0.0034}$} & 0.0793 \scriptsize{$\textcolor{gray}{\pm 0.0104}$} & 0.0654  \scriptsize{$\textcolor{gray}{\pm 0.0083}$}\\

             \bottomrule[1.1pt]
\end{tabular}}
\caption{Average recommendation performance by applying \model to other text-based recommendation modules across five different train/test splits.}
    \label{tab:other_baselines_with_llm_rec}
\end{table*}

\subsection{Additional Discussions on Integration Process}\label{appendix_sec: integration}
 In our setup, the text encoder is frozen (not fine-tuned), with a fixed output dimension for all vectors. The fundamental difference between the {\tt Concat-All} and Text Concatenation methods lies in their processing sequence. The {\tt Concat-All} method initially transforms individual text segments into embeddings and subsequently concatenates these embeddings. In contrast, the Text Concatenation method first concatenates the text segments and then converts this combined text into a single embedding.

The observed superiority of the {\tt Concat-All} method can be attributed to how these processes handle information density. When lengthy text segments are concatenated before encoding, there is a higher likelihood of \textbf{information loss}, particularly given the constraints of a frozen text encoder. This encoder, \textbf{not} being fine-tuned for the specific nuances of our data, may struggle to effectively capture and retain crucial information from longer text inputs. Therefore, processing shorter text segments individually before concatenation (as in {\tt Concat-All}) may help in preserving important features and nuances in the embeddings.

\subsection{Additional Discussions on Prompt Design}\label{appendix:prompt_design}

To investigate whether modifying the word choice in the designed prompts significantly affects the augmented output, we construct one variant prompt for each of \model's prompts, ensuring they convey the same meaning but with different word choices. Take $p_{para}$ as an example, $p_{para}$ is {\it ``The description of an item is as follows `\{{\tt description}\}', paraphrase it.''}. One variant is {\it ``Summarize the given item description, `\{{\tt description}\}', using different words.''} Next, we randomly sample 50 items from Movielens-1M, and prompt {\sc GPT-3} with these variants. The cosine similarity between the responses generated from the variant prompt and \model's prompt is computed and shown in Table~\ref{tab:appendix_variant_quantitative}. Tables~\ref{tab:variant_basic_para}-\ref{tab:variant_rec_eng} shows the example responses.

\subsection{Additional Discussions on Dynamic Prompts}\label{appendix:dynamic_prompt}
The concept of dynamic prompts in recommendation systems is an intriguing area that holds the potential for enhancing personalization. By incorporating descriptions of a user's most recently interacted items into prompts, the system can generate item descriptions on-the-fly that are more closely aligned with the user's current interests and preferences. This approach could lead to more precise and tailored recommendations, as the generated descriptions would reflect the user's evolving tastes.

One of the primary considerations is the computational cost associated with generating dynamic prompts. Each user interaction would require real-time processing to update the prompt, which could be resource-intensive, especially for large-scale systems with many users and items.

To mitigate computational costs, several strategies can be employed.
Developing efficient algorithms for prompt generation and item description generation can help mitigate computational costs. 
Implementing caching mechanisms for frequently accessed data can reduce the processing time required for updating prompts.
Instead of completely regenerating prompts after each interaction, the system could employ incremental updates to modify prompts based on recent changes in user behavior.

While the implementation of dynamic prompts presents several challenges, it also offers a promising avenue for enhancing personalization in recommendation systems. With careful consideration, this approach has the potential to cater more effectively to individual user needs.

\section{Extended Related Work}\label{sec:detailed_related_work}
\textbf{Augmentation in Text-based Recommendation.} Traditionally, researchers have advocated the augmentation of item descriptions through the incorporation of external knowledge sources~\citep{di2012linked,musto2018semantics,sachdeva2020useful}. Notably, \citet{di2012linked} harnesse data from external databases such as {\tt DBpedia}~\citep{bizer2009dbpedia}, {\tt Freebase}~\citep{bollacker2008freebase}, and {\tt LinkedMDB}~\citep{hassanzadeh2009linked} to gather comprehensive information pertaining to movies, including details about actors, directors, genres, and categories. This approach aimed to enrich the background knowledge available to movie recommender systems. The explicit semantics embedded in these external knowledge sources have demonstrated a discernible enhancement in recommendation performance~\citep{musto2017introducing}. However, this process necessitates a profound domain expertise to effectively and efficiently select and leverage the precise database, ensuring the incorporation of genuinely valuable information into item descriptions~\citep{dumitru2011demand}.

\noindent\textbf{LLM for Recommendation.} The use of large language models in recommender systems has garnered significant attention in recent research~\cite{lin2023can, chen2023large}. Many studies have explored the direct use of LLMs as recommender models. The underlying principle of these approaches involves constructing prompts that encompass the recommendation task, user profiles, item attributes, and user-item interactions. These task-specific prompts are then presented as input to the LLMs, which is instructed to predict the likelihood of interaction between a given user and item~\citep{dai2023uncovering, gao2023chat,geng2022recommendation,li2023exploring,liu2023chatgpt,zhang2023recommendation}. For instance, \citet{wang2023zero} designed a three-step prompting strategy to directly guide LLMs to capture users' preferences, select representative previously interacted items, and recommend a ranked list of 10 items. While these works demonstrate the potential of LLMs as powerful recommender models, the focus primarily revolves around utilizing the LLMs directly for recommendation purposes. However, a potential issue of these methods is that LLMs may generate predictions merely from memorizing training samples which poses a challenge for conducting effective evaluations. In this study, we approach the problem from a different perspective. Rather than using LLMs as recommender models, this study explores diverse prompting strategies to \textit{augment input text} with LLMs for personalized content recommendation. The actual recommendation process still relies on existing recommendation methodologies.

\section{Example Responses}\label{sec:example_responses_appendix}

Tables~\ref{tab:example_response_ml1m} and \ref{tab:example_response_recipe} show example responses by {\sc GPT-3} and the 7B {\sc Llama-2-Chat} on Movielens-1M~\citep{harper2015movielens} and Recipe~\citep{majumder2019generating}. Augmented components are highlighted (recommendation-driven: \purpletext{blue}; engagement-guided: \greentext{green}; rec+eng: \orangetext{orange}). In summary, both {\sc GPT-3} and {\sc Llama-2} exhibit the capability to enrich item descriptions with supplementary information. Nevertheless, the {\sc Llama-2-Chat} model with its 7B parameters demonstrates comparatively poorer performance, which could be attributed to its limited parameter scale. This limitation offers insight into the diminished recommendation quality when using {\sc Llama-2} responses in contrast to {\sc GPT-3}. Future research endeavors should focus on optimizing the \model framework, particularly concerning the selection of different large language models as backbones, to enhance recommendation outcomes.

\begin{table*}[t]
    \centering
    \small
    \adjustbox{max width=\textwidth}{
    \begin{tabular}{p{\textwidth}}
        \toprule[1.1pt]

        \textbf{Description}: A troubled child psychologist helps a young boy who is able to see and communicate with the dead.
\\
\vspace{-1mm}
$p_{tag}$: \#Psychology \#Child \#YoungBoy \#Dead \#Communication \\
$p^{rec}_{tag}$: Tags: Child Psychology, Troubled Child, Young Boy, Communicate with Dead, \purpletext{Supernatural}, \purpletext{Paranormal}, \purpletext{Ghost Story}, Psychological \purpletext{Thriller} \\
\midrule
        \textbf{Description}: This unique Tex-Mex recipe was created by a distant relative who was stationed at the Alamo almost 160 years ago. Family stories say that he was searching for quality tomatoes when Santa Ana launched an attack on the now-iconic Texas landmark. Fortunately, our ancestor had hidden this recipe in his coon-skin cap's lining, ensuring that both he and his avocado dip survived to see another day. Years later, as he lay dying from a chihuahua bite infection, he gazed deeply into his son's eyes and, with his final breath, whispered the memorable words, ``Remember... the... guacamole.''
\\
\vspace{-1mm}
$p_{tag}$: \#TexMex \#Alamo \#AvocadoDip \#CoonSkinCap \#SantaAna \#Chihuahua \#Guacamole \\
$p^{rec}_{tag}$: Tags: Tex-Mex, Alamo, Santa Ana, Avocado Dip, Chihuahua, Guacamole, \purpletext{Recipe}, \purpletext{Family Lore} \\

        \bottomrule[1.1pt]
    \end{tabular}}
    \caption{Example responses to $p_{tag}$ and $p^{rec}_{tag}$ generated by {\sc GPT-3}. The additional information augmented via the
recommendation-driven prompting is highlighted in \purpletext{blue}. }
    \label{tab:example_gpt_tag}
\end{table*}

\begin{table*}[t]
    \centering
    \small
    \adjustbox{max width=\textwidth}{
    \begin{tabular}{p{\textwidth}}
        \toprule[1.1pt]

        \textbf{Description}: A former Roman General is forced to become a gladiator and seek revenge against the corrupt emperor who murdered his family.
\\
\vspace{-1mm}
$p_{infer}$: This description can evoke a range of emotions, including anger, sadness, revenge, determination, and courage. \\
$p^{rec}_{infer}$: This item is sure to evoke strong emotions in its viewers. It tells the story of a former Roman General who is forced to become a gladiator and seek revenge against the corrupt emperor who murdered his family. It is a tale of tragedy, courage, and justice that will leave you feeling \purpletext{inspired} and \purpletext{moved}. \\
\midrule
\textbf{Description}: Our garden's potatoes are ripe and ready, and this recipe is always the first one I go for! They're incredibly delicious!
\\
\vspace{-1mm}
$p_{infer}$: This description can evoke feelings of excitement, anticipation, joy, and satisfaction. \\
$p^{rec}_{infer}$: This recipe is sure to bring a smile to your face! The potatoes are \purpletext{fresh} from the garden, and the \purpletext{flavor is unbeatable}. Enjoy the feeling of \purpletext{accomplishment} as you savor the deliciousness of these potatoes. You won't regret it!
 \\

        \bottomrule[1.1pt]
    \end{tabular}}
    \caption{Example responses to $p_{infer}$ and $p^{rec}_{infer}$ generated by {\sc GPT-3}. The additional information augmented via the
recommendation-driven prompting is highlighted in \purpletext{blue}. }
    \label{tab:example_gpt_infer}
\end{table*}

\begin{table*}[h]
    \centering
    \small
    \adjustbox{max width=\textwidth}{
    \begin{tabular}{p{\textwidth}}
        \toprule[1.1pt]

        \textbf{Description}: Two agents from a secret organization protect the Earth from alien threats while maintaining the public's ignorance of extraterrestrial life.
\\
\vspace{-1mm}
$p_{para}$: `` Two government agents keep Earth safe from intergalactic dangers while concealing the truth about alien life from the public.'' \\
$p^{rec}_{para}$: ``If you're a fan of \purpletext{action-packed sci-fi with a dash of humor}, then this is the show for you! The agents are \purpletext{hilarious} and the alien threats they face are so \purpletext{creative and thrilling}. Plus, the way they manage to keep the public in the dark about alien life is just genius. It's like a \purpletext{sci-fi comedy version of 'Men in Black' with a lot more explosions. Give it a try and let me know what you think!}'' \\
\midrule
        \textbf{Description}: Indulge in something unique tonight. It's tasty and relatively simple to make. Plus, I usually have all the ingredients on hand.
\\
\vspace{-1mm}
$p_{para}$: Sure! Here's an example paraphrase of the original description \\
$p^{rec}_{para}$: If you're looking for something different and easy to make, try this recipe! \purpletext{It's good and has everything you need.} I usually have these ingredients on hand, so it's a great option when you're feeling \purpletext{spontaneous}. \purpletext{Give it a try and let me know what you think!}\\

        \bottomrule[1.1pt]
    \end{tabular}}
    \caption{Example responses to $p_{para}$ and $p^{rec}_{para}$ generated by {\sc Llama-2}. The additional information augmented via the
recommendation-driven prompting is highlighted in \purpletext{blue}. }
    \label{tab:example_llama2_para}
\end{table*}

\begin{table*}[h]
    \centering
    \small
    \adjustbox{max width=\textwidth}{
    \begin{tabular}{p{\textwidth}}
        \toprule[1.1pt]

        \textbf{Description}: An ant embarks on a journey to save his colony from the oppressive rule of a group of grasshoppers.
\\
\vspace{-1mm}
$p_{tag}$: Sure! Here is an example of how you could summarize the description of the item with tags \\
$p^{rec}_{tag}$: One tag you could use to recommend this book to others is \purpletext{``insects''} or \purpletext{``animals in fiction.''} This tag would \purpletext{help others find the book if they are interested in stories that feature insects or animals as main characters.} \\
\midrule
        \textbf{Description}: I adore this crock-pot chicken recipe for a couple of reasons: first, my family can't get enough of it, and second, it's incredibly simple to prepare! My sister shared this recipe with me. She has two kids, and they're big fans as well! Plus, if there are any leftovers, they're the best you'll ever have!
\\
\vspace{-1mm}
$p_{tag}$: Tags \\
$p^{rec}_{tag}$: If you want to recommend the recipe to others, you could use tags such as \purpletext{``chicken recipe'', ``easy recipe'', ``family-friendly recipe'', ``leftovers recipe'', and ``sister's recipe''}. Explanation \\
        \bottomrule[1.1pt]
    \end{tabular}}
    \caption{Example responses to $p_{tag}$ and $p^{rec}_{tag}$ generated by {\sc Llama-2}. The additional information augmented via the
recommendation-driven prompting is highlighted in \purpletext{blue}. }
    \label{tab:example_llama2_tag}
\end{table*}

\begin{table*}[h]
    \centering
    \small
    \adjustbox{max width=\textwidth}{
    \begin{tabular}{p{\textwidth}}
        \toprule[1.1pt]

        \textbf{Description}: A young police officer must race against time to save a bus full of passengers from a bomb planted by a vengeful criminal.
\\
\vspace{-1mm}
$p_{infer}$: Suspense  Explanation \\
$p^{rec}_{infer}$: I would recommend this movie to others because it evokes a \purpletext{sense of urgency and suspense that is both thrilling and emotionally resonant}. The scene where the young police officer races against time to defuse the bomb is \purpletext{heart-pumping and intense}, and the \purpletext{emotional stakes are high as the officer works to save the passengers}. The movie also explores themes of \purpletext{justice and revenge}, making it a \purpletext{thought-provoking and emotionally impactful viewing experience}. Overall, I would \purpletext{highly recommend this movie to anyone looking for a gripping and emotional thriller}.\\
\midrule
\textbf{Description}: Bored with the same old ground beef recipes? Give this spicy dish a try! If you prefer it extra saucy, feel free to double the sauce amount. Update: Since hoisin sauce is quite sweet, you might want to start with a small amount and gradually increase it to your taste!
\\
\vspace{-1mm}
$p_{infer}$: 1. Desire \\
$p^{rec}_{infer}$: ``Hey fellow foodies! Have you tried this Spicy Beef and Noodles recipe? It's a \purpletext{game changer}! The hoisin sauce adds a depth of flavor that will leave you wanting more, and the spicy kick is perfect for those who like a little heat in their meals. \purpletext{I made it for my family last night and they all loved it! Give it a try and let me know what you think! \#foodie \#delicious \#homemade}''
 \\

        \bottomrule[1.1pt]
    \end{tabular}}
    \caption{Example responses to $p_{infer}$ and $p^{rec}_{infer}$ generated by {\sc Llama-2}. The additional information augmented via the
recommendation-driven prompting is highlighted in \purpletext{blue}. }
    \label{tab:example_llama2_infer}
\end{table*}

\begin{table*}[h]
    \centering
    \small
    \adjustbox{max width=\textwidth}{
    \begin{tabular}{p{\textwidth}}
        \toprule[1.1pt]

        \textbf{Description}: A troubled child psychologist helps a young boy who is able to see and communicate with the dead.
\\
\vspace{-1mm}
$p^{eng}$:  Despite their different subjects and genres, the descriptions share a common theme of \greentext{personal transformation and empowerment.} Each story revolves around a \greentext{character's journey towards self-discovery and self-acceptance, as they navigate challenges and obstacles that help them grow and develop as individuals.}\\
        \bottomrule[1.1pt]
    \end{tabular}}
    \caption{Example responses to $p^{eng}$ generated by {\sc Llama-2}. The additional information augmented via the
engagement-guided prompting is highlighted in \greentext{green}. }
    \label{tab:example_llama2_eng}
\end{table*}

\begin{table*}[h]
    \centering
    \small
    \adjustbox{max width=\textwidth}{
    \begin{tabular}{p{\textwidth}}
        \toprule[1.1pt]

        \textbf{Description}: The Paper Chase follows a first-year Harvard Law student as he struggles to balance his studies and personal life while facing the demanding expectations of his professor.
\\
\vspace{-1mm}
$p^{rec+eng}$:  If you're looking for a movie that will keep you on the edge of your seat, The Paper Chase is a great choice. It follows a first-year Harvard Law student as he navigates the pressures of his studies and personal life while facing the expectations of his professor. It's a story of \orangetext{determination} and \orangetext{resilience} in the face of adversity, and it will leave you feeling \orangetext{inspired} and \orangetext{motivated}.\\
\midrule
\textbf{Description}: Our garden's potatoes are ripe and ready, and this recipe is always the first one I go for! They're incredibly delicious!
\\
\vspace{-1mm}
$p^{rec+eng}$: This potato recipe is a \orangetext{must-try}! It's \orangetext{packed with flavor and so easy to make}. \orangetext{Plus, you don't need to add any extra salt - the ranch dressing mix does the trick}. It's a great way to get your \orangetext{family} to enjoy potatoes \orangetext{in a new way}. Plus, it's a great way to use up \orangetext{leftovers if you have any}. \orangetext{Give it a try - you won't be disappointed!}\\
        \bottomrule[1.1pt]
\end{tabular}}
\caption{Example responses to $p^{rec+eng}$ generated by {\sc GPT-3}. The additional information augmented via the
engagement-guided prompting is highlighted in \orangetext{orange}. }
    \label{tab:example_gpt_rec_eng}
\end{table*}

\begin{table*}[h]
    \centering
    \small
    \adjustbox{max width=\textwidth}{
    \begin{tabular}{p{\textwidth}}
        \toprule[1.1pt]

        \textbf{Description}: In this Cold War-era thriller, a Soviet submarine captain attempts to defect to the United States with his state-of-the-art vessel.
\\ \midrule
$p_{para}$: The description of an item is as follows: `\{description\}', paraphrase it. \\
\vspace{-1mm}
\textbf{Response:} In this suspenseful movie set during the Cold War, a captain of a Soviet submarine attempts to defect to the US with his advanced submarine.\\ \midrule
\textbf{Variant Prompt:} Summarize the given item description, `\{description\}', using different words. \\
\vspace{-1mm}
\textbf{Response:} This suspenseful movie from the Cold War period follows a Soviet submarine commander who attempts to defect to America with his advanced submarine.\\ 
        \bottomrule[1.1pt]
    \end{tabular}}
    \caption{Example responses generated by {\sc GPT-3} for $p_{para}$ and its variant that shares the same meaning but differs in word choices.}
    \label{tab:variant_basic_para}
\end{table*}

\begin{table*}[h]
    \centering
    \small
    \adjustbox{max width=\textwidth}{
    \begin{tabular}{p{\textwidth}}
        \toprule[1.1pt]

        \textbf{Description}: In this Cold War-era thriller, a Soviet submarine captain attempts to defect to the United States with his state-of-the-art vessel.
\\ \midrule
$p_{tag}$: The description of an item is as follows: `\{description\}', summarize it with tags.\\
\vspace{-1mm}
\textbf{Response:} Cold War, Soviet, submarine, defect, United States, state-of-the-art.\\ \midrule
\textbf{Variant Prompt:} Condense the provided item description `\{description\}' using tags. \\
\vspace{-1mm}
\textbf{Response:} \#ColdWar \#Thriller \#SovietSubmarine \#Defect \#US \#StateOfTheArtVessel\\ 
        \bottomrule[1.1pt]
    \end{tabular}}
    \caption{Example responses generated by {\sc GPT-3} for $p_{tag}$ and its variant that shares the same meaning but differs in word choices.}
    \label{tab:variant_basic_tag}
\end{table*}

\begin{table*}[h]
    \centering
    \small
    \adjustbox{max width=\textwidth}{
    \begin{tabular}{p{\textwidth}}
        \toprule[1.1pt]

        \textbf{Description}: In this Cold War-era thriller, a Soviet submarine captain attempts to defect to the United States with his state-of-the-art vessel.
\\ \midrule
$p_{infer}$: The description of an item is as follows: `\{description\}', what kind of emotions can it evoke? \\
\vspace{-1mm}
\textbf{Response:} This description can evoke feelings of suspense, tension, and excitement. It can also evoke feelings of patriotism and loyalty as the Soviet submarine captain attempts to defect to the United States.\\ \midrule
\textbf{Variant Prompt:} Based on the description `\{description\}', what emotions might it elicit? \\
\vspace{-1mm}
\textbf{Response:} -Tension-Suspense-Fear-Anxiety-Dread-Unease-Paranoia\\ 
        \bottomrule[1.1pt]
    \end{tabular}}
     \caption{Example responses generated by {\sc GPT-3} for $p_{infer}$ and its variant that shares the same meaning but differs in word choices.}
    \label{tab:variant_basic_infer}
\end{table*}

\begin{table*}[h]
    \centering
    \small
    \adjustbox{max width=\textwidth}{
    \begin{tabular}{p{\textwidth}}
        \toprule[1.1pt]

        \textbf{Description}: In this Cold War-era thriller, a Soviet submarine captain attempts to defect to the United States with his state-of-the-art vessel.
\\ \midrule
$p^{rec}_{para}$: The description of an item is as follows: `\{description\}', what else should I say if I want to recommend it to others? \\
\vspace{-1mm}
\textbf{Response:} This Cold War-era thriller is a gripping and suspenseful story that will keep you on the edge of your seat. It follows the daring journey of a Soviet submarine captain as he attempts to defect to the United States with his state-of-the-art vessel. With its thrilling plot and captivating characters, this movie is sure to be an unforgettable experience.\\ \midrule
\textbf{Variant Prompt:} In addition to the description `\{description\}', what other details would you include to endorse it to others? \\
\vspace{-1mm}
\textbf{Response:} This Cold War-era thriller follows the story of a Soviet submarine captain, Marko Ramius, who attempts to defect to the United States with his state-of-the-art vessel. Along the way, he must outwit and outmaneuver a relentless CIA analyst, Jack Ryan, who is determined to stop him. With intense action sequences and a gripping plot, this movie will keep you on the edge of your seat as the two men battle for control of the submarine. Featuring an all-star cast including Sean Connery, Alec Baldwin, and Scott Glenn, The Hunt for Red October is a thrilling and suspenseful ride that will leave you wanting more.\\ 

        \bottomrule[1.1pt]
    \end{tabular}}
    \caption{Example responses generated by {\sc GPT-3} for $p^{rec}_{para}$ and its variant that shares the same meaning but differs in word choices.}
    \label{tab:variant_rec_para}
\end{table*}

\begin{table*}[h]
    \centering
    \small
    \adjustbox{max width=\textwidth}{
    \begin{tabular}{p{\textwidth}}
        \toprule[1.1pt]

        \textbf{Description}: In this Cold War-era thriller, a Soviet submarine captain attempts to defect to the United States with his state-of-the-art vessel.
\\ \midrule
$p^{rec}_{tag}$: The description of an item is as follows: `\{description\}', what tags should I use if I want to recommend it to others?\\
\vspace{-1mm}
\textbf{Response:} Tags: Cold War, Thriller, Soviet Submarine, Defection, United States, State-of-the-Art Vessel\\ \midrule
\textbf{Variant Prompt:} If you aim to recommend `\{description\}' to others, what tags or terms would you use? \\
\vspace{-1mm}
\textbf{Response:} Tags/Terms: Cold War, Thriller, Soviet Submarine, Defection, United States, State-of-the-Art Vessel.\\ 
        \bottomrule[1.1pt]
    \end{tabular}}
    \caption{Example responses generated by {\sc GPT-3} for $p^{rec}_{tag}$ and its variant that shares the same meaning but differs in word choices.}
    \label{tab:variant_rec_tag}
\end{table*}

\begin{table*}[h]
    \centering
    \small
    \adjustbox{max width=\textwidth}{
    \begin{tabular}{p{\textwidth}}
        \toprule[1.1pt]

        \textbf{Description}: In this Cold War-era thriller, a Soviet submarine captain attempts to defect to the United States with his state-of-the-art vessel.
\\ \midrule
$p^{rec}_{infer}$: The description of an item is as follows: `\{description\}', recommend it to others with a focus on the emotions it can evoke.\\
\vspace{-1mm}
\textbf{Response:} This Cold War-era thriller will take you on an intense journey of suspense and intrigue. Follow a Soviet submarine captain as he attempts to defect to the United States with his state-of-the-art vessel, and experience the tension and drama of a high-stakes game of cat and mouse. Feel the fear and anticipation as the captain navigates a dangerous course of political and personal risk.\\ \midrule
\textbf{Variant Prompt:} Using the description `\{description\}', endorse this item to others while emphasizing the emotions it can inspire. \\
\vspace{-1mm}
\textbf{Response:} This Cold War-era thriller will have you on the edge of your seat! Follow the gripping story of a Soviet submarine captain as he attempts to defect to the United States with his state-of-the-art vessel. Feel the tension and suspense as you watch this thrilling tale unfold. Don't miss out on this captivating story of courage and determination!\\ 
        \bottomrule[1.1pt]
    \end{tabular}}
     \caption{Example responses generated by {\sc GPT-3} for $p^{rec}_{infer}$ and its variants that shares the same meaning but differs in word choices.}
    \label{tab:variant_rec_infer}
\end{table*}

\begin{table*}[h]
    \centering
    \small
    \adjustbox{max width=\textwidth}{
    \begin{tabular}{p{\textwidth}}
        \toprule[1.1pt]

        \textbf{Description}: The Paper Chase follows a first-year Harvard Law student as he struggles to balance his studies and personal life while facing the demanding expectations of his professor.\\
        \vspace{-1mm}
        \textbf{Descriptions of other important neighbors}: `The Paper Chase follows a first-year Harvard Law student as he struggles to balance his studies and personal life while facing the demanding expectations of his professor.'; `The Shawshank Redemption is a story of hope, friendship, and resilience in the face of despair.'; `A criminal fakes insanity to escape prison, only to find himself in a mental institution where he must battle the oppressive head nurse.'
\\ \midrule
$p^{eng}$: Summarize the commonalities among the following descriptions: `\{description\}'; `\{descriptions of other important neighbors\}'.\\
\vspace{-1mm}
\textbf{Response:} The commonalities among the descriptions are that they all involve characters facing difficult challenges and having to find a way to overcome them. In The Paper Chase, the protagonist must balance his studies and personal life while facing the demanding expectations of his professor. In The Shawshank Redemption, the characters must find hope, friendship, and resilience in the face of despair. In the other description, the criminal must fake insanity to escape prison and then battle the oppressive head nurse in the mental institution.\\ \midrule
\textbf{Variant Prompt:}  Outline the shared characteristics of the following descriptions: `\{description\}' and `\{descriptions of other important neighbors\}'.\\
\vspace{-1mm}
\textbf{Response:} The shared characteristics of the descriptions are: - A protagonist facing a difficult situation - A struggle to balance personal life and studies - Overcoming adversity - Hope, friendship, and resilience - Oppressive forces to be overcome.\\ 

        \bottomrule[1.1pt]
    \end{tabular}}
    \caption{Example responses generated by {\sc GPT-3} for $p^{eng}$ and its variant that shares the same meaning but differs in word choices.}
    \label{tab:variant_eng}
\end{table*}

\begin{table*}[h]
    \centering
    \small
    \adjustbox{max width=\textwidth}{
    \begin{tabular}{p{\textwidth}}
        \toprule[1.1pt]

        \textbf{Description}: The Paper Chase follows a first-year Harvard Law student as he struggles to balance his studies and personal life while facing the demanding expectations of his professor.\\
        \vspace{-1mm}
        \textbf{Descriptions of other important neighbors}: `The Paper Chase follows a first-year Harvard Law student as he struggles to balance his studies and personal life while facing the demanding expectations of his professor.'; `The Shawshank Redemption is a story of hope, friendship, and resilience in the face of despair.'; `A criminal fakes insanity to escape prison, only to find himself in a mental institution where he must battle the oppressive head nurse.'
\\ \midrule
$p^{rec+eng}$: The description of an item is as follows: `\{description\}'. What else should I say if I want to recommend it to others? This content is considered to hold some similar attractive characteristics as the following descriptions: `\{descriptions of other important neighbors\}'.\\
\vspace{-1mm}
\textbf{Response:} If you're looking for a movie that will keep you on the edge of your seat, The Paper Chase is a great choice. It follows a first-year Harvard Law student as he navigates the pressures of his studies and personal life while facing the expectations of his professor. It's a story of determination and resilience in the face of adversity, and it will leave you feeling inspired and motivated.\\ \midrule
\textbf{Variant Prompt:} If you wish to recommend `\{description\}' to others, what other details would you provide? This content possesses certain akin attractive attributes as the following descriptions: `\{descriptions of other important neighbors\}'. \\
\vspace{-1mm}
\textbf{Response:} The Paper Chase is a classic drama that follows the struggles of a first-year Harvard Law student as he navigates the demanding expectations of his professor while trying to balance his studies and personal life. It is a story of perseverance, determination, and resilience in the face of adversity. The film features an all-star cast, including John Houseman, Timothy Bottoms, and Lindsay Wagner, and is sure to captivate viewers with its gripping plot and powerful performances.\\
        \bottomrule[1.1pt]
    \end{tabular}}
    \caption{Example responses generated by {\sc GPT-3} for $p^{rec+eng}$ and its variant that shares the same meaning but differs in word choices.}
    \label{tab:variant_rec_eng}
\end{table*}

\begin{table*}[h]
    \centering
    \small
    \resizebox{\textwidth}{!}{
    \begin{tabular}{lcccccc}
    \toprule
    & \multicolumn{3}{c}{\textbf{Movielens-1M}}   & \multicolumn{3}{c}{\textbf{Recipe}}  \\ 
      & Precision@10  & Recall@10      & NDCG@10       & Precision@10      & Recall@10    & NDCG@10         \\ \midrule
      Original Item Description & 0.2922 \scriptsize{$\textcolor{gray}{\pm 0.0019}$} & 0.2455 \scriptsize{$\textcolor{gray}{\pm 0.0031}$}& 0.3640 \scriptsize{$\textcolor{gray}{\pm 0.0039}$} & 0.0325 \scriptsize{$\textcolor{gray}{\pm 0.0021}$}& 0.0684 \scriptsize{$\textcolor{gray}{\pm 0.0066}$}& 0.0580 \scriptsize{$\textcolor{gray}{\pm 0.0054}$}\\
  Duplicating Original Description Embeddings & 0.2858 \scriptsize{$\textcolor{gray}{\pm 0.0020}$}&0.2417 \scriptsize{$\textcolor{gray}{\pm 0.0032}$}&0.3567 \scriptsize{$\textcolor{gray}{\pm 0.0043}$}& 0.0327 \scriptsize{$\textcolor{gray}{\pm 0.0030}$}&0.0694 \scriptsize{$\textcolor{gray}{\pm 0.0082}$} &0.0590 \scriptsize{$\textcolor{gray}{\pm 0.0068}$}\\
  Text Concatenation & 0.3075 \scriptsize{$\textcolor{gray}{\pm 0.0015}$}& 0.2636 \scriptsize{$\textcolor{gray}{\pm 0.0035}$}& 0.3853 \scriptsize{$\textcolor{gray}{\pm 0.0027}$} & 0.0332 \scriptsize{$\textcolor{gray}{\pm 0.0026}$}& 0.0714 \scriptsize{$\textcolor{gray}{\pm 0.0083}$}& 0.0591 \scriptsize{$\textcolor{gray}{\pm 0.0065}$} \\

             {\tt Concat-All} (ours)     & \textbf{0.3150}  \scriptsize{$\textcolor{gray}{\pm 0.0023}$} &\textbf{0.2766}  \scriptsize{$\textcolor{gray}{\pm 0.0030}$} & \textbf{0.3951}  \scriptsize{$\textcolor{gray}{\pm 0.0035}$}& \textbf{0.0394}  \scriptsize{$\textcolor{gray}{\pm 0.0033}$} & \textbf{0.0842}  \scriptsize{$\textcolor{gray}{\pm 0.0098}$}&  \textbf{0.0706}  \scriptsize{$\textcolor{gray}{\pm 0.0084}$}\\ 
             \bottomrule
\end{tabular}}
\caption{Concatenating embeddings of the responses augmented by \model outperforms concatenating duplicate embeddings of original description. It also achieves a superior performance than concatenating the raw text.}
\label{tab:concat}
\end{table*}

\begin{table*}[htbp]
    \centering
    \small
    \adjustbox{max width=0.9\textwidth}{
    \begin{tabular}{p{\textwidth}}
        \toprule[1.1pt]

        \textbf{Description}: This traditional meatloaf makes a fantastic meal, whether served hot or cold. It's perfect for slicing and enjoying cold at a picnic or in sandwiches. For a different twist, consider brushing your favorite salsa on top instead of the typical topping. A medium or hot chipotle salsa adds an especially tasty kick.
\\ \midrule
\textbf{KAR:} This easy and tasty meatloaf is a great dish to serve either hot or cold. It is perfect for a picnic or sandwiches, and can be enjoyed by everyone. 

\redtext{Ingredients: - 2 lbs ground beef - 1/2 cup bread crumbs - 1/4 cup onion, finely chopped - 1/4 cup green pepper, finely chopped - 1 egg - 1/4 cup ketchup - 1 teaspoon Worcestershire sauce - 1 teaspoon garlic powder - 1 teaspoon salt - 1/4 teaspoon black pepper} - Optional: your favorite salsa 

\redtext{Cuisine type: American}

\redtext{Cooking time: 1 hour}

\redtext{Nutritional value: High in protein, low in fat}

\redtext{Allergies: None}

\redtext{Taste preferences: Mildly seasoned with a hint of sweetness from the ketchup}

Skill level: Easy 

Occasion: Perfect for a picnic, lunch, dinner, or snack 

Cost: Inexpensive 

\redtext{Health and wellness goals: Low in fat, high in protein}

Food allure: Comfort food that is easy to make 

Reviews and ratings: Highly rated 

\redtext{Cooking equipment: Baking sheet}

Personal experience: Delicious and easy to make 

Season and weather: Perfect for any season 

\redtext{Cultural or ethical considerations: None}

Creativity and variety: Can be topped with your favorite salsa 

Simplicity: Easy to make 

Popularity and trends: A classic dish that is always popular 

\redtext{Instructions: 1. Preheat oven to 350F. 2. In a large bowl, combine ground beef, bread crumbs, onion, green pepper, egg, ketchup, Worcestershire sauce, garlic powder, salt, and pepper. Mix until all ingredients are combined. 3. Place mixture into a greased 9x5 inch loaf pan. 4. Bake for 1 hour or until internal temperature reaches 160F. 5. Optional: Brush your favorite salsa on top of the meatloaf before serving. 6. Enjoy!}
 \\

        \bottomrule[1.1pt]
    \end{tabular}}
\caption{An example response generated via knowledge augmentation prompts~\cite{xi2023towards}. The additional information is highlighted in \redtext{red}.}
    \label{tab:kar_recipe_example_complete}
\end{table*}

\begin{table*}[t]
\centering
\adjustbox{max width=\textwidth}{
\begin{tabular}{lcccccc}
\toprule[1.1pt]
&              & \textbf{Movielens-1M} &         &              & \textbf{Recipe}    &         \\
                               & Precision@10 & Recall@10    & NDCG@10 & Precision@10 & Recall@10 & NDCG@10 \\\midrule
Item Description & 0.2922 \scriptsize{$\textcolor{gray}{\pm 0.0019}$}& 0.2455 \scriptsize{$\textcolor{gray}{\pm 0.0031}$}& 0.3640 \scriptsize{$\textcolor{gray}{\pm 0.0039}$}& 0.0325 \scriptsize{$\textcolor{gray}{\pm 0.0021}$}& 0.0684 \scriptsize{$\textcolor{gray}{\pm 0.0066}$}& 0.0580 \scriptsize{$\textcolor{gray}{\pm 0.0054}$} \\\midrule
 Basic   & 0.3001  \scriptsize{$\textcolor{gray}{\pm 0.0027}$}     & 0.2569 \scriptsize{$\textcolor{gray}{\pm 0.0028}$}       & 0.3747 \scriptsize{$\textcolor{gray}{\pm 0.0042}$} & 0.0356 \scriptsize{$\textcolor{gray}{\pm 0.0024}$}       & 0.0754 \scriptsize{$\textcolor{gray}{\pm 0.0089}$}    & 0.0644  \scriptsize{$\textcolor{gray}{\pm 0.0068}$}\\
 Recommendation-driven     & 0.3025   \scriptsize{$\textcolor{gray}{\pm 0.0023}$}    & 0.2577 \scriptsize{$\textcolor{gray}{\pm 0.0053}$}       & 0.3786  \scriptsize{$\textcolor{gray}{\pm 0.0041}$}& \textbf{0.0361} \scriptsize{$\textcolor{gray}{\pm 0.0031}$}      & \textbf{0.0771} \scriptsize{$\textcolor{gray}{\pm 0.0086}$}   & \textbf{0.0649} \scriptsize{$\textcolor{gray}{\pm 0.0069}$}\\
            Engagement-guided     & 0.3036 \scriptsize{$\textcolor{gray}{\pm 0.0020}$}       & \textbf{0.2608} \scriptsize{$\textcolor{gray}{\pm 0.0030}$}      & 0.3801  \scriptsize{$\textcolor{gray}{\pm 0.0032}$} & 0.0348 \scriptsize{$\textcolor{gray}{\pm 0.0031}$}       & 0.0732 \scriptsize{$\textcolor{gray}{\pm 0.0088}$}    & 0.0628 \scriptsize{$\textcolor{gray}{\pm 0.0077}$} \\
                         
        Recommendation+Engagement & \textbf{0.3038} \scriptsize{$\textcolor{gray}{\pm 0.0020}$}       & 0.2603    \scriptsize{$\textcolor{gray}{\pm 0.0042}$}   & \textbf{0.3802}  \scriptsize{$\textcolor{gray}{\pm 0.0037}$}& 0.0349 \scriptsize{$\textcolor{gray}{\pm 0.0024}$}       & 0.0732  \scriptsize{$\textcolor{gray}{\pm 0.0066}$}   & 0.0625 \scriptsize{$\textcolor{gray}{\pm 0.0060}$}  \\
                        \bottomrule[1.1pt]       
\end{tabular}
}
\caption{Average recommendation performance among different prompting strategies across five different splits. The best performance among the three Basic Prompting and three Recommendation-driven Prompting strategies is reported. The overall best results are highlighted in \textbf{bold}.}
    \label{tab:strategy_comp}
\end{table*}

\begin{figure*}[t]
\centering
\begin{subfigure}{0.3\textwidth}
    \includegraphics[width=\textwidth]{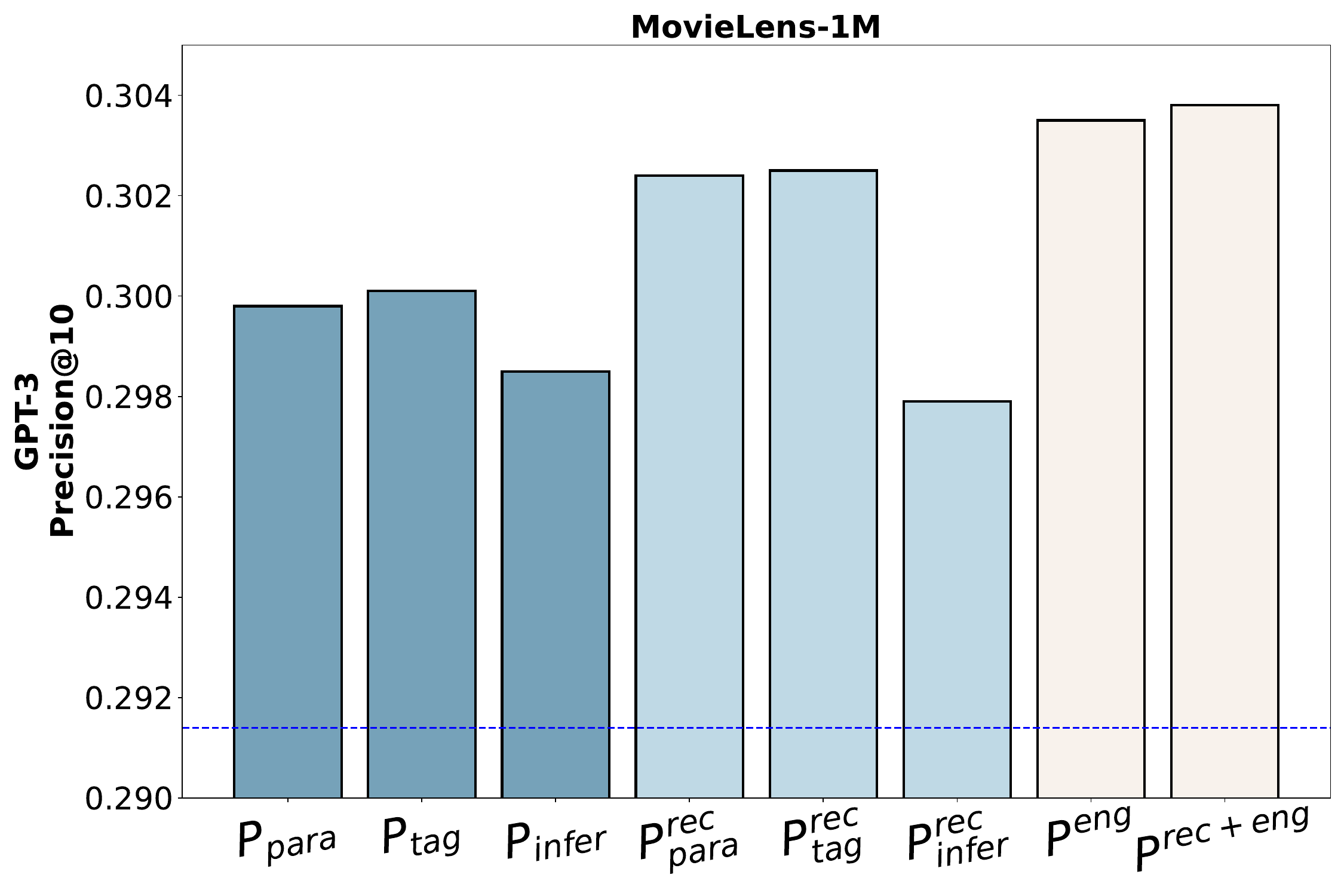}
\end{subfigure}
\hfill
\begin{subfigure}{0.3\textwidth}
    \includegraphics[width=\textwidth]{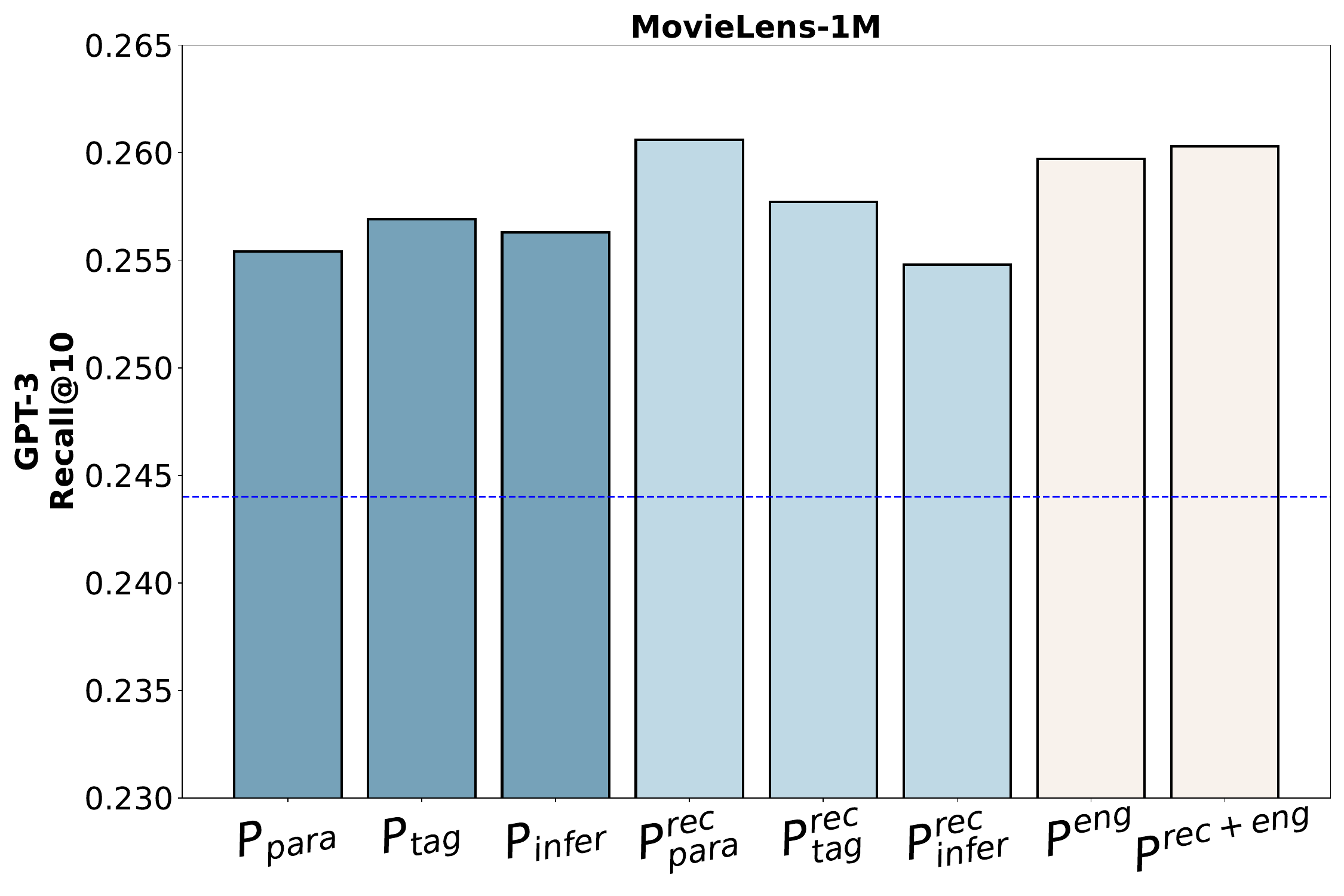}
   
\end{subfigure}
\hfill
\begin{subfigure}{0.3\textwidth}
    \includegraphics[width=\textwidth]{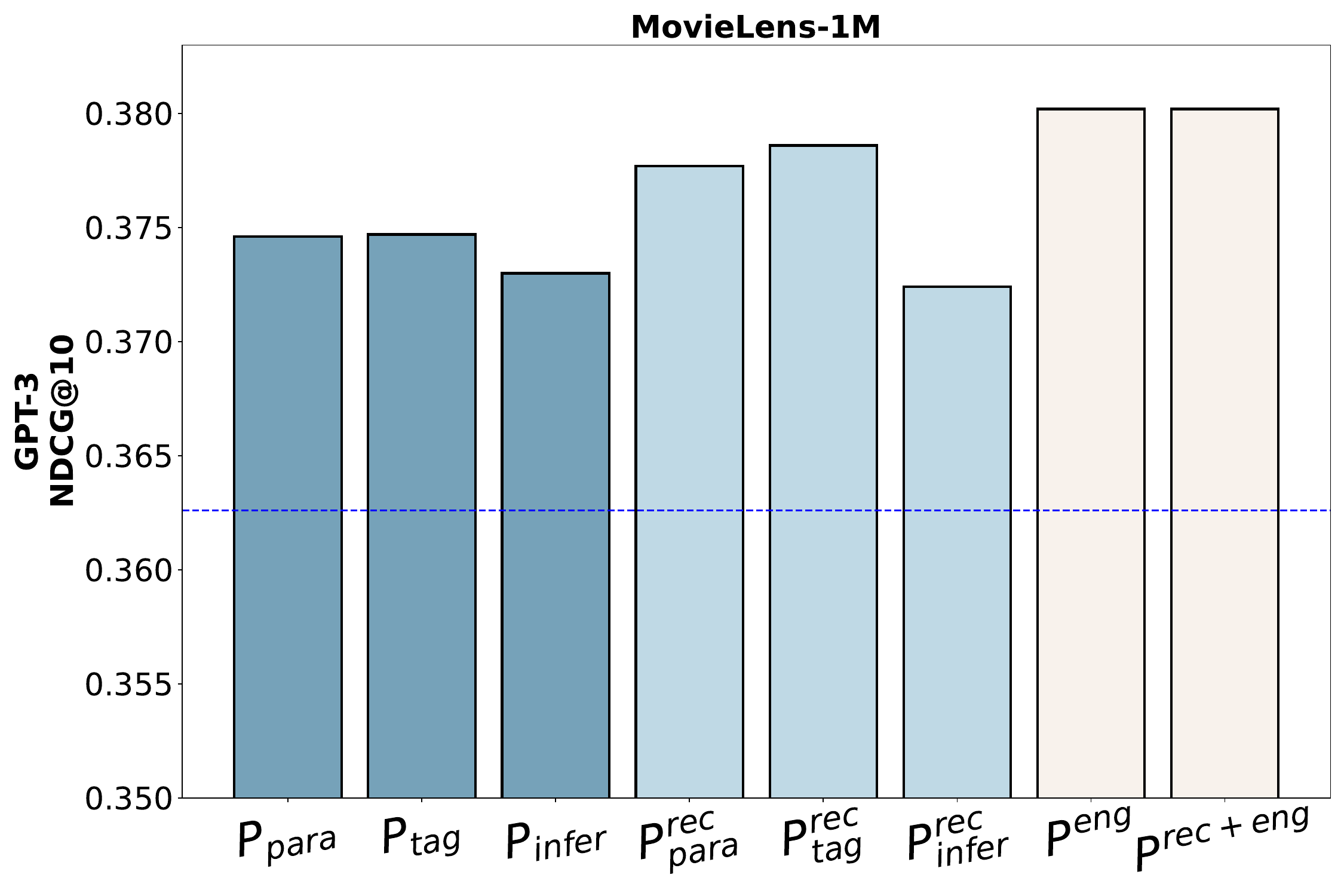}
    
\end{subfigure}
\hfill
\begin{subfigure}{0.3\textwidth}
    \includegraphics[width=\textwidth]{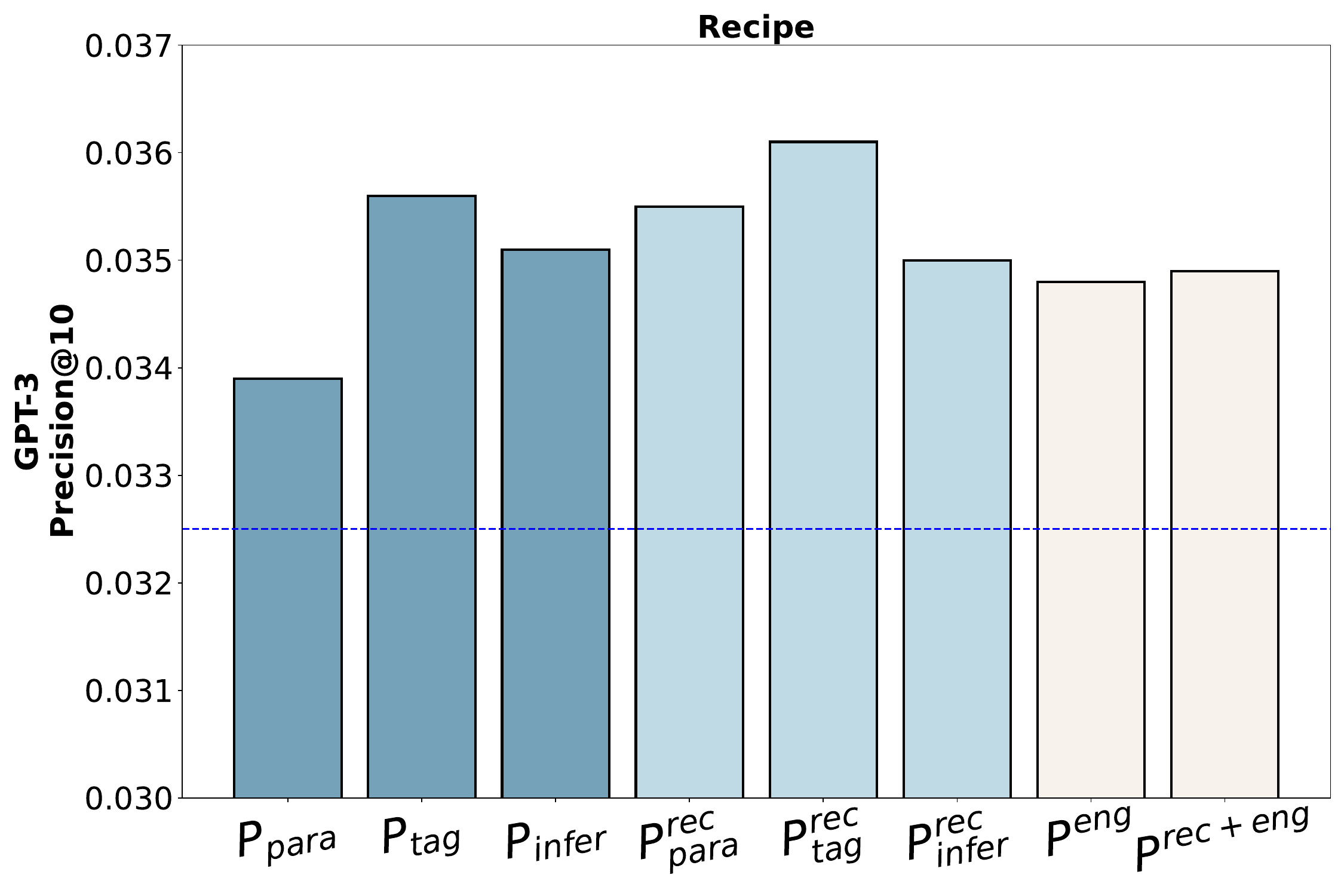}
\end{subfigure}
\hfill
\begin{subfigure}{0.3\textwidth}
    \includegraphics[width=\textwidth]{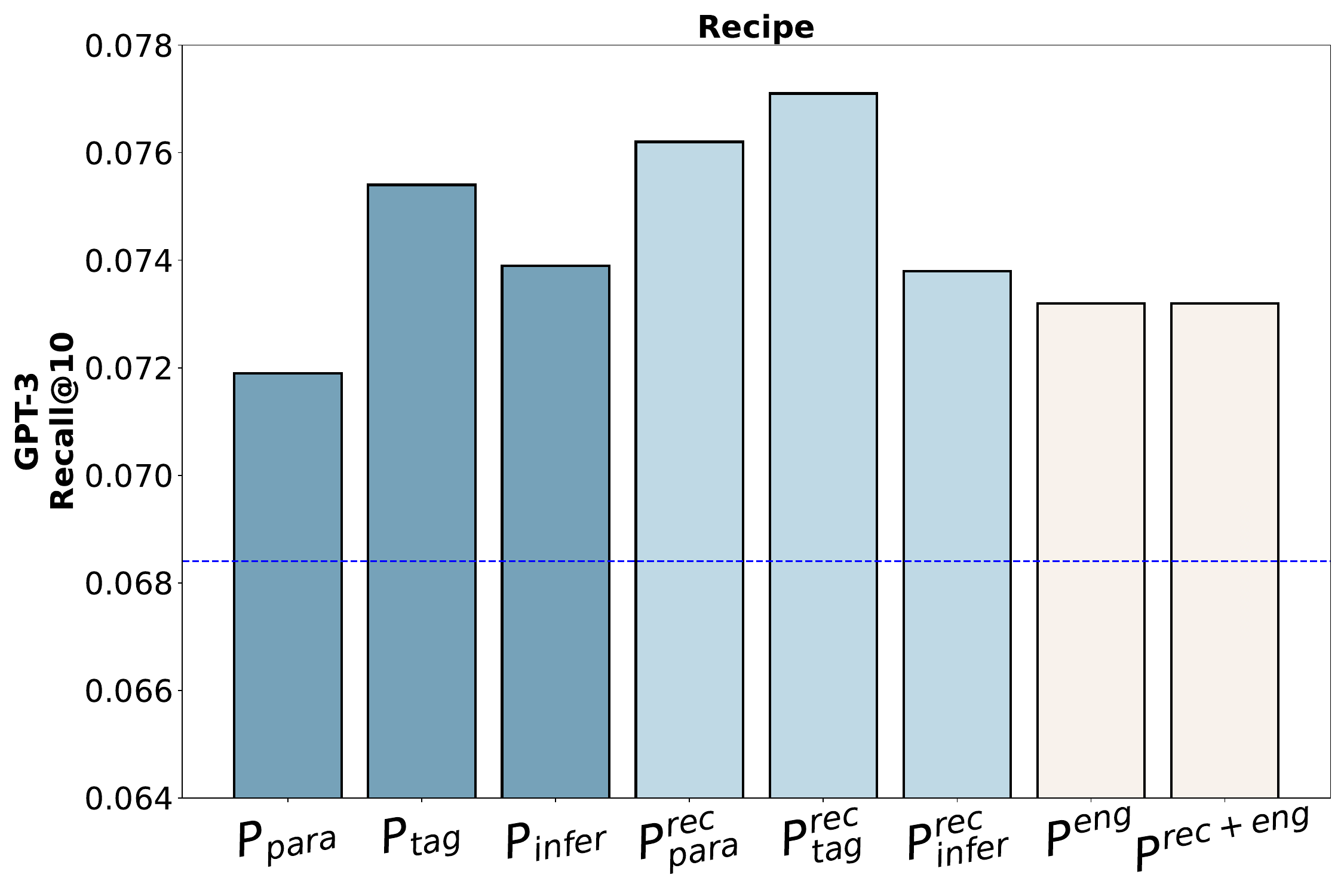}
\end{subfigure}
\hfill
\begin{subfigure}{0.3\textwidth}
    \includegraphics[width=\textwidth]{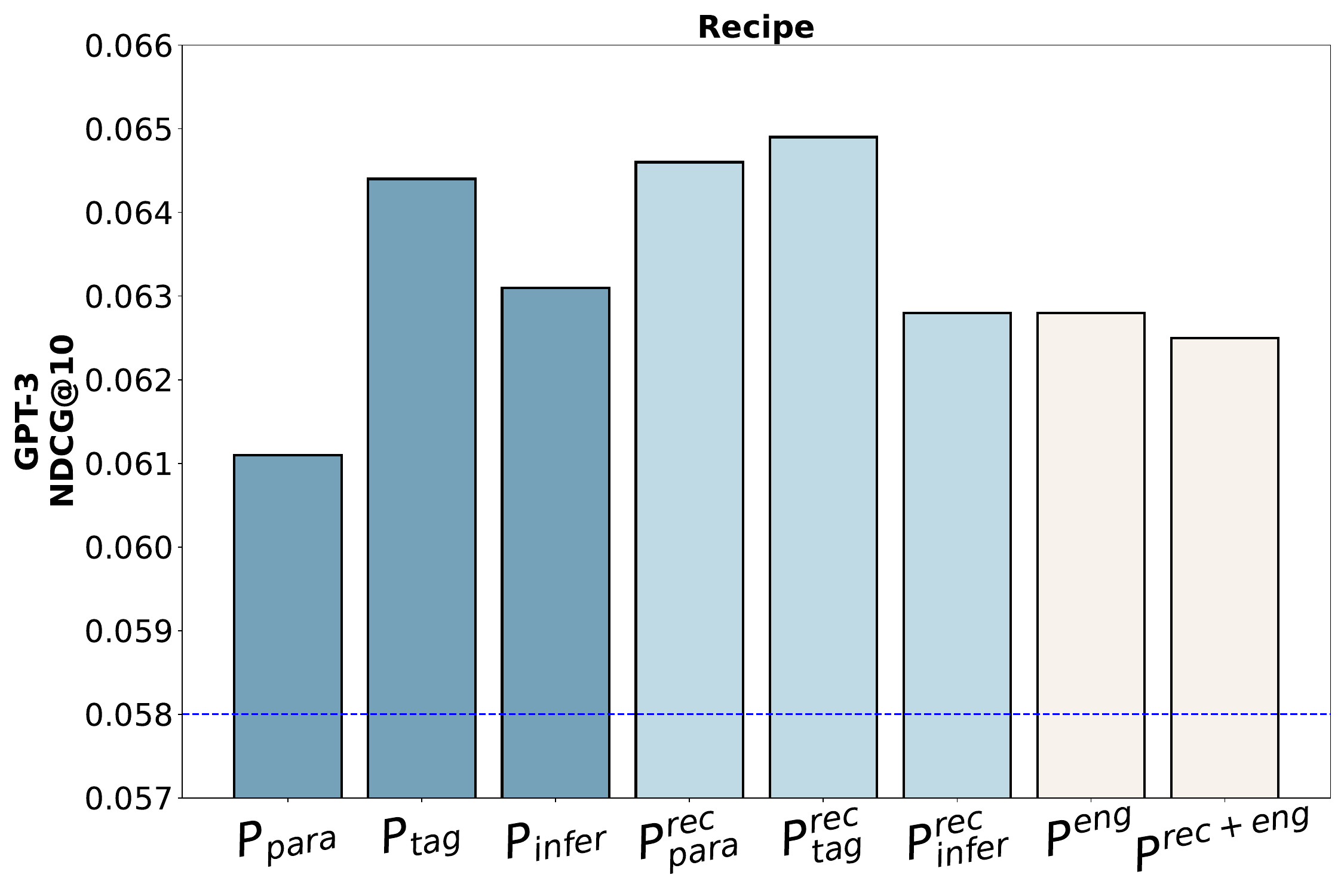}
\end{subfigure}
\hfill
\begin{subfigure}{0.3\textwidth}
    \includegraphics[width=\textwidth]{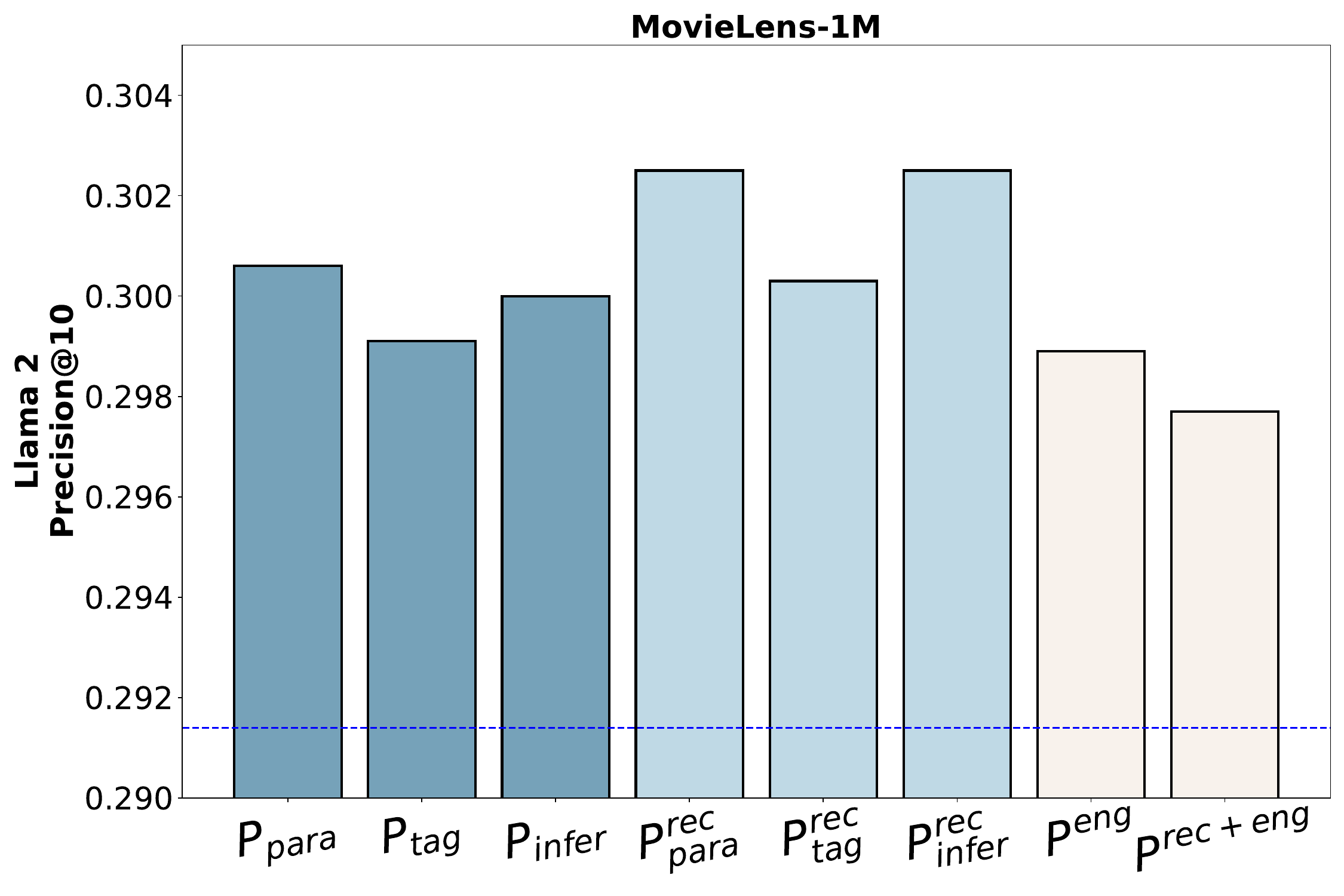}
\end{subfigure}
\hfill
\begin{subfigure}{0.3\textwidth}
    \includegraphics[width=\textwidth]{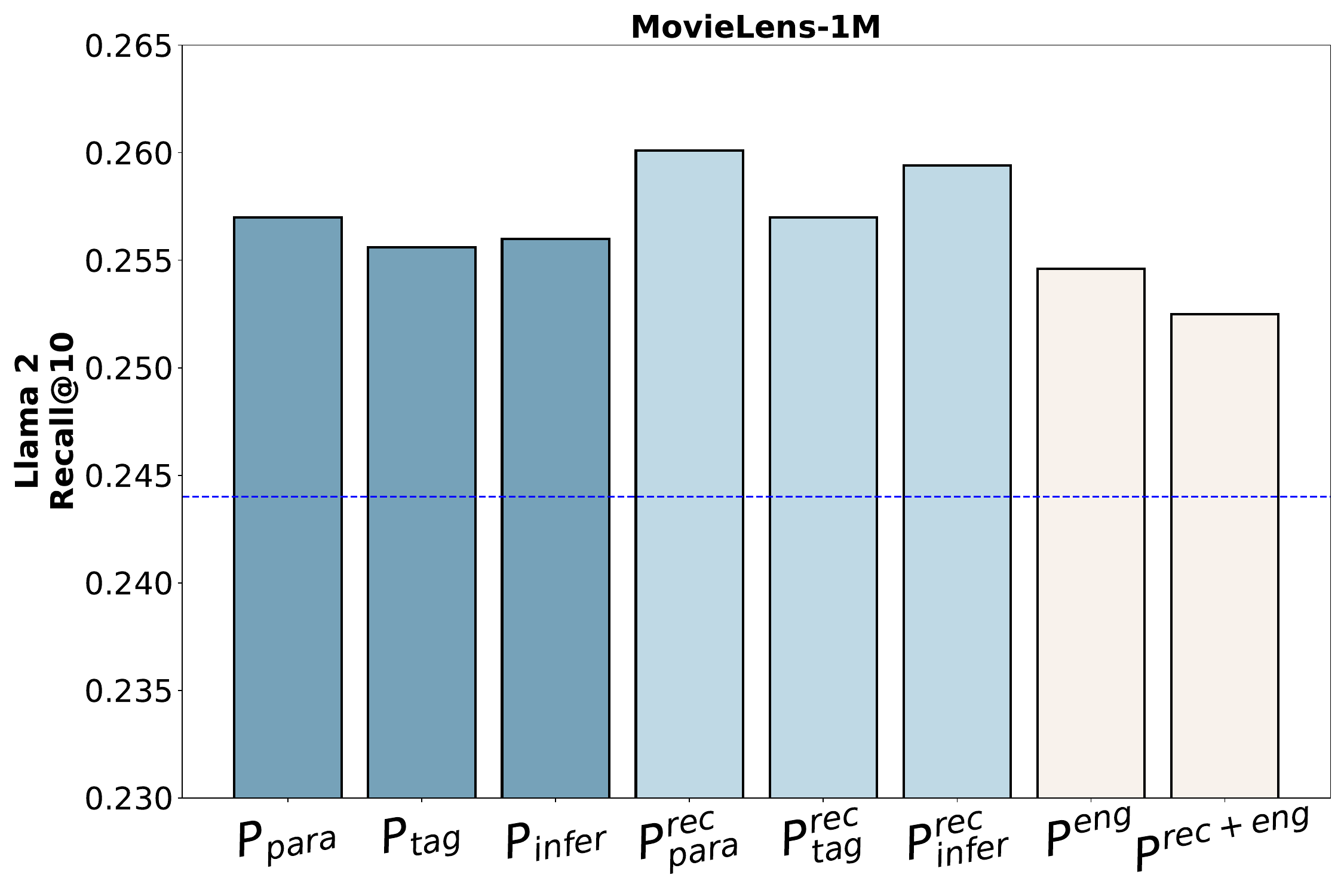}
   
\end{subfigure}
\hfill
\begin{subfigure}{0.3\textwidth}
    \includegraphics[width=\textwidth]{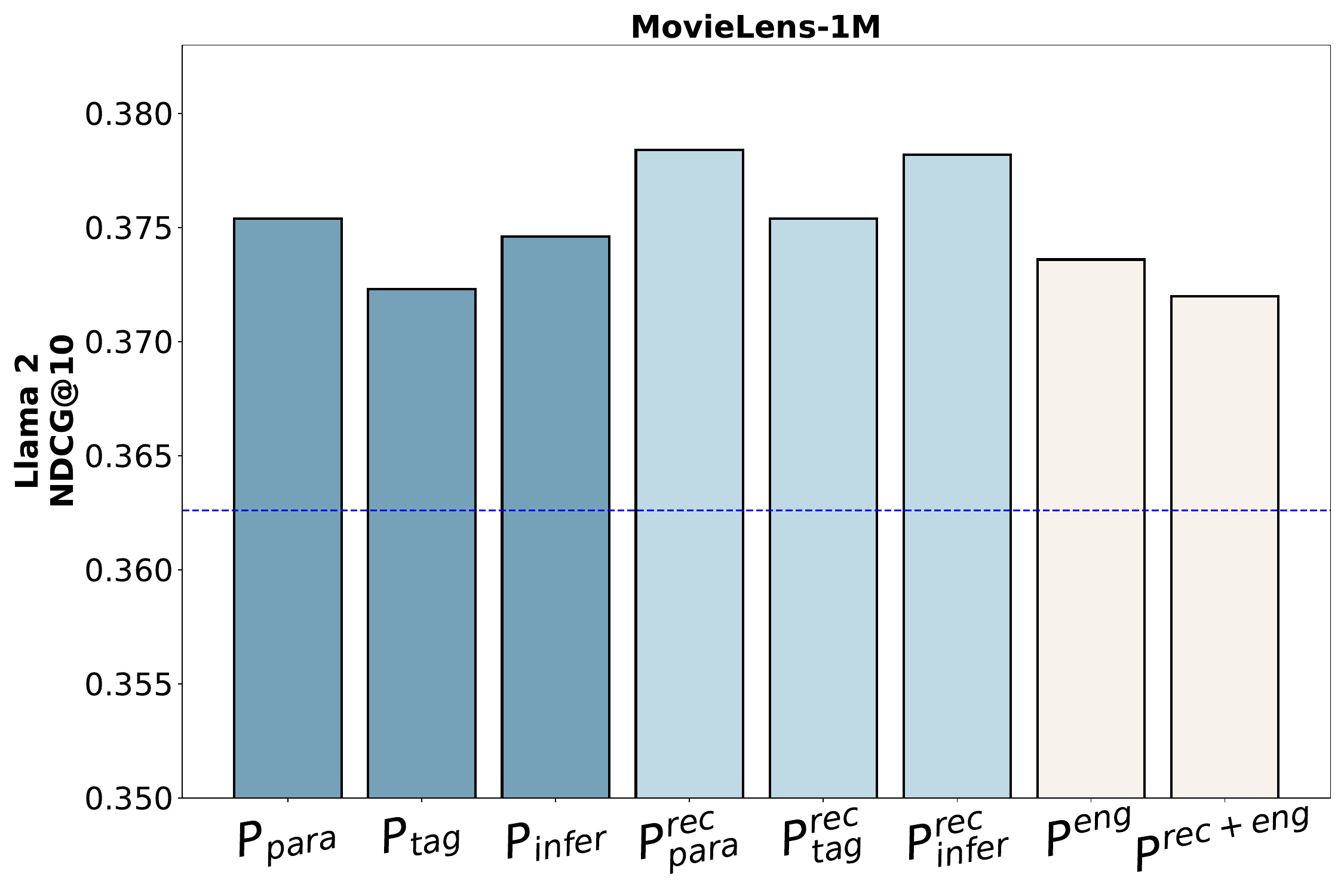}

\end{subfigure}
\hfill
\begin{subfigure}{0.3\textwidth}
    \includegraphics[width=\textwidth]{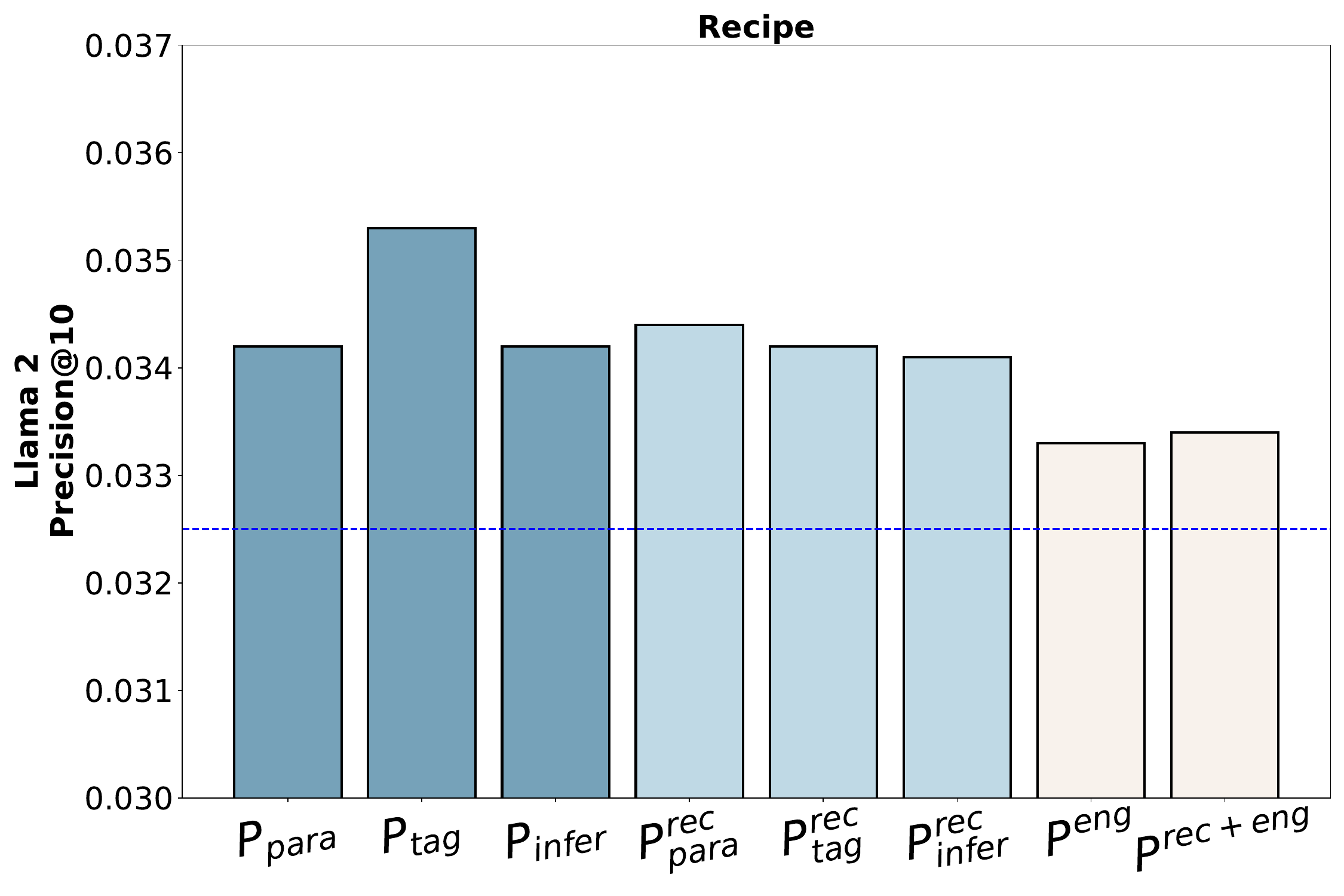}
\end{subfigure}
\hfill
\begin{subfigure}{0.3\textwidth}
    \includegraphics[width=\textwidth]{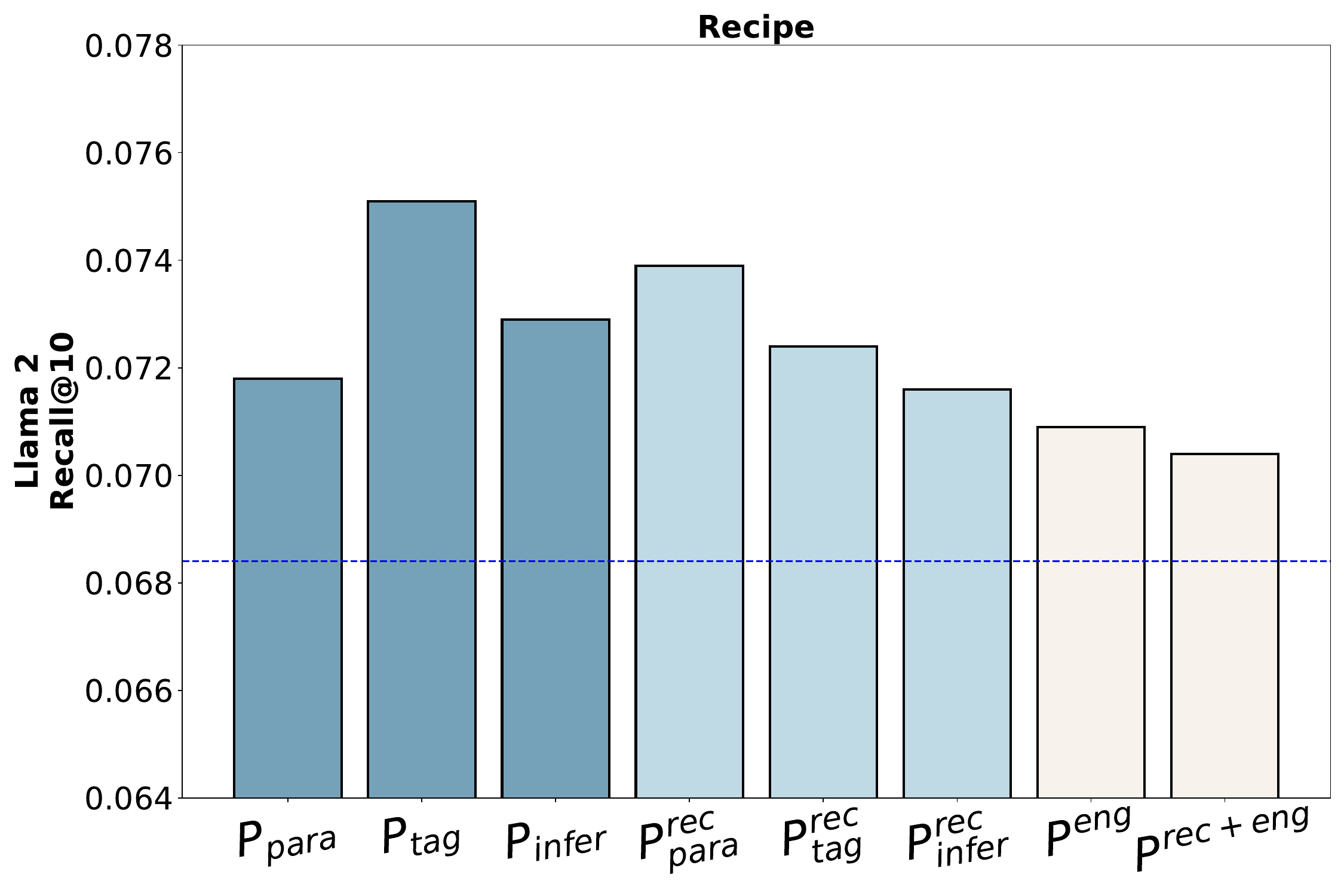}
\end{subfigure}
\hfill
\begin{subfigure}{0.3\textwidth}
    \includegraphics[width=\textwidth]{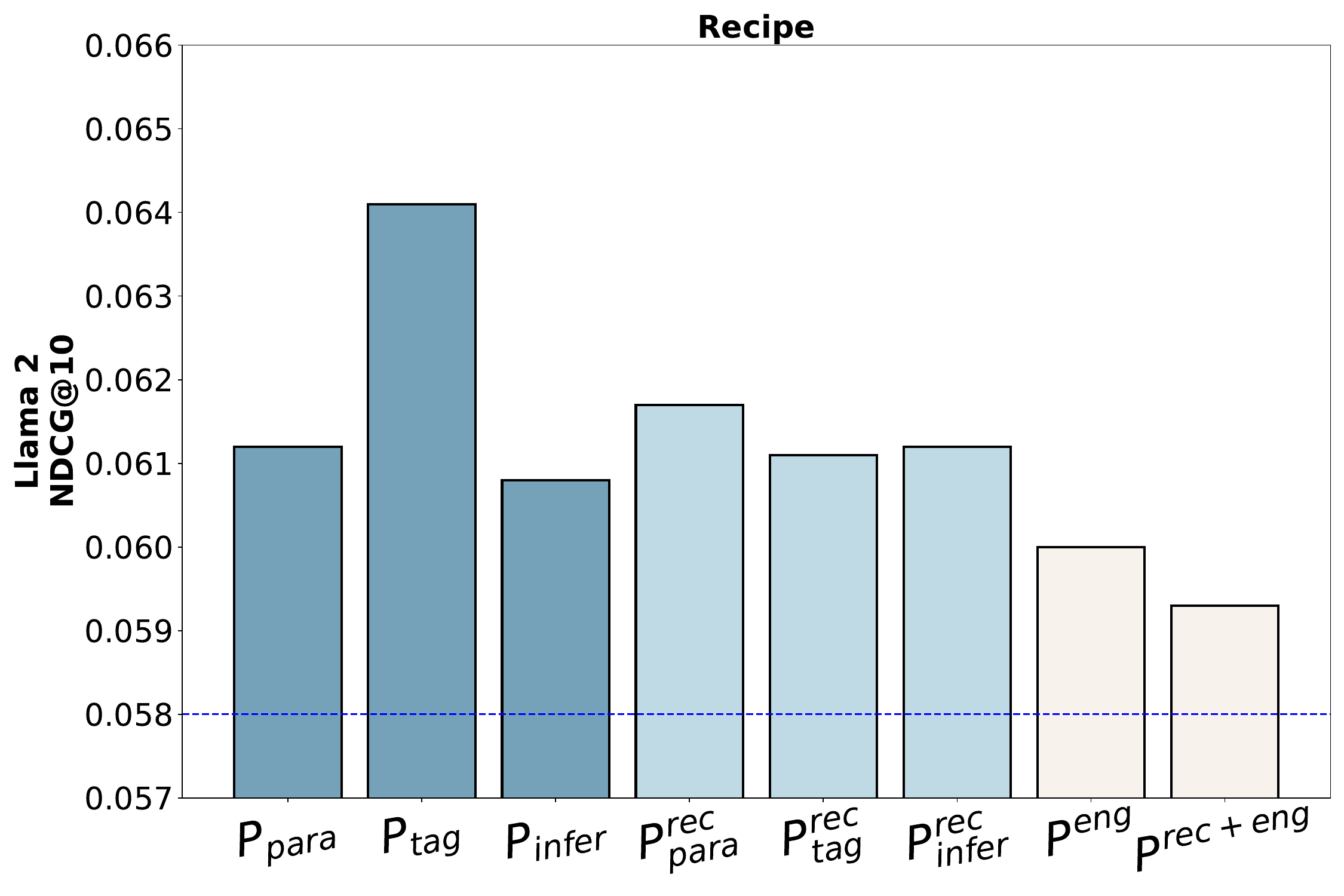}
\end{subfigure}
\caption{\textbf{The ablation study conducted on different prompting strategies shows that augmenting the input text with responses generated by large language models using our proposed prompting strategies enhances recommendation performance. However, the extent of this improvement may vary depending on the characteristics of the datasets used.} The \textit{basic prompting} strategy includes three variants: $p_{para}$ (paraphrase), $p_{tag}$ (tag), and $p_{infer}$ (infer). The \textit{recommendation-driven} versions of these three variants are denoted by $p_{para}^{rec}$, $p_{tag}^{rec}$, and $p_{infer}^{rec}$. In addition, $p^{eng}$ represents the \textit{engagement-guided} prompts, and $p^{rec\text{+}eng}$ stands for the \textit{recommendation + engagement} prompts. The \textcolor{blue}{blue line} in each figure indicates the performance achieved by using only the original description embeddings. Note that in this ablation study, all the baselines use a concatenation of the the original description embeddings and prompt response embeddings as their model input. }
\label{fig:ablation_different_strategy}
\end{figure*}

\begin{figure*}[t]
\centering
\begin{subfigure}{0.3\textwidth}
    \includegraphics[width=\textwidth]{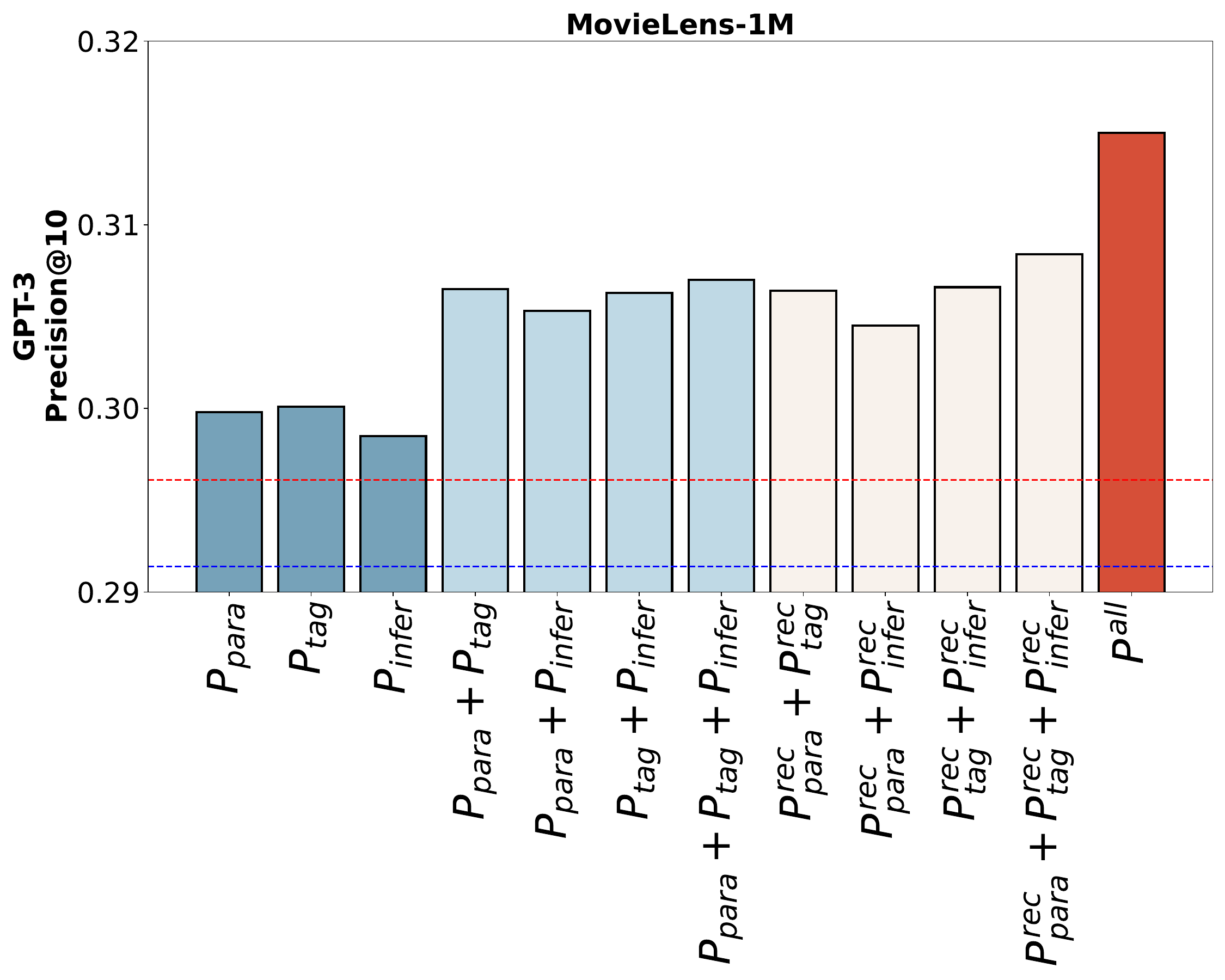}
\end{subfigure}
\hfill
\begin{subfigure}{0.3\textwidth}
    \includegraphics[width=\textwidth]{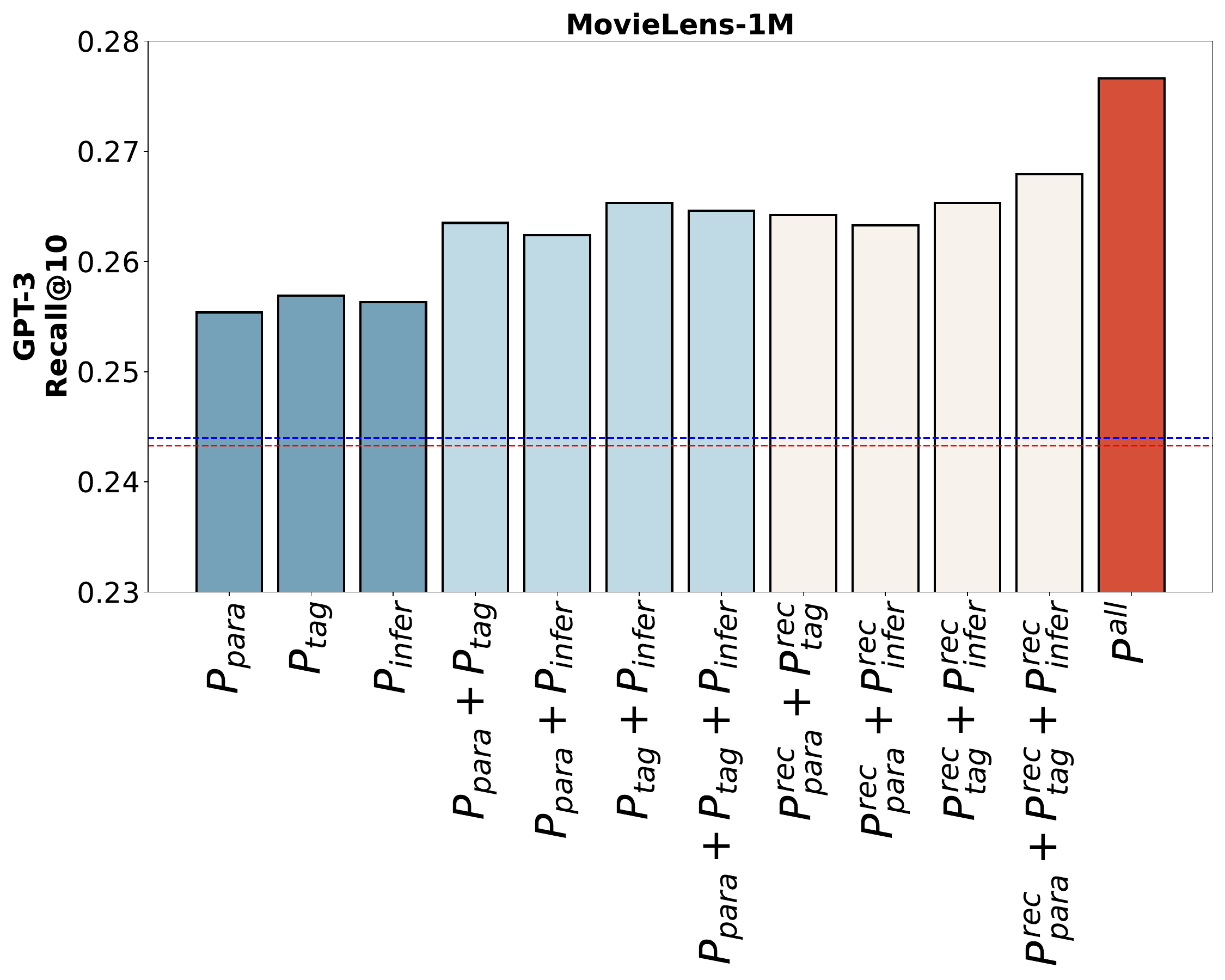}
   
\end{subfigure}
\hfill
\begin{subfigure}{0.3\textwidth}
    \includegraphics[width=\textwidth]{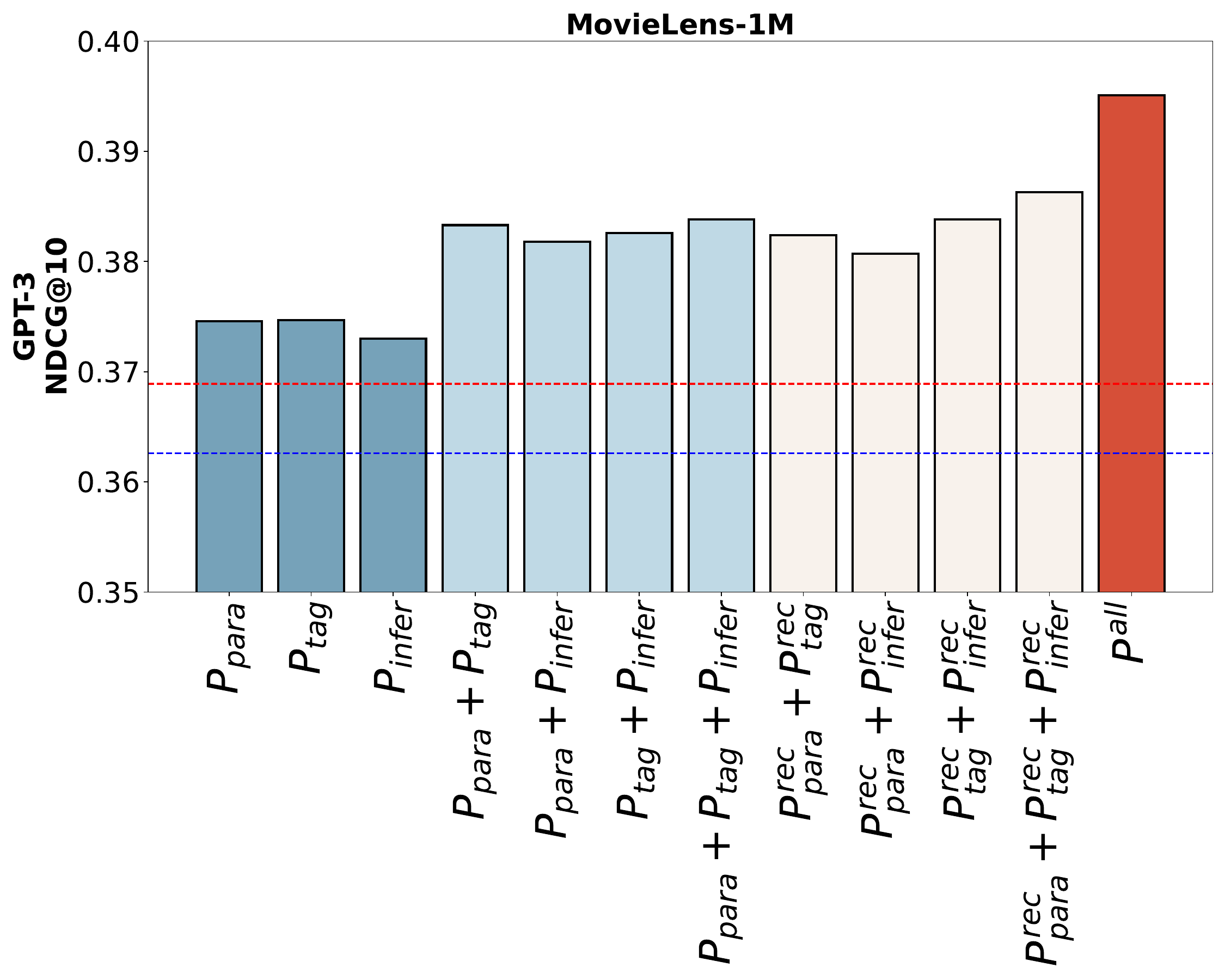}
    
\end{subfigure}
\hfill
\begin{subfigure}{0.3\textwidth}
    \includegraphics[width=\textwidth]{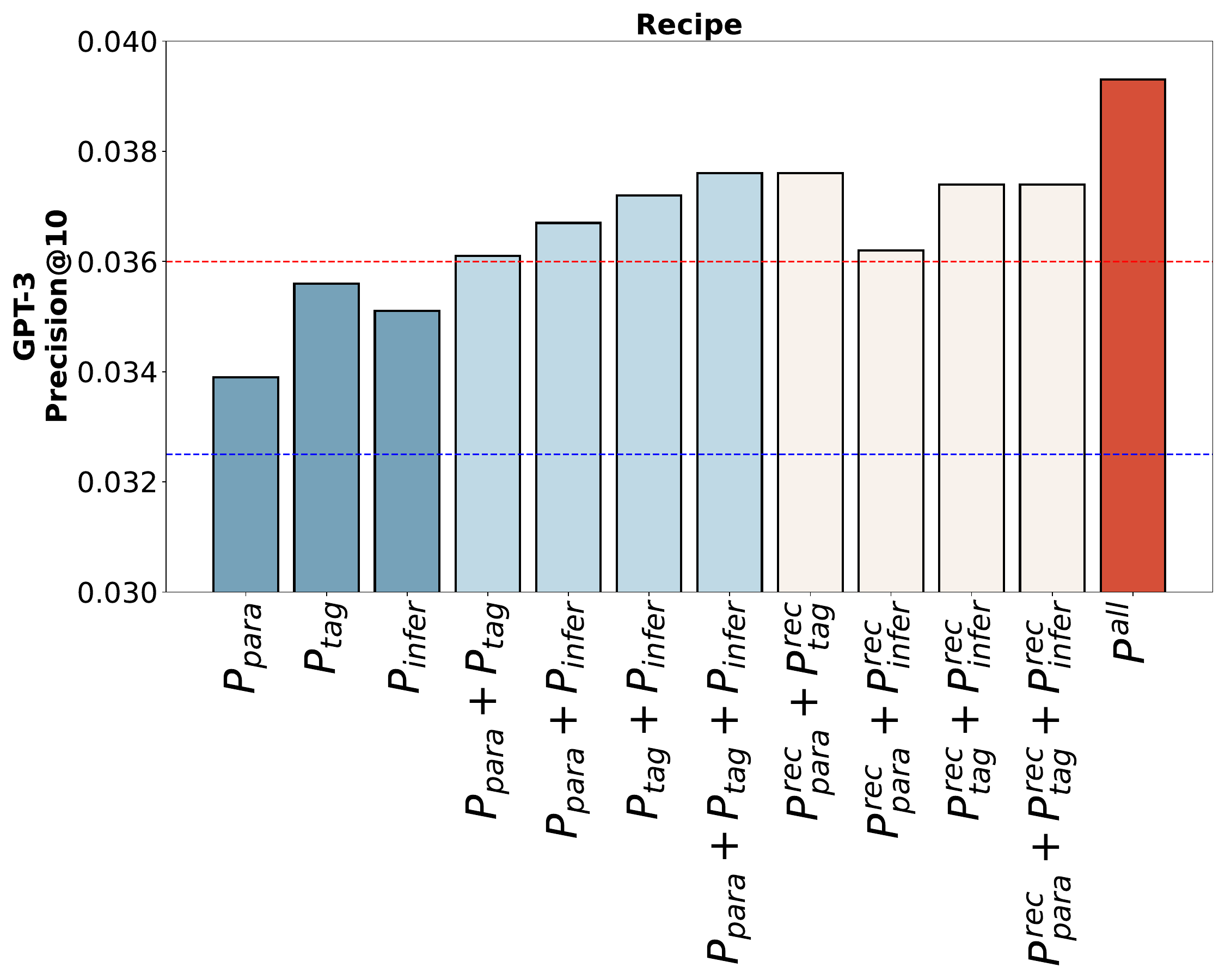}
\end{subfigure}
\hfill
\begin{subfigure}{0.3\textwidth}
    \includegraphics[width=\textwidth]{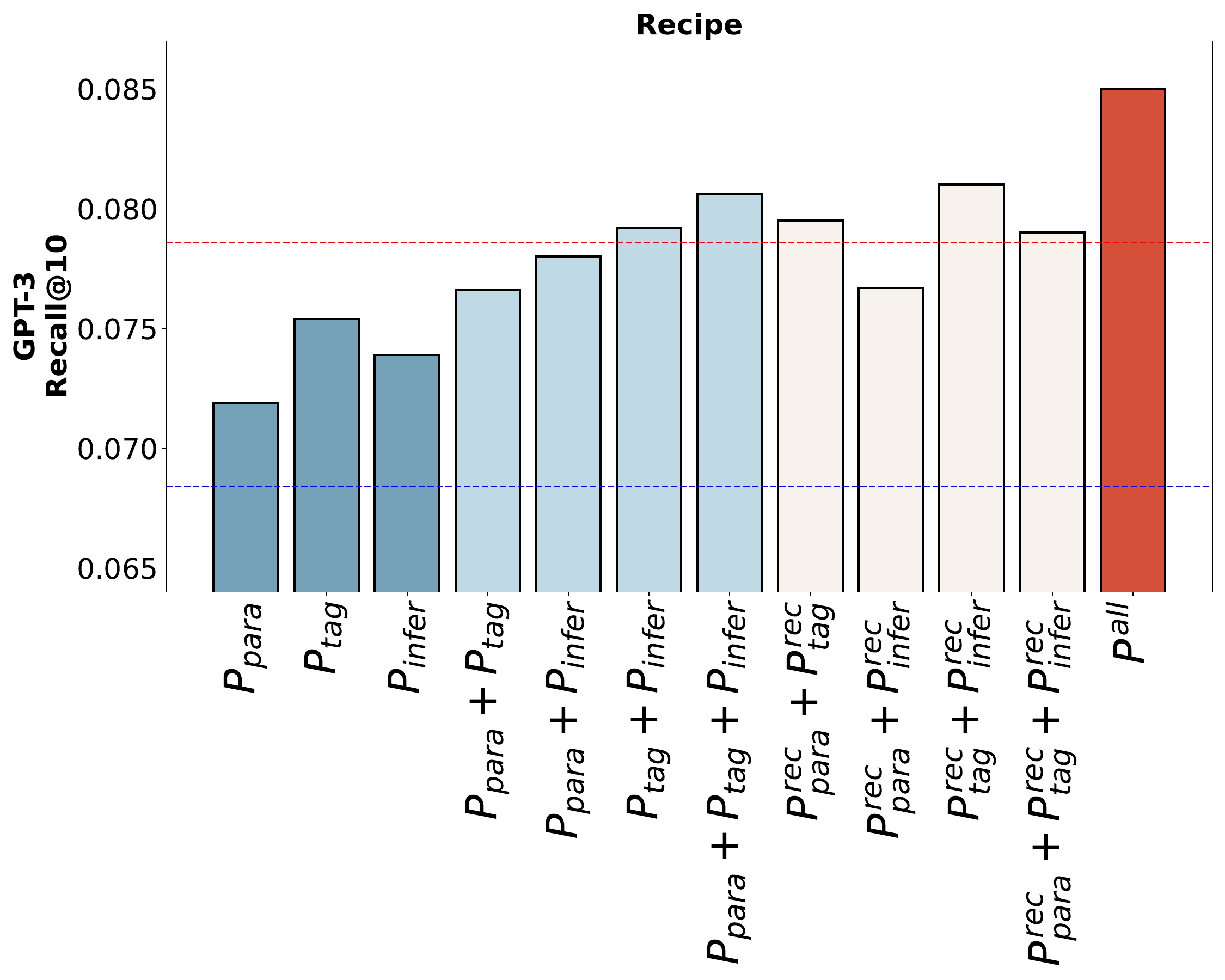}
\end{subfigure}
\hfill
\begin{subfigure}{0.3\textwidth}
    \includegraphics[width=\textwidth]{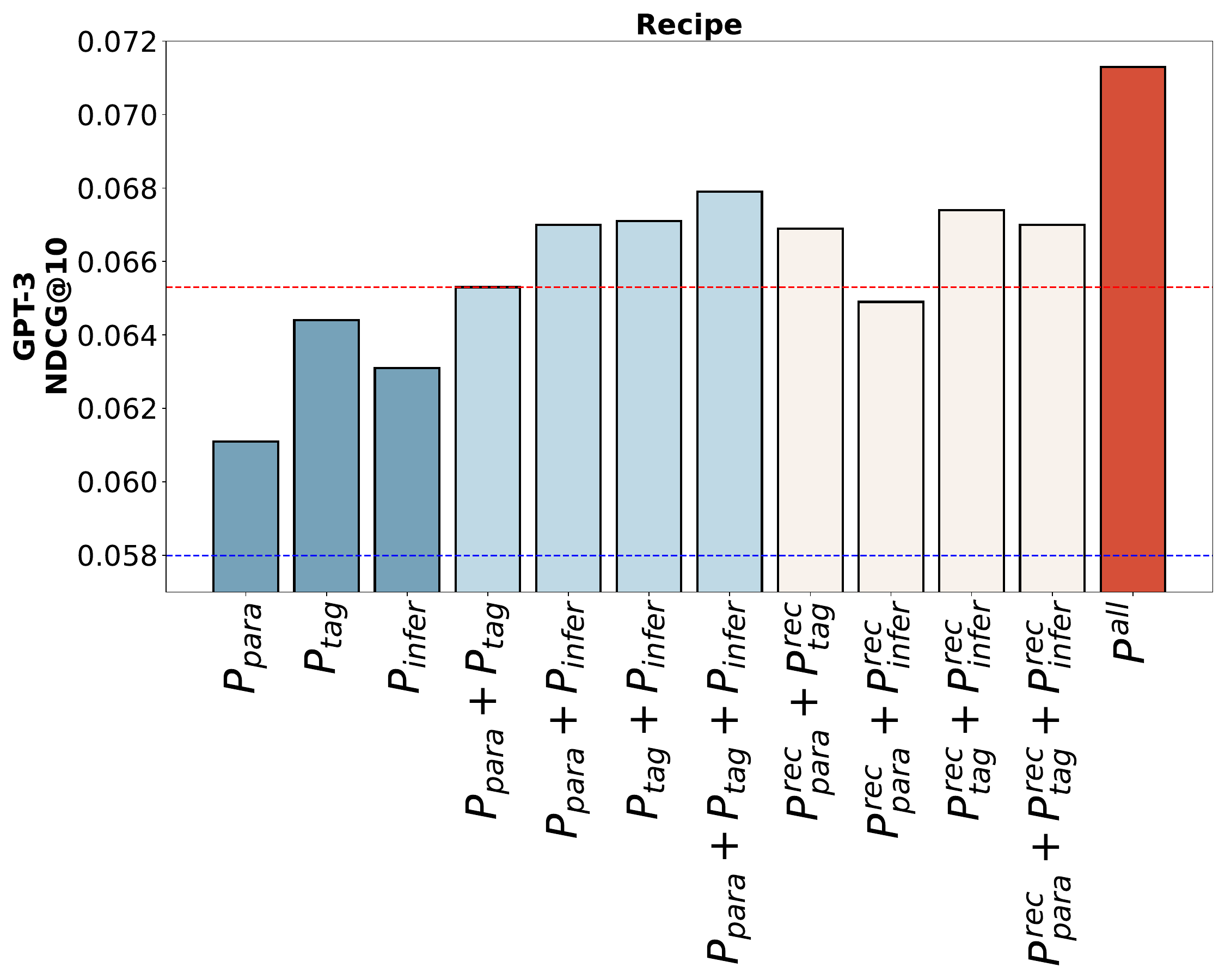}
\end{subfigure}
\hfill
\begin{subfigure}{0.3\textwidth}
    \includegraphics[width=\textwidth]{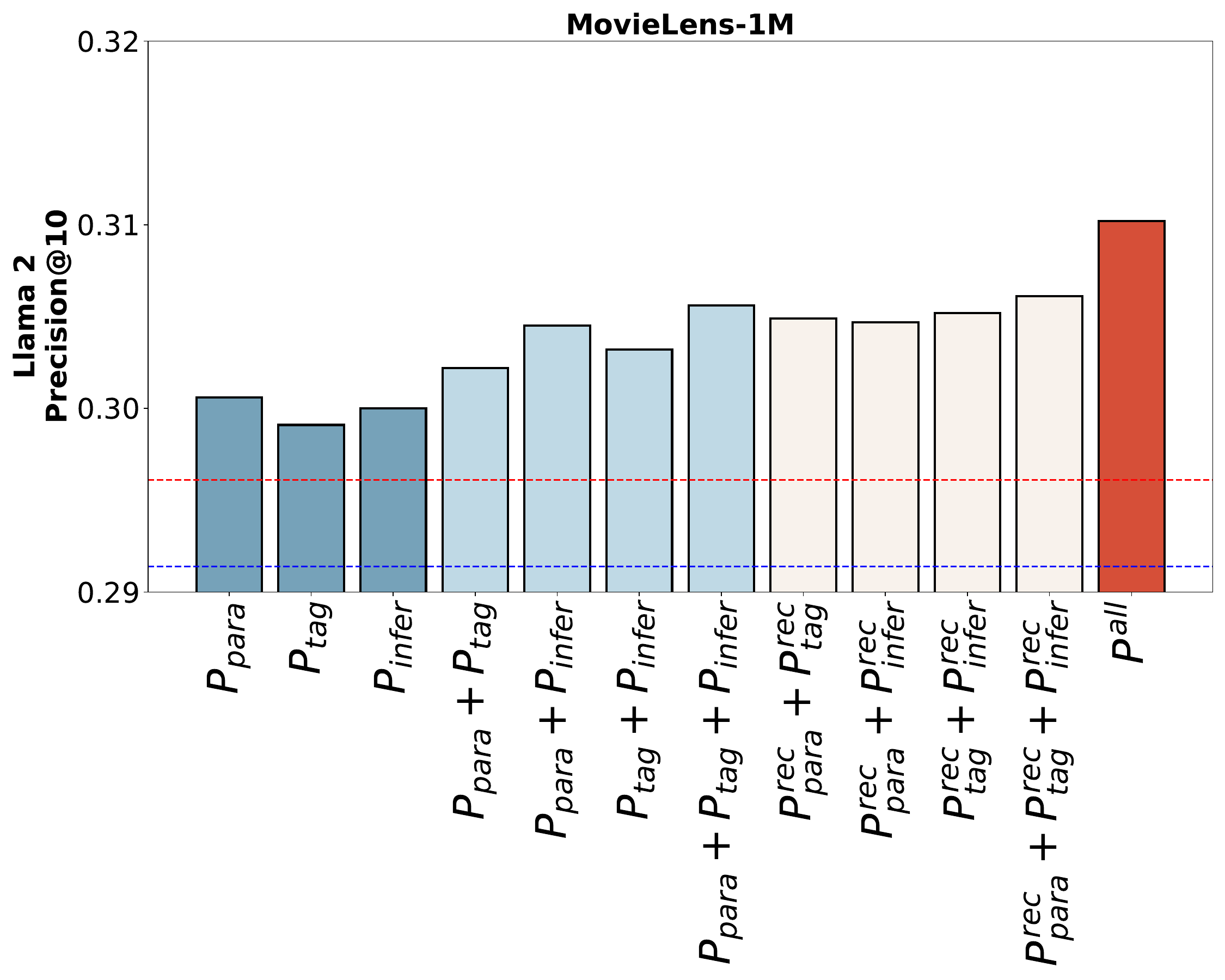}
\end{subfigure}
\hfill
\begin{subfigure}{0.3\textwidth}
    \includegraphics[width=\textwidth]{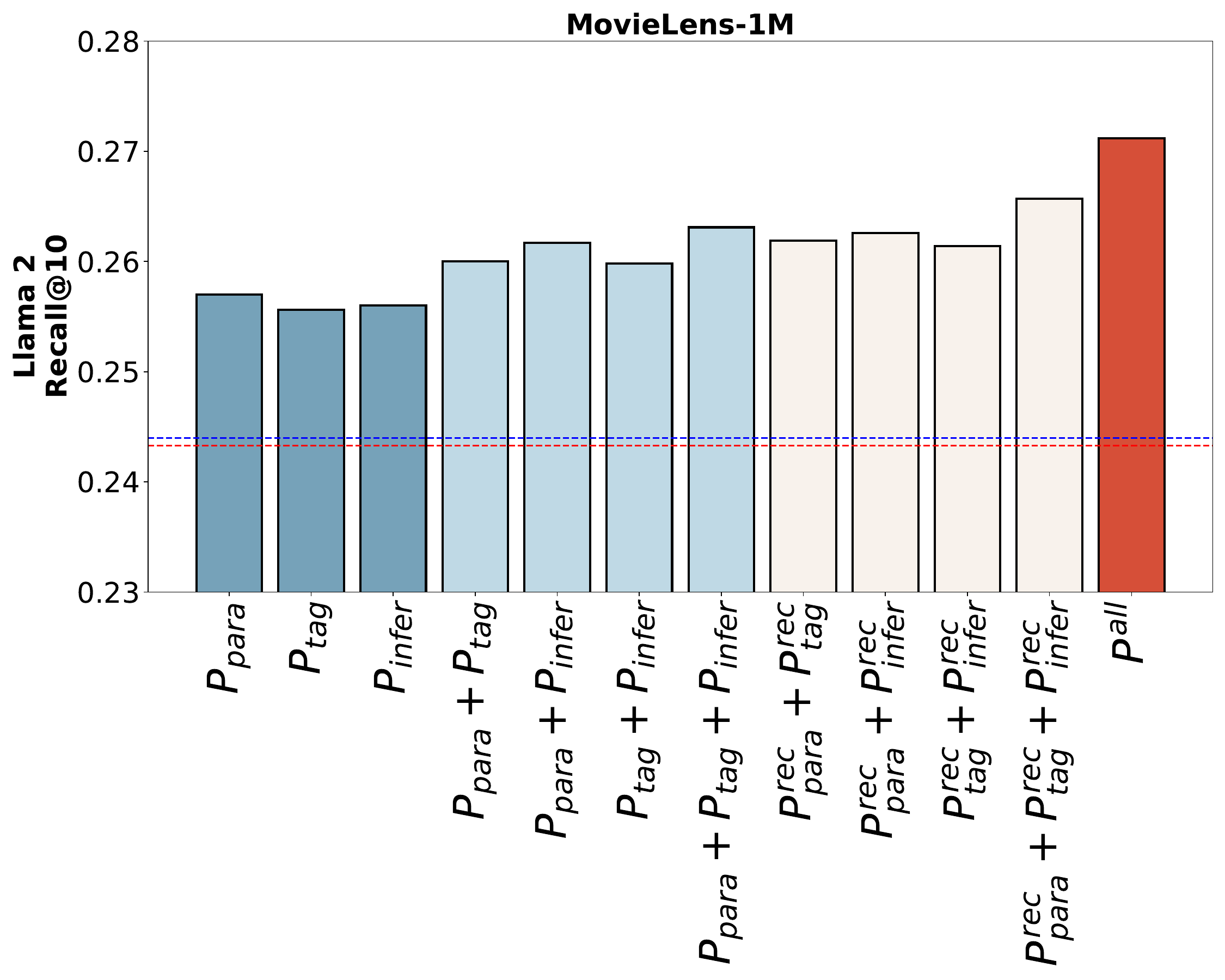}
   
\end{subfigure}
\hfill
\begin{subfigure}{0.3\textwidth}
    \includegraphics[width=\textwidth]{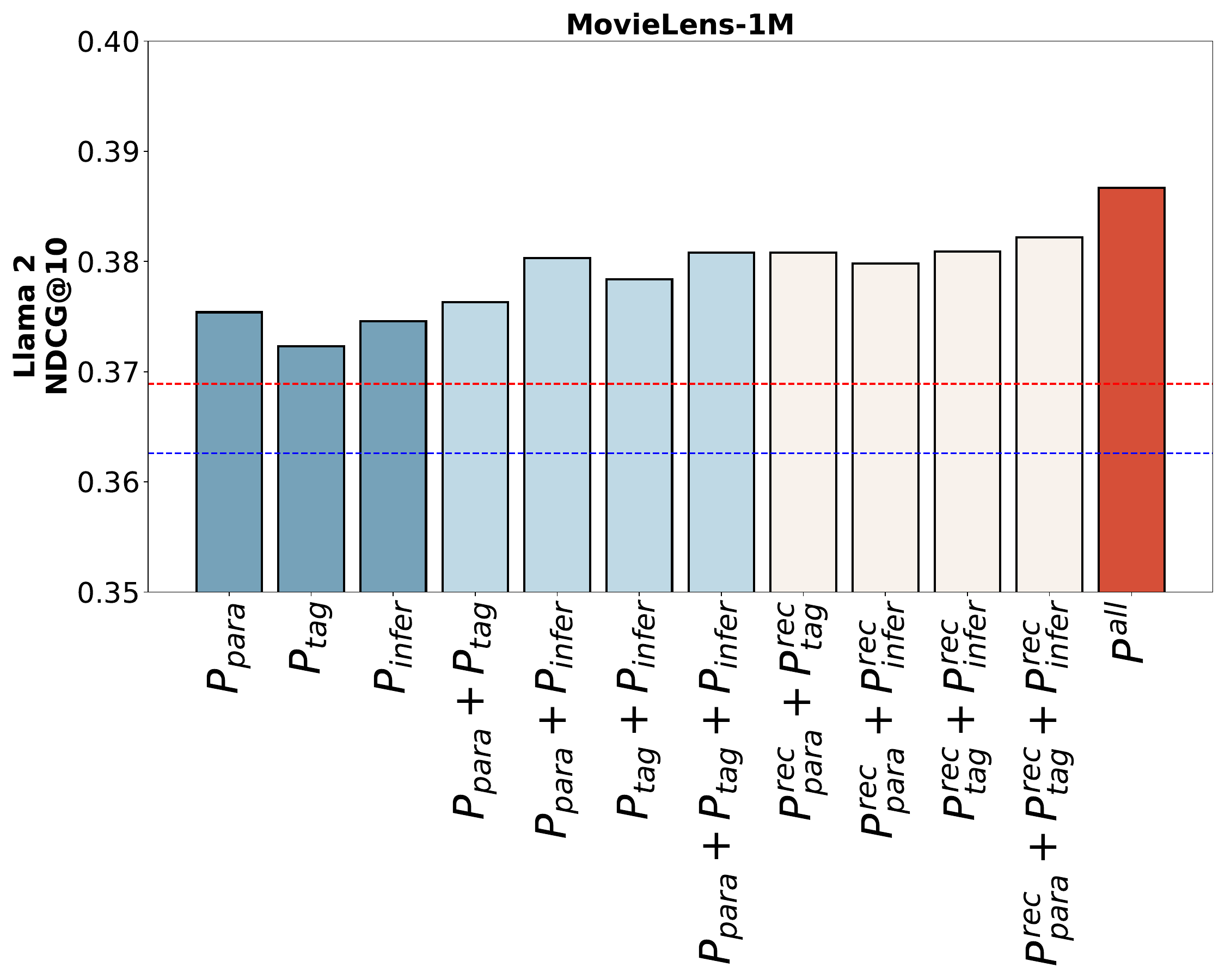}

\end{subfigure}
\hfill
\begin{subfigure}{0.3\textwidth}
    \includegraphics[width=\textwidth]{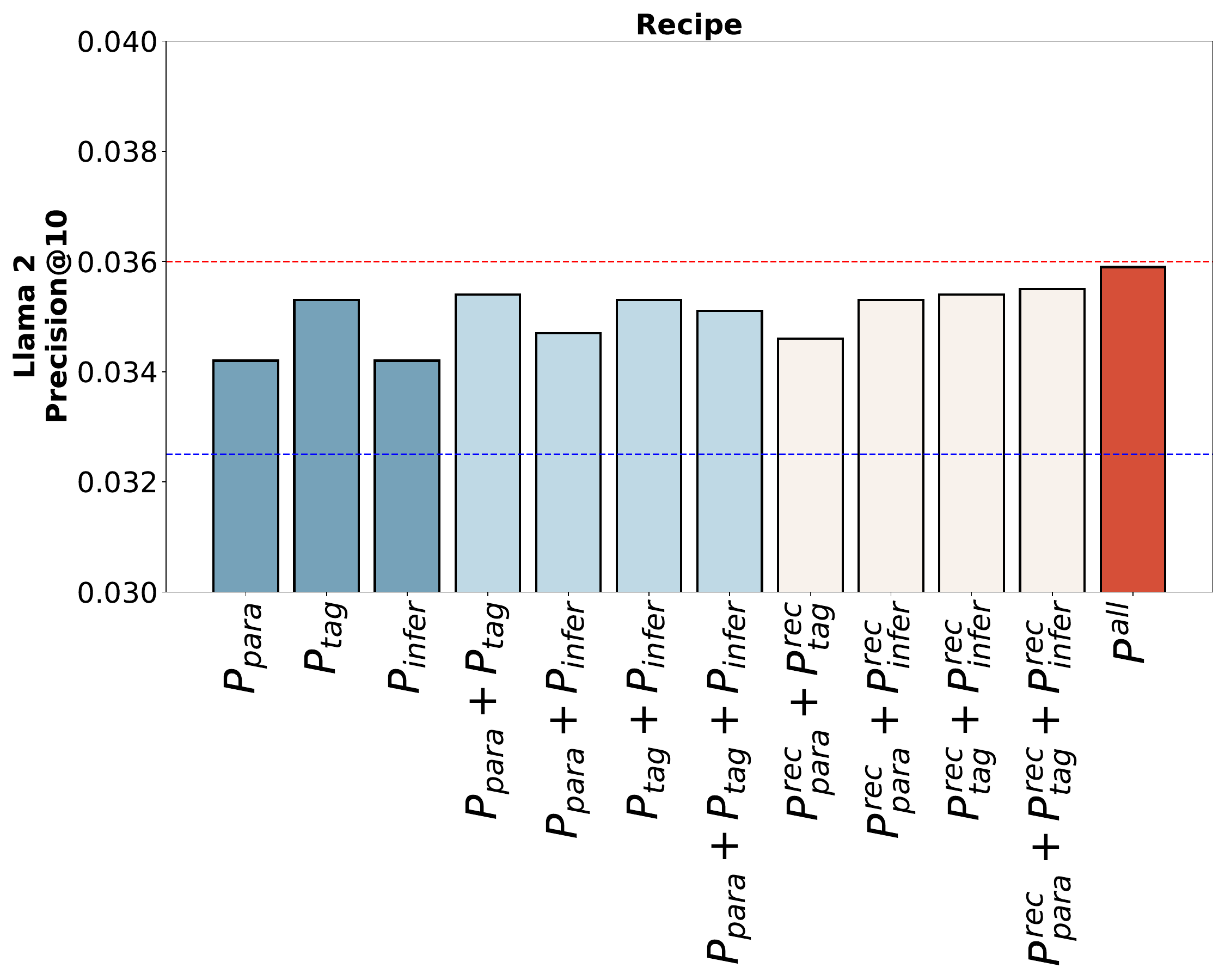}
\end{subfigure}
\hfill
\begin{subfigure}{0.3\textwidth}
    \includegraphics[width=\textwidth]{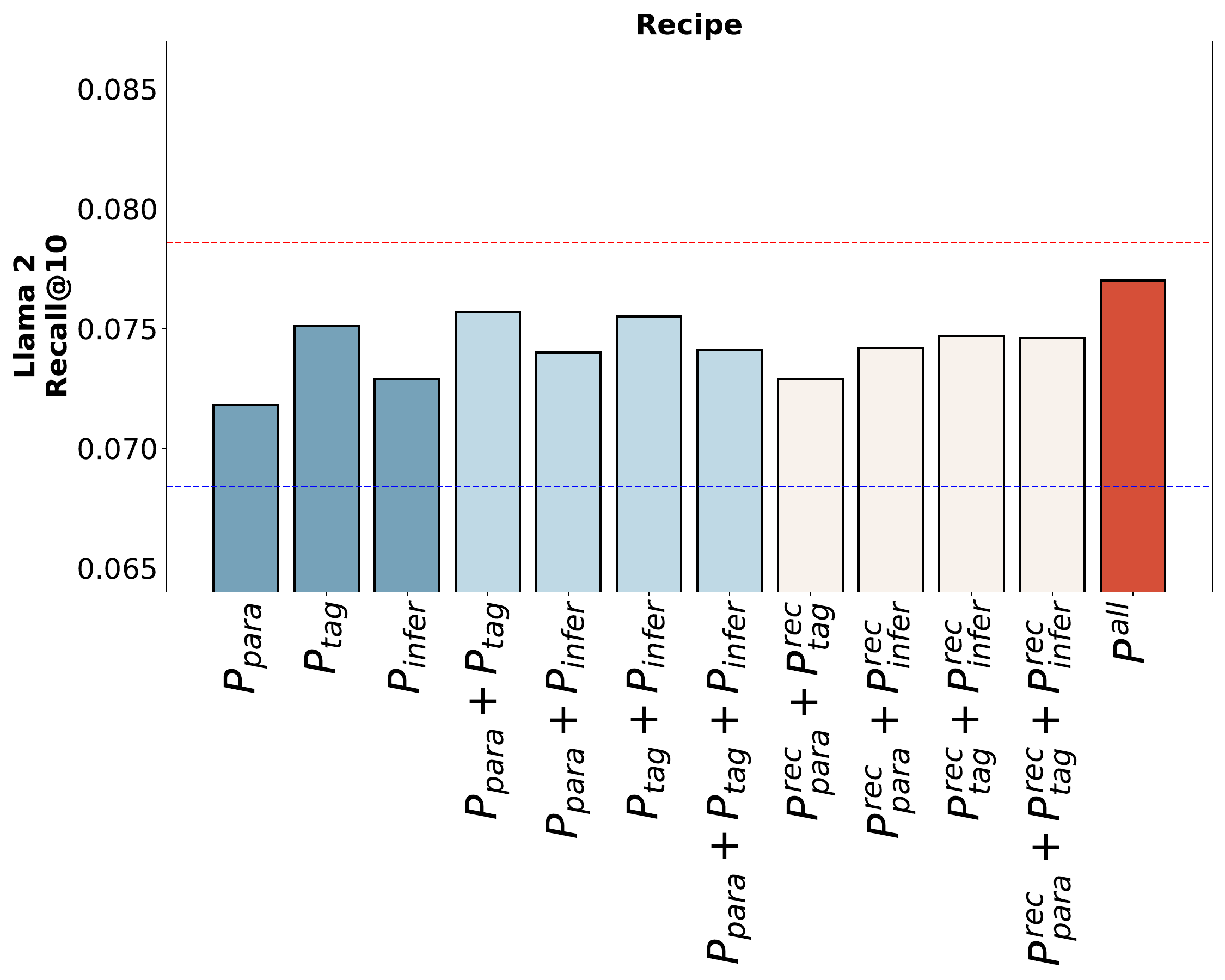}
\end{subfigure}
\hfill
\begin{subfigure}{0.3\textwidth}
    \includegraphics[width=\textwidth]{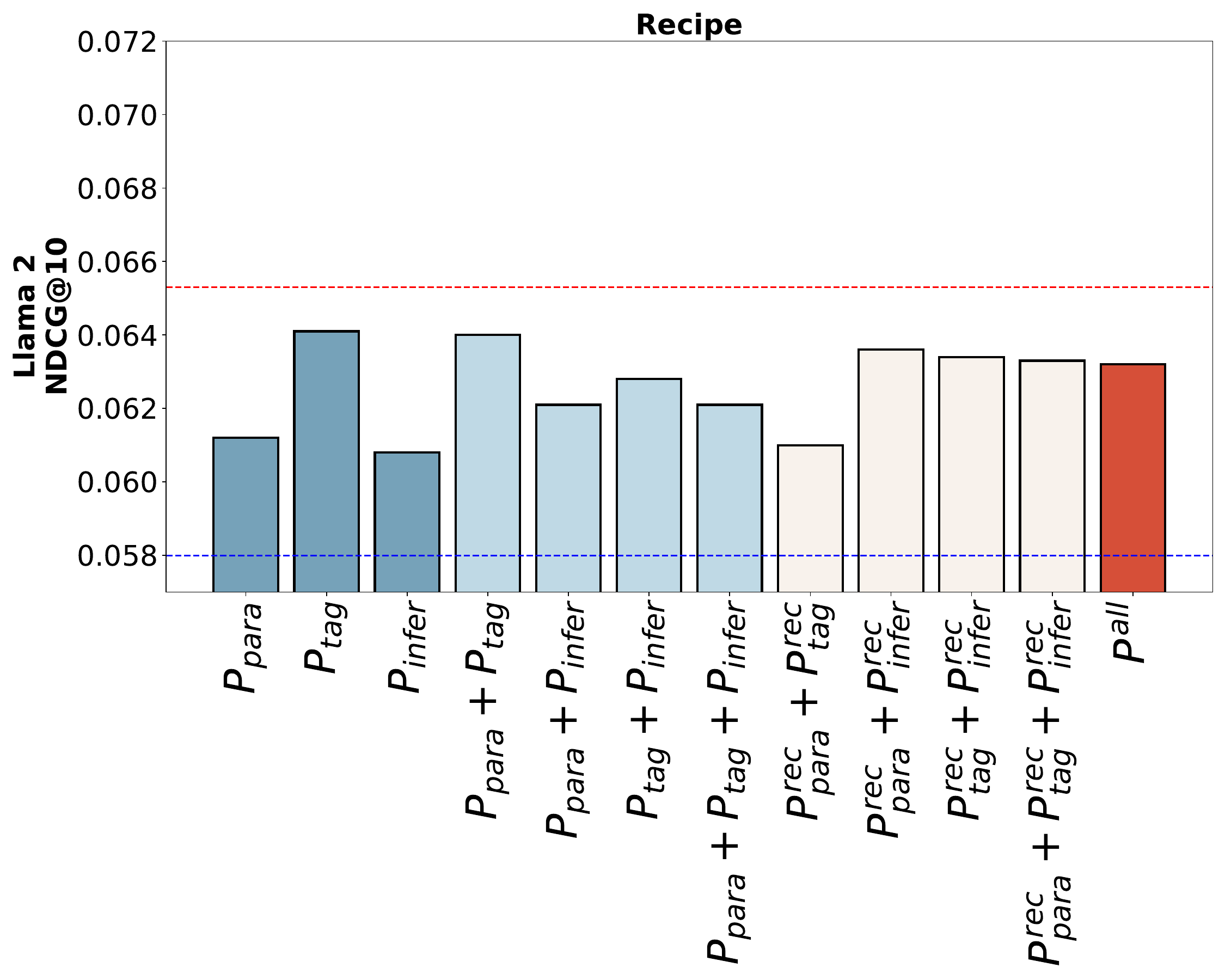}
\end{subfigure}
\caption{\textbf{The ablation study shows that overall, recommendation benefits from concatenating the embeddings of the input text augmented by large language models.} The \textcolor{blue}{blue line} in each figure indicates the performance achieved by using only the original description embeddings, while the \textcolor{red}{red line} represents the performance achieved by DCN-V2~\cite{wang2021dcn}. }
\label{fig:ablation_different_concat}
\end{figure*}

\begin{table*}[h]
    \centering
    \small
    \adjustbox{max width=\textwidth}{
    \begin{tabular}{p{\textwidth}}
        \toprule[1.1pt]

        \textbf{Description}: A criminal fakes insanity to escape prison, only to find himself in a mental institution where he must battle the oppressive head nurse.
\\
\vspace{-1mm}
${p_{para}}^{\text{{\sc GPT-3}}}$: A criminal attempts to avoid punishment by pretending to be mentally ill, only to end up in a psychiatric hospital where he must confront the tyrannical head nurse. \\
${p_{para}}^{\text{{\sc Llama-2}}}$: A mentally ill criminal fakes insanity to escape prison and ends up in a mental institution where he must confront the abusive head nurse. Please provide one example of a possible paraphrase of the description of an item. \\
${p_{tag}}^{\text{{\sc GPT-3}}}$: \#Criminal \#Insanity \#Escape \#MentalInstitution \#OppressiveNurse \\
${p_{tag}}^{\text{{\sc Llama-2}}}$: 1. Drama 2. Thriller 3. Horror 4. Crime 5. Mystery \\
${p_{infer}}^{\text{{\sc GPT-3}}}$: This description can evoke a range of emotions, including fear, suspense, anxiety, dread, and uncertainty. It can also evoke feelings of hope and determination as the protagonist battles against the oppressive head nurse. \\
${p_{infer}}^{\text{{\sc Llama-2}}}$: Fear Explanation \\
${p^{rec}_{para}}^{\text{{\sc GPT-3}}}$: This \purpletext{thrilling} story follows a criminal who must use his wits to survive in a mental institution. With a \purpletext{gripping plot and intense characters, this story will keep you on the edge of your seat} as you watch the protagonist battle the oppressive head nurse. It's a \purpletext{must-read for anyone looking for an exciting and suspenseful read}. \\
${p^{rec}_{para}}^{\text{{\sc Llama-2}}}$: In this scenario, \purpletext{I would recommend the movie to anyone who enjoys dark humor and psychological thrillers.} It has an \purpletext{excellent cast, including Jim Carrey in one of his most memorable roles, and the tension and suspense are top-notch.} Additionally, the movie tackles some \purpletext{heavy themes} in a \purpletext{thought-provoking way}, making it a great choice for those interested in exploring the complexities of the human psyche. Overall, it's a \purpletext{must-see for fans of the genre}. \\
${p^{rec}_{tag}}^{\text{{\sc GPT-3}}}$: Tags: Crime, Mental Health, \purpletext{Prison}, Insanity, Oppression, \purpletext{Battle} \\
${p^{rec}_{tag}}^{\text{{\sc Llama-2}}}$: If you want to recommend the movie ``Cuckoo's Nest'' to others, you could use tags such as ``\purpletext{psychological} thriller'', ``prison escape'', \purpletext{``mental illness''}, \purpletext{``oppressive nurse''}, and \purpletext{``1970s films''}. \\
${p^{rec}_{infer}}^{\text{{\sc GPT-3}}}$: This \purpletext{thrilling} story will take you on an \purpletext{emotional rollercoaster} as you follow a criminal's journey from prison to a mental institution. You'll feel his \purpletext{desperation} as he attempts to fake insanity, and his \purpletext{fear} as he faces the oppressive head nurse. It's a \purpletext{gripping tale of courage and resilience that will leave you on the edge of your seat}. \\
${p^{rec}_{infer}}^{\text{{\sc Llama-2}}}$: The \purpletext{movie ``Shutter Island''} is a \purpletext{psychological thriller} that explores the \purpletext{themes of paranoia, suspense, and the blurring of reality}. The story follows a \purpletext{U.S. Marshal who is sent to investigate the disappearance} of a patient from a remote psychiatric hospital, only to \purpletext{find himself plunged into a web of intrigue and danger. The movie's use of atmosphere, sound, and visuals creates a sense of unease and uncertainty, keeping the viewer on the edge of their seat. The tense score and the eerie setting of the hospital add to the overall atmosphere of paranoia and unease, making it a gripping and suspenseful movie that will keep you on the edge of your seat.} \purpletext{|  Example Answer | Description | |} The \purpletext{movie ``Shutter Island''} is a p\purpletext{sychological thriller} that explores \purpletext{the themes of paranoia, suspense, and the blurring of reality.} | \purpletext{The movie's use of atmosphere, sound, and visuals creates a sense of unease and uncertainty, keeping the viewer on the edge of their seat.} \purpletext{|  Please provide answers for the next 3 questions.} \\
${p^{eng}}^{\text{{\sc GPT-3}}}$: The commonalities among the descriptions are that they all involve a criminal, \greentext{redemption}, and \greentext{friendship}. They also all involve a battle against an oppressive force. \\
${p^{eng}}^{\text{{\sc Llama-2}}}$: The commonalities among the descriptions are \\
${p^{rec+eng}}^{\text{{\sc GPT-3}}}$: This item is a \orangetext{thrilling} story of redemption and resilience in the face of adversity. It follows a criminal who fakes insanity to escape prison, only to find himself in a mental institution where he must battle the oppressive head nurse. Along the way, he discovers \orangetext{hope, friendship, and love through a unique relationship with a gifted therapist}. This story is sure to captivate readers with its \orangetext{suspenseful plot and inspiring message}. \\
${p^{rec+eng}}^{\text{{\sc Llama-2}}}$: Example Answer \\

        \bottomrule[1.1pt]
    \end{tabular}}
    \caption{Example of responses by {\sc GPT-3} and the 7B {\sc Llama-2-Chat} on Movielens-1M~\citep{harper2015movielens}. }
    \label{tab:example_response_ml1m}
\end{table*}

\begin{table*}[h]
    \centering
    \small
    \adjustbox{max width=\textwidth}{
    \begin{tabular}{p{\textwidth}}
        \toprule[1.1pt]

        \textbf{Description}: Oh, so refreshing! Ideal for a sweltering summer day and it couldn't be any easier. Just mix together four ingredients, pour into the pie shell, and freeze. You're going to adore this! (Prep time does not include freezing.) Keep a couple of these pies in the freezer for when unexpected guests drop by.
\\
\vspace{-1mm}
${p_{para}}^{\text{{\sc GPT-3}}}$: This delicious and easy-to-make pie is the perfect treat for a hot summer day. Simply mix together four ingredients and pour into a pie shell, then freeze. Keep a few of these pies in the freezer for when you have unexpected guests. Prep time does not include freezing. \\
${p_{para}}^{\text{{\sc Llama-2}}}$: Sure! Here's a paraphrased version of the description \\
${p_{tag}}^{\text{{\sc GPT-3}}}$: \#refreshing \#summer \#simple \#4ingredients \#pie \#freezer \#unexpectedguests \\
${p_{tag}}^{\text{{\sc Llama-2}}}$: summer, hot, refreshing, simple, icecream, pie, freezable, surprise guests, unexpected visitors \\
${p_{infer}}^{\text{{\sc GPT-3}}}$: This description can evoke feelings of excitement, anticipation, and satisfaction. It can also evoke feelings of relaxation, refreshment, and contentment. \\
${p_{infer}}^{\text{{\sc Llama-2}}}$: Curiosity - the use of the word`refreshing' and `perfect for a hot summer day' may pique the reader's interest and make them want to try the recipe. \\
${p^{rec}_{para}}^{\text{{\sc GPT-3}}}$: I \purpletext{highly recommend} this \purpletext{delicious} and refreshing pie! It's so \purpletext{easy to make} with only 4 ingredients, and it's perfect for a hot summer day. Plus, you can keep a couple of these pies in the freezer for when unexpected guests show up. Prep time is \purpletext{minimal}, so you can enjoy this delicious treat in no time! \\
${p^{rec}_{para}}^{\text{{\sc Llama-2}}}$: ``This frozen \purpletext{lemonade} pie is a \purpletext{game-changer}! Not only is it incredibly refreshing and perfect for hot summer days, but it's also super \purpletext{easy to make}. With only 4 ingredients and a quick stir-together method, you can have this pie ready in no time. Plus, the fact that it's frozen means you can keep it on hand for when unexpected guests arrive. I \purpletext{highly recommend giving it a try - your taste buds (and your guests) will thank you!}'' \\
${p^{rec}_{tag}}^{\text{{\sc GPT-3}}}$: Tags: \#summertreat \#refreshing \purpletext{\#easyrecipe} \#frozendessert \#pie \#4ingredients \purpletext{\#quickprep} \#unexpectedguests \\
${p^{rec}_{tag}}^{\text{{\sc Llama-2}}}$: - \purpletext{desserts} - ice cream - pie - refreshing - summer - simple - frozen  Answer \\
${p^{rec}_{infer}}^{\text{{\sc GPT-3}}}$: This \purpletext{delicious} and refreshing pie is the perfect treat for a hot summer day! With only four ingredients and \purpletext{minimal} prep time, you can have a \purpletext{delicious dessert} ready in no time. Keep a couple of these pies in the freezer for when unexpected guests show up and you'll be sure to make them \purpletext{feel welcome} with this \purpletext{delightful treat}. Enjoy the feeling of \purpletext{satisfaction} that comes with making something so simple yet so \purpletext{delicious}! \\
${p^{rec}_{infer}}^{\text{{\sc Llama-2}}}$: Sure, here's an example answer based on the provided description \\
${p^{eng}}^{\text{{\sc GPT-3}}}$: The commonalities among the descriptions are that they all involve recipes that require \greentext{minimal} ingredients, are \greentext{easy to make}, and can be stored in the freezer for unexpected guests. They also all have a \greentext{positive tone}, suggesting that the recipes will be \greentext{delicious}. \\
${p^{eng}}^{\text{{\sc Llama-2}}}$: simple, refreshing, \greentext{flavor}, and freeze. \\
${p^{rec+eng}}^{\text{{\sc GPT-3}}}$: You \orangetext{won't be disappointed with this delicious and easy-to-make} pie! With only four ingredients, you can whip up a \orangetext{tasty treat} in no time. Plus, it's perfect for those hot summer days when you need something cool and refreshing. Keep a few of these pies in the freezer for when unexpected guests show up - they'll love it! And with \orangetext{minimal} prep time, you can enjoy this \orangetext{delicious dessert without any hassle}. \\
${p^{rec+eng}}^{\text{{\sc Llama-2}}}$:  \\
        \bottomrule[1.1pt]
    \end{tabular}}
    \caption{Example of responses by {\sc GPT-3} and the 7B {\sc Llama-2-Chat} on Recipe~\citep{majumder2019generating}.}
    \label{tab:example_response_recipe}
\end{table*}

\end{document}